%% file: main-acm.tex
\def\arxiv{}
\PassOptionsToPackage{table}{xcolor}
\ifdefined\arxiv
\documentclass[manuscript,nonacm]{acmart}
\else
\documentclass[manuscript]{acmart}
\fi
\AtBeginDocument{%
  }

\input{config/preamble}
\input{config/macros}
\input{navigation_experiment/plots/paper_support}

\ifdefined\arxiv
\newcommand{\supplementname}{the appendix}
\let\arxivnewlabel\newlabel
\def\newlabel#1#2{\arxivnewlabel{#1}{#2}\arxivnewlabel{M-#1}{#2}\arxivnewlabel{S-#1}{#2}}
\else
\newcommand{\supplementname}{the electronic supplement}
\usepackage{xr-hyper}
\fi

\setcopyright{none}
\acmJournal{TAAS}

\begin{document}

\title{Plan your Self-Adaptation! -- Ontology-Mediated Task and Architecture Co-adaptation Planning for Robots}

\author{Gustavo Rezende Silva}
\email{g.rezendesilva@tudelft.nl}
\affiliation{%
  \institution{Cognitive Robotics, Delft University of Technology}
  \city{Delft}
  \country{The Netherlands}
}

\author{Tobias John}
\email{tobiasj@posteo.de}
\affiliation{%
  \institution{Robotron Datenbank Software GmbH}
  \city{Dresden}
  \country{Germany}
}

\author{Francisco Martín}
\email{francisco.rico@urjc.es}
\affiliation{%
  \institution{Intelligent Robotics Lab, Universidad Rey Juan Carlos}
  \city{Fuenlabrada, Madrid}
  \country{Spain}
}

\author{Einar Broch Johnsen}
\email{einarj@ifi.uio.no}
\affiliation{%
  \institution{Department of Informatics, University of Oslo}
  \city{Oslo}
  \country{Norway}
}

\author{Carlos Hernandez Corbato}
\email{c.h.corbato@tudelft.nl}
\affiliation{%
  \institution{Cognitive Robotics, Delft University of Technology}
  \city{Delft}
  \country{The Netherlands}
}

\renewcommand{\shortauthors}{Silva et al.}

\begin{abstract}
  Autonomous robots may need to adapt both their task plans and architectural configurations as operating conditions change. These decisions are interdependent: architectural configurations determine which actions a robot can perform, while task plans determine which capabilities are required. This paper presents \acronym, an ontology-mediated approach to task and architecture co-adaptation that considers these dependencies during mission planning. \acronym uses an ontology expressed in the Web Ontology Language to represent architectural alternatives and their associated functional and quality requirements, while the Planning Domain Definition Language defines task actions and mission goals. By automatically translating and integrating these models, \acronym enables joint planning of task and reconfiguration actions. The approach uses satisficing planning and replans when changes during execution invalidate the current plan. \acronym is evaluated in two robotic scenarios involving underwater inspection and ground navigation. The results support the feasibility of runtime co-adaptation and show lower planning time and memory consumption compared with optimal-policy generation in the evaluated navigation scenario, while producing plans of similar quality. %
\end{abstract}



\maketitle

\input{sections/introduction}

\input{sections/running_example}
\input{sections/related-works}

\input{sections/background}

\input{sections/model}

\input{sections/framework}

\input{sections/implementation}

\input{sections/evaluation}

\input{sections/discussion}

\input{sections/threats}

\input{sections/conclusions}

\begin{acks}
This work was supported by the European Union’s Horizon 2020 Framework Programme through the MSCA network REMARO (Grant Agreement No 956200);  Horizon Europe CoreSense (no. 101070254); and Proyectos de Generación de Conocimiento 2021 call of Ministry of Science and Innovation of Spain co-funded by the European Union, project PID2021-126592OB-C22 CASCAR/DMARCE\@.
\end{acks}

\bibliographystyle{styles/ACM-Reference-Format}
\bibliography{references}

\ifdefined\arxiv
\appendix
\input{sections/supplement}
\fi

\end{document}

%% file: config/preamble.tex
\usepackage{graphicx} %
\usepackage{csquotes}

\usepackage{todonotes}

\usepackage{algorithm}
\usepackage{algpseudocode}
\usepackage{amsmath}

\usepackage{amssymb}
\usepackage{listings}
\usepackage{cleveref}
\usepackage{booktabs} %

\usepackage{xspace}
\usepackage{multirow}

\usepackage{subcaption}

\usepackage{tabularx}
\usepackage{makecell}  %
\raggedright

\usepackage{pgfplots}
\usepackage{pgfplotstable}
\pgfplotsset{compat=1.18} 

\usetikzlibrary{calc, arrows.meta, decorations.pathmorphing, shapes}

\usepgfplotslibrary{fillbetween, statistics}

\definecolor{codegreen}{rgb}{0,0.6,0}
\definecolor{codegray}{rgb}{0.5,0.5,0.5}
\definecolor{codepurple}{rgb}{0.58,0,0.82}
\definecolor{backcolour}{rgb}{0.95,0.95,0.92}

\lstdefinestyle{mystyle}{
    frame=single,
    commentstyle=\color{blue}\sl\sffamily,
    morecomment=[l]{\#},
    keywordstyle=\color{black}\ttfamily\bfseries,
    numberstyle=\tiny\color{codegray},
    stringstyle=\color{olive},
    basicstyle=\sffamily\footnotesize,
    columns=fullflexible,
    numberstyle=\tiny,
    breakatwhitespace=false,         
    breaklines=true,                 
    captionpos=b,                    
    keepspaces=true,                 
    numbers=left,                    
    numbersep=5pt,                  
    showspaces=false,                
    showstringspaces=false,
    showtabs=false,                  
    tabsize=2,
    literate={\{}{{\color{purple}{\{}}}1 {\}}{{\color{purple}{\}}}}1
}
\lstdefinelanguage{PDDL}
{
  sensitive=false,    %
  morecomment=[l]{;}, %
  alsoletter={:,-,?},   %
  keywords=[1]{ },
  keywords=[2]{and, or, over,all, at, end, not},
  keywords=[3]{
      define,domain,problem,when,forall,exists,either,
      :domain,:requirements,:types,:objects,:constants, %
      :predicates,:action,:parameters,:precondition,:effect, %
      :fluents,:primary-effect,:side-effect,:init,:goal, %
      :strips,:adl,:equality,:typing,:conditional-effects, %
      :negative-preconditions,:disjunctive-preconditions, %
      :existential-preconditions,:universal-preconditions,:quantified-preconditions, %
      :functions,assign,increase,decrease,scale-up,scale-down, %
      :metric,minimize,maximize, %
      :durative-actions,:duration-inequalities,:continuous-effects, %
      :durative-action,:duration,:condition, %
      :derived, %
  },
  keywordstyle=[2]\color{\blueColor}\bfseries,
  keywordstyle=[3]\color{\greenColor}\bfseries,
  keywordsprefix=?
}

\definecolor{pvalue_green}{RGB}{210,245,210} %
\definecolor{pvalue_blue}{RGB}{220,230,245}  %

%% file: config/macros.tex
\newcommand{\pddlTask}{\ensuremath{\Pi}\xspace}
\newcommand{\pddlDomain}{\ensuremath{\mathcal{D}}\xspace}
\newcommand{\pddlInstance}{\ensuremath{\mathcal{P}}\xspace}
\newcommand{\pddlObjects}{\ensuremath{O}\xspace}
\newcommand{\pddlInitial}{\ensuremath{I}\xspace}
\newcommand{\pddlGoal}{\ensuremath{G}\xspace}
\newcommand{\pddlState}{\ensuremath{S}\xspace}

\newcommand{\pddlPredicates}{\ensuremath{\mathsf{N}_\mathsf{P}}\xspace}
\newcommand{\pddlActions}{\ensuremath{A}\xspace}
\newcommand{\pddlDerRules}{\ensuremath{\text{Der}}\xspace}

\newcommand{\pddlActParam}{\ensuremath{\vec{x}}\xspace}
\DeclareMathOperator{\pddlActName}{name}
\DeclareMathOperator{\pddlActPre}{pre}
\DeclareMathOperator{\pddlActAdd}{add}
\DeclareMathOperator{\pddlActDel}{del}
\DeclareMathOperator{\pddlActionReq}{actionReqs}

\newcommand{\Ont}{\ensuremath{\mathcal{O}}\xspace}

\newcommand{\N}{\ensuremath{\mathsf{N}}\xspace}
\newcommand{\NC}{\ensuremath{\mathsf{N}_\mathsf{C}}\xspace}
\newcommand{\NR}{\ensuremath{\mathsf{N_R}}\xspace}
\newcommand{\NI}{\ensuremath{\mathsf{N_I}}\xspace}

\newcommand{\NRobject}{\ensuremath{\mathsf{N_R^{object}}}\xspace}
\newcommand{\NRdata}{\ensuremath{\mathsf{N_R^{data}}}\xspace}

\newcommand{\ALCI}{\ensuremath{\mathcal{ALCI}}\xspace}

\DeclareMathOperator{\antecedent}{antecedent}
\DeclareMathOperator{\consequent}{consequent}

\DeclareMathOperator{\FunctionTomasys}{Function}
\DeclareMathOperator{\FunctionsTomasys}{Functions}
\DeclareMathOperator{\FunctionDesign}{FunctionDesign}
\DeclareMathOperator{\FunctionDesigns}{FunctionDesigns}
\DeclareMathOperator{\Component}{Component}
\DeclareMathOperator{\QAtype}{QAtype}
\DeclareMathOperator{\QAtypes}{QAtypes}
\DeclareMathOperator{\QAvalue}{QAvalue}
\DeclareMathOperator{\Objective}{Objective}

\DeclareMathOperator{\FunctionGrounding}{FunctionGrounding}

\DeclareMathOperator{\solvesF}{solvesF}
\DeclareMathOperator{\requiredBy}{requiredBy}
\DeclareMathOperator{\isQAtype}{isQAtype}
\DeclareMathOperator{\hasQAestimation}{hasQAestimation}

\DeclareMathOperator{\cStatus}{c\_status}

\DeclareMathOperator{\fdRealisability}{fd\_realisability}

\DeclareMathOperator{\qaHasValue}{qa\_has\_value}

\DeclareMathOperator{\Action}{Action}
\DeclareMathOperator{\fdBetterUtility}{fdBetterUtility}
\DeclareMathOperator{\functionGrounding}{functionGrounding}
\DeclareMathOperator{\requiresC}{requiresC}
\DeclareMathOperator{\requiresF}{requiresF}
\DeclareMathOperator{\fActive}{f\_active}
\DeclareMathOperator{\fdUtility}{fd\_utility}

\DeclareMathOperator{\fdSpiralMed}{fd\_spiral\_medium}

\DeclareMathOperator{\fdSpiralHigh}{fd\_spiral\_high}
\DeclareMathOperator{\aSearchPipelineOwl}{a\_search\_pipeline}
\DeclareMathOperator{\fGenerateSearchPath}{f\_generate\_search\_path}

\DeclareMathOperator{\fFollowPipeline}{f\_follow\_pipeline}

\DeclareMathOperator{\varC}{?c}
\DeclareMathOperator{\varFD}{?fd}
\DeclareMathOperator{\varFDtwo}{?fd2}

\DeclareMathOperator{\varF}{?f}
\DeclareMathOperator{\varU}{?u}
\DeclareMathOperator{\varUtwo}{?u2}

\DeclareMathOperator{\varQA}{?qa}
\DeclareMathOperator{\varQAvalue}{?qav}

\DeclareMathOperator{\fdUnground}{fd\_unground}

\DeclareMathOperator{\instanceFD}{fd}
\DeclareMathOperator{\instanceF}{f}
\DeclareMathOperator{\instanceA}{a}

\DeclareMathOperator{\instanceTrue}{true}
\DeclareMathOperator{\instanceFalse}{false}

\DeclareMathOperator{\bestFD}{bestFD}

\DeclareMathOperator{\functionalReq}{functionalReq}

\DeclareMathOperator{\algE}{E}
\DeclareMathOperator{\algEff}{Eff}
\DeclareMathOperator{\algDerive}{Derive}
\DeclareMathOperator{\algGraph}{Graph}
\DeclareMathOperator{\algRoots}{Roots}
\DeclareMathOperator{\algS}{S}
\DeclareMathOperator{\algC}{C}
\DeclareMathOperator{\algP}{P}
\DeclareMathOperator{\algNodes}{Nodes}
\DeclareMathOperator{\algSCC}{SCC}
\DeclareMathOperator{\algSCCs}{SCCs}

\newcommand{\hide}[1]{}

\newcommand\owltopddl{Owl2PDDL\xspace}
\newcommand\acronym{PLANTA\xspace}
\newcommand{\CamaraApproach}{\textsc{Camara}\xspace}
\newcommand\plansys{PlanSys2\xspace}
\newcommand\tomasys{TOMASys\xspace}
\newcommand\pddltomasys{PDDL-TOMASys\xspace}

\newcommand{\Uone}{\emph{U1}\xspace}
\newcommand{\Utwo}{\emph{U2}\xspace}
\newcommand{\Uthree}{\emph{U3}\xspace}

\newcommand{\Sone}{\emph{S1}\xspace}
\newcommand{\Stwo}{\emph{S2}\xspace}
\newcommand{\Sthree}{\emph{S3}\xspace}
\newcommand{\Sfour}{\emph{S4}\xspace}
\newcommand{\Sfive}{\emph{S5}\xspace}

\definecolor{MidnightBlue}{HTML}{006795}
\definecolor{OliveGreen}{HTML}{3C8031}
\definecolor{BrickRed}{HTML}{B6321C}
\definecolor{Purple}{HTML}{800080}
\definecolor{Orange}{HTML}{F59527}
\newcommand{\greenColor}{OliveGreen} 	%
\newcommand{\blueColor}{MidnightBlue}	%

\newcommand{\purpleColor}{Purple}
\newcommand{\orangeColor}{Orange}

\tikzstyle{lineDefault} = [
    line width=0.75pt
]

\tikzstyle{artifact} = [
lineDefault,
draw=black,
inner sep=3pt,
minimum width=1cm,
minimum height=0.5cm,
font=\sffamily,
fill=white!20
]

\tikzstyle{artifact_transparent} = [
lineDefault,
draw=black,
inner sep=3pt,
minimum width=1cm,
minimum height=0.5cm,
font=\sffamily,
]

\tikzstyle{process} = [
artifact,
rounded corners=0.2cm,
minimum height=0.3cm,
fill=black!5
]

\tikzstyle{tool} = [
    process,
    fill=\greenColor!20,
    draw=\greenColor
]

\tikzstyle{tool_3rd_party} = [
    process,
    fill=\blueColor!20,
    draw=\blueColor
]

\tikzstyle{system} = [
    process,
    fill=\purpleColor!5,
    draw=\purpleColor
]

\tikzstyle{runtime} = [
    artifact,
    fill=\blueColor!20,
    draw=\blueColor
]

\tikzstyle{arrow} = [
    line width=1pt,
    -Latex
]

\tikzstyle{arrow_dashed} = [
    line width=1pt,
    dashed,
    -Latex
]

\tikzstyle{bubble} = [
    ellipse,
    line width=1pt,
    draw
]

\tikzstyle{bubblePurple} = [
    bubble,
    fill=\purpleColor!20,
    draw=\purpleColor
]

\tikzstyle{bubbleGreen} = [
    bubble,
    fill=\greenColor!20,
    draw=\greenColor
]

\tikzstyle{bubbleBlue} = [
bubble,
fill=\blueColor!20,
draw=\blueColor
]

\newcommand{\gridNodeSize}{3pt}
\newcommand{\gridNodeLine}{0.75pt}	%
\newcommand{\gridNodeSizePlus}{\gridNodeSize+\gridNodeLine+\gridNodeLine}

\tikzstyle{gridNode} = [
    minimum height=\gridNodeSize,
    minimum width=\gridNodeSize,
    inner sep=2.5pt,
    draw=black!70,
    line width=\gridNodeLine,
    fill=white,
    shape=circle
]

\tikzstyle{corridor} = [
    line width=\gridNodeLine,
    double,
    double distance=\gridNodeSize,
    draw=black!70
]

\tikzstyle{unsafe} = [
    decorate=true,
    decoration=snake,
    line cap=round
]

\tikzstyle{dark} = [
    draw=black!80!\blueColor!60!white
]

\tikzstyle{corridorDark} = [
    line width=\gridNodeSizePlus,
    dark
]

\tikzstyle{corridorUnsafe} = [
    unsafe,
    corridor
]

\tikzstyle{corridorDarkUnsafe} = [
    unsafe,
    line width=\gridNodeSizePlus,
    dark
]

\tikzstyle{goal} = [
    fill=\greenColor,
    text=white,
    shape=circle,
    draw=black!70,
    line width=\gridNodeLine
]
\tikzstyle{start} = [
    fill=\greenColor,
    text=white,
    shape=circle,
    draw=black!70,
    line width=\gridNodeLine
]

\tikzstyle{hideNode} = [
    fill=white,
    minimum height=\gridNodeSize+\gridNodeLine+7pt,
    minimum width=\gridNodeSize+\gridNodeLine+7pt,
    shape=circle
]

\pgfdeclarelayer{background}
\pgfsetlayers{background,main}

\makeatletter
\pgfplotsset{
    boxplot prepared from table/.code={
        \def\tikz@plot@handler{\pgfplotsplothandlerboxplotprepared}%
        \pgfplotsset{
            /pgfplots/boxplot prepared from table/.cd,
            #1,
        }
    },
    /pgfplots/boxplot prepared from table/.cd,
    table/.code={\pgfplotstablecopy{#1}\to\boxplot@datatable},
    row/.initial=0,
    make style readable from table/.style={
        #1/.code={
            \pgfplotstablegetelem{\pgfkeysvalueof{/pgfplots/boxplot prepared from table/row}}{##1}\of\boxplot@datatable
            \pgfplotsset{boxplot/#1/.expand once={\pgfplotsretval}}
        }
    },
    make style readable from table=lower whisker,
    make style readable from table=upper whisker,
    make style readable from table=lower quartile,
    make style readable from table=upper quartile,
    make style readable from table=median,
    make style readable from table=lower notch,
    make style readable from table=upper notch
}
\makeatother

%% file: navigation_experiment/plots/paper_support.tex
\pgfplotsset{compat=1.18}
\usepgfplotslibrary{statistics,fillbetween}
\definecolor{paperPlanta}{cmyk}{0.64,0,0.95,0.40}
\definecolor{paperCamara}{cmyk}{0.98,0.13,0,0.43}
\definecolor{paperBaseline}{cmyk}{0,0.61,0.87,0}
\providecommand{\greenColor}{paperPlanta}
\providecommand{\blueColor}{paperCamara}
\providecommand{\orangeColor}{paperBaseline}
\providecommand{\acronym}{PLANTA}
\providecommand{\CamaraApproach}{\textsc{Camara}}
\providecommand{\paperEagerLabel}{\acronym}

%% file: sections/introduction.tex
\section{Introduction}

Autonomous robots are required to operate reliably in increasingly complex environments and to perform a larger number of tasks with distinct functional and quality requirements, potentially requiring different architectural configurations to satisfy them. A key challenge in designing such systems is ensuring that the robot continues to meet its requirements throughout mission execution, particularly under uncertainty, such as sensor failures or environmental disturbances~\cite{Correll-2017, Krotkov-2018}. For example, an underwater robot might require different sensor configurations depending on the water visibility, or it might need to navigate through different paths depending on water currents. Addressing these challenges requires coordinating task execution and architectural adaptation while accounting for their interdependencies.

Task and architecture co-adaptation (TACA)~\cite{CAMARASoftwareArchitectureTask2020} addresses the interdependencies between task execution and architectural adaptation. Reactive TACA approaches such as ROSA~\cite{rosa} select a configuration for each action only when that action begins executing, without accounting for how the choice constrains the rest of the mission. However, selecting a suitable configuration for the current action alone does not establish whether subsequent actions can be performed under the conditions anticipated along the mission. Considering these dependencies during mission planning allows configuration requirements to influence the task plan before execution begins, so that each configuration choice is made with its downstream task consequences in view. To the best of our knowledge, the approach by Cámara et al.~\cite{CAMARASoftwareArchitectureTask2020} is the only approach that jointly plans task actions and reconfiguration actions and produces a complete plan ahead of execution. Their approach, however, requires substantial planning time and memory and relies on domain models expressed across multiple specification languages with manually developed integration code, increasing modeling effort and the potential for errors. Furthermore, their approach requires modeling probabilities to enable optimal planning, although these probabilities can be difficult to estimate accurately.

To address these limitations, this paper proposes \acronym, an ontology-mediated approach that formulates task and architecture co-adaptation as a single planning problem by automatically integrating task and architectural knowledge. \acronym uses a Web Ontology Language (OWL) ontology~\cite{aguado2024survey} to represent the functional requirements of robot actions, the architectural alternatives that realize those requirements, and the quality attributes and requirements used to assess the applicability and utility of each alternative. The Planning Domain Definition Language (PDDL)~\cite{karpas2020automated} describes task actions, the current system state, and mission goals. %
Architectural knowledge is automatically translated into PDDL and integrated with the task model, allowing task actions and architectural reconfiguration actions to be considered within the same search space. This integration preserves a separation between task modeling and architectural knowledge modeling. The resulting problem can be solved by standard PDDL planners. To support timely runtime adaptation, \acronym favors satisficing planning over guaranteed mission-level optimality. Rather than explicitly modeling uncertainty and its probabilities, it plans from the currently observed state and handles changes during execution through replanning when monitored conditions invalidate the current plan. By relying on two established modeling languages and automating their integration, \acronym aims to reduce modeling effort.

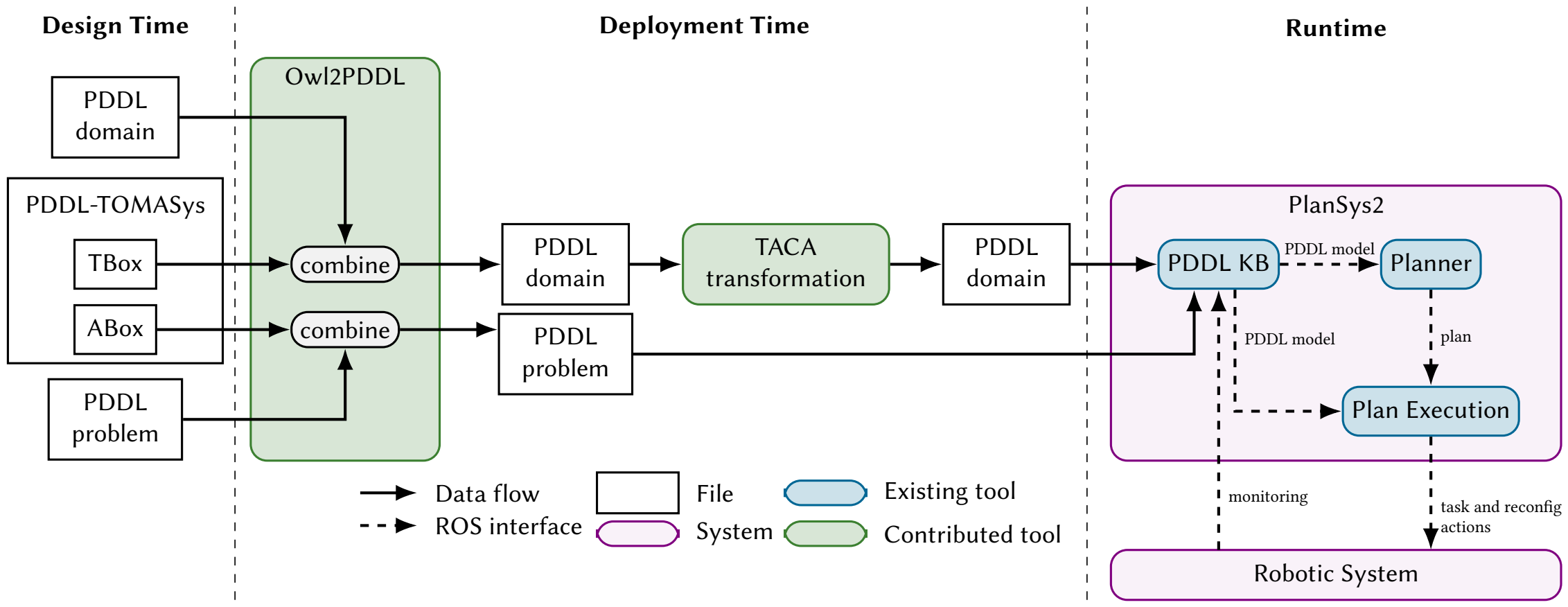
\begin{figure}[tbp]
	\centering
        \resizebox{\textwidth}{!}{
            \input{figures/framework.tex}
        }
		\caption{\acronym framework, organized into design time, deployment time, and runtime, separated by dashed vertical lines. Some \plansys nodes are omitted for clarity, and the Planner and Plan Execution components shown were extended as part of this work.}%
	\label{fig:architecture}
\Description{Left-to-right diagram of the \acronym workflow. At design time, a PDDL domain and
problem and a \pddltomasys ontology are specified. At deployment time, \owltopddl
combines the ontology TBox with the domain and the ABox with the problem. The TACA
transformation then modifies the domain. At runtime, the resulting domain and problem
populate the \plansys PDDL knowledge base, which supplies the planner and plan execution
component. The planner sends a plan to plan execution, which sends task and
reconfiguration actions to the robotic system. Monitoring feeds observations from the
robot back to the knowledge base. Solid arrows denote data flow and dashed arrows denote
ROS interfaces.}
\end{figure}

\Cref{fig:architecture} depicts the \acronym framework across design time, deployment time, and runtime. At design time, the system architecture is modeled using the \pddltomasys ontology, and the task problem is modeled in PDDL\@. At deployment time, \owltopddl translates the ontology into PDDL, combines it with the planning problem, and incorporates their mutual dependencies, i.e., TACA concerns. At runtime, the planning system, based on \plansys, jointly plans and executes task and reconfiguration actions. \pddltomasys builds upon the Teleological and Ontological Metamodel for Autonomous Systems~(\tomasys)~\cite{aguado2021functional,corbato2013model} and draws inspiration from the ROSA model~\cite{rosa}. %

\acronym is evaluated with respect to feasibility, performance efficiency, and scalability on two use cases: an autonomous underwater vehicle~(AUV) scenario from the SUAVE exemplar~\cite{suave}, and an autonomous ground vehicle~(AGV) scenario adapted from~\cite{CAMARASoftwareArchitectureTask2020}. These dimensions are formulated as research questions RQ1--RQ3 in \Cref{sec:evaluation}.
Feasibility (RQ1) is addressed by implementing \acronym and integrating it with \plansys, an existing planning system for robotics. This required extending \plansys with derived predicates and existential preconditions~\cite{edelkamp2004pddl2}, PDDL constructs that the models generated by \owltopddl rely on.
The results of running \acronym on the ROS~2-based SUAVE exemplar support the feasibility of applying it to robotic systems.
Performance efficiency (RQ2) is assessed on the AGV use case, where \acronym is, at the median, $3.00$ times faster than the approach by Cámara et al.~\cite{CAMARASoftwareArchitectureTask2020} and its peak memory usage is $6.84$ times smaller, while both approaches produce plans with near-identical path length and estimated remaining battery at the end of the mission.
In the AGV use case, scalability (RQ3) is assessed by varying the map size, the number of architectural configurations, the number of quality requirements, and the number of mission-level PDDL action schemas, independently and in combination. The results show that satisficing planning remains tractable in scenarios where optimal planning becomes impractical, supporting the use of \acronym in robotic applications that require online task-level replanning.

The main contributions of this paper are:
\begin{enumerate}
    \item an ontology-mediated planning approach for mission-wide task and architecture co-adaptation that combines task and reconfiguration actions in a single PDDL formulation and supports satisficing planning with runtime replanning.
    \item an open-source implementation of the proposed approach, designed for integration with ROS~2-based robotic systems, including an extension of the \plansys framework to support derived predicates and existential preconditions in PDDL\@.
    \item a systematic evaluation of the proposed approach, including a comparison with the state-of-the-art approach by Cámara et al.~\cite{CAMARASoftwareArchitectureTask2020} and a scalability analysis that demonstrates the feasibility of satisficing planning in regimes where optimal planning becomes impractical.
\end{enumerate}

%% file: figures/framework.tex
\begin{tikzpicture}[font=\sffamily]

    \coordinate (origin) at (0,0);
    \draw (origin) node (design_time) {\textbf{Design Time}};

    \draw ($(design_time) + (0, -1.1)$) node[artifact] (domain)
    {\begin{tabular}{c}
        PDDL\\
        domain
    \end{tabular}};

    \draw ($(domain) + (0.0, -1.1)$) node (ontology) {PDDL-TOMASys};

    \draw ($(ontology) + (0.0, -0.7)$) node[artifact] (tbox) {TBox\strut};
    \draw ($(tbox) + (0.0, -0.8)$) node[artifact] (abox) {ABox\strut};

    \coordinate (bottom_point_ontology) at (abox.south -| ontology.east);
    \draw[artifact_transparent] ($(ontology.north west) + (-0.1, 0.1)$) rectangle ($(bottom_point_ontology) + (0.1, -0.1)$);

    \draw ($(abox) + (0, -1.1)$) node[artifact] (problem)
    {\begin{tabular}{c}
        PDDL\\
        problem
    \end{tabular}};

    \coordinate (domain1_north_ontology_east) at (domain.north -| ontology.east);
    \draw ($(domain1_north_ontology_east) + (1.6, 0)$) node[minimum width=2.3cm] (owlTool) {\owltopddl};
    \begin{pgfonlayer}{background}
        \coordinate (bottom_owl2pddl) at (problem.south -| owlTool.east);
        \draw[tool] (owlTool.north west) rectangle ($(bottom_owl2pddl)$);
    \end{pgfonlayer}

    \draw (owlTool |- tbox) node[process] (combine_domain) {combine} ;
    \draw (owlTool |- abox) node[process] (combine_problem) {combine}  ;

    \draw[arrow] (domain) -| (combine_domain);
    \draw[arrow] (problem) -| (combine_problem);
    \draw[arrow] (tbox) -- (combine_domain);
    \draw[arrow] (abox) -- (combine_problem);

    \draw ($(combine_domain) + (2.7, 0.0)$) node[artifact] (domain2)
    {\begin{tabular}{c}
        PDDL\\
        domain
    \end{tabular}};
    \draw ($(combine_problem) + (2.7, -0.3)$) node[artifact] (problem2)
    {\begin{tabular}{c}
        PDDL\\
        problem
    \end{tabular}};
    \draw[arrow] (combine_domain) -- (domain2);
    \draw[arrow] (combine_problem) -- (combine_problem -| problem2.west);

    \draw ($(domain2) + (2.7, 0)$) node[tool] (taca)
    {\begin{tabular}{c}
        TACA\\
        transformation
    \end{tabular}};

    \draw[arrow] (domain2) -- (taca);

    \draw ($(origin.north -| taca) + (-1,0)$) node (deployment_time) {\textbf{Deployment Time}};

    \draw ($(taca) + (2.7, 0.0)$) node[artifact] (domain3)
    {\begin{tabular}{c}
        PDDL\\
        domain
    \end{tabular}};
    \draw[arrow] (taca) -- (domain3);

    \coordinate (domain1_north_domain3_east) at (domain3.north -| domain3.east);
    \draw ($(domain1_north_domain3_east) + (3.25, 0.2)$) node[minimum width=5.5cm] (plansys) {PlanSys2};
    \begin{pgfonlayer}{background}
        \coordinate (bottom_plansys) at (problem.south -| plansys.east);
        \draw[system] (plansys.north west) rectangle ($(bottom_plansys)$);
    \end{pgfonlayer}

    \draw ($(domain3) + (2.6, 0.0)$) node[tool_3rd_party] (pddl_kb){PDDL KB\strut};
    \draw[arrow] (domain3) -- (pddl_kb);
    \coordinate (problem2_pddl_kb_in) at (pddl_kb.south -| pddl_kb);
    \draw[arrow] (problem2) -| ($(pddl_kb.south -| pddl_kb) + (-0.3, 0)$);

    \draw ($(pddl_kb) + (2.6, 0.0)$) node[tool_3rd_party] (planner){Planner\strut};
    \draw[arrow_dashed] (pddl_kb) -- (planner)
        node[midway, above, font=\scriptsize] {PDDL model};

    \draw ($(planner) + (0.0, -1.8)$) node[tool_3rd_party] (plan_exec){Plan Execution\strut};
    \draw[arrow_dashed] (planner) -- (plan_exec)
        node[midway, right, font=\scriptsize] {plan};
    \coordinate (pddlkb_plan_exec_arrow) at (pddl_kb.south -| pddl_kb);
    \draw[arrow_dashed] ($(pddlkb_plan_exec_arrow) + (0.2, 0.0)$) |- (plan_exec)
        node[pos=0.2, right, font=\scriptsize] {PDDL model};

    \draw ($(problem.south -| plansys) + (0.0, -1.4)$) node[minimum width=5.5cm] (robot){Robotic System\strut};
    \begin{pgfonlayer}{background}
        \coordinate (bottom_plansys) at (problem.south -| plansys.east);
        \draw[system] (robot.north west) rectangle ($(robot.south east) + (0.0, -0.0)$);
    \end{pgfonlayer}

    \draw[arrow_dashed] (plan_exec) -- (robot.north -| plan_exec)
        node[pos=0.7, right, font=\scriptsize, align=left] {task and reconfig\\actions};
    \draw[arrow_dashed] (robot.north -| pddl_kb) -- (pddl_kb)
        node[pos=0.2, right, font=\scriptsize] {monitoring};

    \draw[dashed] ($(robot.south -| ontology.east) + (0.25, 0)$) -- ($(design_time.north -| ontology.east) + (0.25, 0)$);
    \draw[dashed] ($(robot.south -| domain3.east) + (0.2, 0)$) -- ($(design_time.north -| domain3.east) + (0.2, 0)$);

    \draw ($(origin.north -| plansys)$) node (run_time) {\textbf{Runtime}};

    \begin{scope}[shift={(3.0,-5.7)}]
        \draw[arrow] (0,0) -- (0.7,0);
        \node[right] at (0.8,0) {Data flow};
        \draw[arrow_dashed] (0,-0.4) -- (0.7,-0.4);
        \node[right] at (0.8,-0.4) {ROS interface};
        \node[artifact] at (3.4, 0) {};
        \node[right] at (4,0) {File};
        \node[system] at (3.4, -0.5) {};
        \node[right] at (4,-0.5) {System};
        \node[tool_3rd_party] at (5.7, 0.0) {};
        \node[right] at (6.3,0.0) {Existing tool};
        \node[tool] at (5.7, -0.5) {};
        \node[right] at (6.3,-0.5) {Contributed tool};
    \end{scope}
\end{tikzpicture}

%% file: sections/running_example.tex
\section{Running Example}\label{sec:running_example}

Throughout this paper, the SUAVE exemplar~\cite{suave} is used as a running example to illustrate the concepts introduced by \acronym. In SUAVE, an Autonomous Underwater Vehicle (AUV) carries out an underwater pipeline inspection mission. The mission is structured as a sequence of task actions: \emph{A1}~search for the pipeline, and \emph{A2}~follow and inspect the pipeline. The SUAVE scenario is extended in ROSA~\cite{rosa} with a third action, \emph{A3}~recharge battery, so that the scenario requires task and architecture co-adaptation.

During mission execution, the AUV is subject to three sources of uncertainty. \Uone corresponds to thruster failures: if a thruster fails during \emph{A1} or \emph{A2}, the system requires reconfiguring the architecture to recover the affected thruster. \Utwo corresponds to the variation in water visibility: reduced visibility during \emph{A1} causes the system to select a lower-altitude search configuration. Both \Uone and \Utwo require architecture-level adaptation only, leaving the task plan unchanged. \Uthree corresponds to a critical drop in battery level: if battery level reaches a critical threshold during \emph{A1} or \emph{A2}, the system must interrupt the current action and execute \emph{A3}, which requires a dedicated architectural configuration. Hereafter, the scenario with \Uone and \Utwo is called the original use case, and the scenario with all three uncertainties the extended use case. %

In this work, each task action in the mission is encoded as a PDDL action in the planning domain. For example, the \texttt{search\_pipeline} action~(Listing~
\ref{lst:search_pipeline}) encodes \emph{A1}. Throughout the paper, this action is used to illustrate how \acronym transforms a regular action into a managed action that integrates functional and quality requirements derived from the \pddltomasys ontology~(Listing~\ref{lst:search_pipeline_managed}), as described in Section~\ref{sec:plan_transformation_taca}.

%% file: sections/related-works.tex
\section{Related Works}\label{sec:related}

\subsection{Self-Adaptive Robotic Systems}\label{sec:related_sas}

The mapping study on ``robotics software architecture-based self-adaptive systems''~(RSASSs) by Alberts et al.~\cite{ALBERTS2025112258} identified 37 primary studies published since 2011. Among these, only four works~\cite{park2012task, runtime_variability_lotz, rra_nico, CAMARASoftwareArchitectureTask2020} explicitly address, to varying extents, the dependencies between task execution and architectural adaptation. Additionally, the two studies~\cite{braberman2018extended, temoto} also explore this relationship,  and the ROSA framework proposes a systematic solution to TACA~\cite{rosa}. However, only the approaches~\cite{braberman2018extended, CAMARASoftwareArchitectureTask2020,rosa} fully address TACA\@.
With respect to knowledge-based methods, Alberts et al.\ identified six works~\cite{park2012task, stefan_adaptive_runtime_models, graph_based_nico, p20_stefan_ice, p26_darko_jasper, suave} that utilize knowledge representation techniques to encode the information required for adaptation. From these, the works~\cite{p26_darko_jasper, suave} focus on demonstrating the use of Metacontrol~\cite{hernandez2018self, bozhinoski2022mros} in specific robotic applications rather than proposing complete solutions for RSASSs.
Concerning planning, using PDDL-based planners for RSASSs is still mostly unexplored. Alberts et al.\ identified only the approach \enquote{Model Based Diagnosis (MBD)}~\cite{zaman2013integrated, zaman2019fault}, which employs PDDL-based planning to generate architectural adaptation plans, without explicitly targeting TACA\@.

Cámara et al.~\cite{CAMARASoftwareArchitectureTask2020} proposed a solution to TACA that generates an optimal policy for task execution and architectural configuration. Their method is decomposed into four steps: (i) generate all legal architectural configurations based on the robot’s architecture model, (ii) generate all possible reconfiguration plans between configurations, (iii) generate the \emph{N} best task plans to accomplish the robot's mission goal, (iv) compute an optimal policy for task execution and configuration, given the reconfiguration and task plans. Their experiments show that the generation time is impractical: Generating an optimal task and configuration policy for an AGV operating in a graph-like environment with 58 nodes and with six architectural configurations available takes, on average, $15.1$ seconds.

Silva et al.~\cite{rosa} proposed ROSA as a solution to TACA, focusing on providing a reference framework for robot self-adaptation that promotes reusability, composability, and extensibility. A limitation of ROSA is that it only supports reactive adaptation of the robot's configuration. ROSA determines the robot's configuration for each action only when it starts executing, and it does not produce a complete configuration plan ahead of time.

To address the limitations of these approaches, this work proposes an ontology-mediated planning approach that enables the timely computation of a satisficing task and configuration plan without relying on probabilistic models of real-world uncertainties.

\subsection{Combining Ontologies and Planning}\label{sec:related_ontology_planning}

\paragraph{Knowledge and Action Bases.}
Several works have explored combining ontologies with planning problems. The approach proposed in this work is most closely related to the work on \emph{Knowledge and Action Bases (KABs)}~\cite{DBLP:journals/jair/HaririCMGMF13}, where an ontology is used to provide additional insights about a planning state by using queries to the ontology in the planning domain. The actions in the planning domain describe changes to the ontology over time.

There are different approaches to generate a plan for KABs, namely (i) rewriting ontology queries to complex preconditions~\cite{DBLP:conf/ijcai/CalvaneseMPS16}, (ii) rewriting ontology axioms to derived predicates~\cite{DBLP:conf/aaai/Borgwardt0KKNS22}, and (iii) enriching the planning domain with grounded answers to the ontology queries~\cite{DBLP:conf/ecai/JohnK24}. All the approaches have in common that they perform planning as a two-step process: First, the ontology is analyzed and a new planning domain is created that encodes the axioms from the ontology. Secondly, an off-the-shelf planner is used to find a plan using the new domain and the original problem. The first approach only works for axioms in DL-Lite~\cite{DBLP:conf/ijcai/CalvaneseMPS16} and leads to rather long planning times~\cite{DBLP:conf/aaai/Borgwardt0KKNS22}, despite optimizations in the expansion algorithm~\cite{DBLP:conf/kr/Borgwardt0KS21}. The second approach requires a Horn-Logic fragment of the ontology axioms, but it is, in general, the fastest among the three~\cite{DBLP:conf/aaai/Borgwardt0KKNS22, DBLP:conf/ecai/JohnK24}. The third approach supports all kinds of axioms, but it is slower than the second approach, especially in instances where finding a plan is relatively simple~\cite{DBLP:conf/ecai/JohnK24}.

The approach presented in this work is most similar to the second approach, where ontology axioms are encoded as derived predicates in the planning formulation. However, due to the special requirements of TACA, a custom translation from axioms to derived predicates is used.

\paragraph{Ontologies as Modeling Languages.} Some studies have explored the use of ontologies as a modeling language for planning problems in the robotics domain~\cite{DBLP:journals/ras/BalakirskyKKPSG13,DBLP:journals/ewc/BorgoCOU19}. These efforts try to make modeling of complex domains easier by providing frameworks for complex classes and their relations. It is a known limitation of PDDL that modeling domains with complex types directly in PDDL is not well supported~\cite{GOLDMANpddlTypeProblems2012}. This work adopts a related approach by encoding PDDL type information within the ontology to support semantically rich planning representations.

%% file: sections/background.tex
\section{Background}\label{sec:background}

\subsection{PDDL Planning}\label{sec:bg_planning}

This work adopts a standard definition of planning problems based on a first-order (lifted) planning formalism, as defined in PDDL~\cite{fox2003pddl2}.

A \emph{planning task} is defined as a tuple $\pddlTask = \langle \pddlDomain, \pddlInstance \rangle$, where \pddlDomain denotes the planning domain and \pddlInstance the problem instance. The domain and the problem instance must be defined over the same finite set of predicate symbols \pddlPredicates.

A \emph{planning problem instance} is defined as a tuple $\pddlInstance= \langle \pddlObjects, \pddlInitial, \pddlGoal \rangle$, where \pddlObjects is a finite set of objects, \pddlInitial is the initial state represented as a set of ground atoms, and \pddlGoal is the goal condition, i.e., a set of ground atoms that must be satisfied in the final state. Ground atoms are constructed using predicate symbols from \pddlPredicates and constants/objects from \pddlObjects.

A \emph{planning domain} is defined as a tuple $\pddlDomain = \langle \pddlActions, \pddlDerRules \rangle$ where \pddlActions is a set of actions and \pddlDerRules is a set of derivation rules.
An \emph{action} is a tuple $\instanceA = \langle \pddlActName, \pddlActParam, \pddlActPre, \pddlActAdd, \pddlActDel \rangle$, where $\pddlActName$ is the action name, $\pddlActParam$ are the action parameters, $\pddlActPre$ is a first-order formula describing the precondition that needs to be satisfied to perform the action, $\pddlActAdd$ and $\pddlActDel$ represent, respectively, the set of atoms that are added and deleted from the current state when the action is performed. A grounded action $\underline{a}$ is an action $a$ in which the variables in $\pddlActParam$ are substituted by objects from \pddlObjects.
A \emph{derivation rule} is of the form $p(\pddlActParam) \leftarrow \phi(\pddlActParam)$ where $\pddlActParam$ is a vector of variables, $p \in \pddlPredicates$ and $\phi(\pddlActParam)$ is a first-order formula with free variables $\pddlActParam$ describing the condition when $p$ should hold. All predicates $p$ that occur in such a rule are called \emph{derived predicates}.

\emph{Planning} is the search for a \emph{plan} $\pi = (\underline{a_{1}}, \ldots{}, \underline{a_{n}})$, i.e., a finite list of grounded actions, where the ordered application of the grounded actions produces a state $s_{\text{goal}}$ that contains the goal atoms, i.e. $\pddlGoal \subseteq s_{\text{goal}}$.

\subsection{PlanSys2}\label{sec:bg_plansys}

This work uses \plansys for automated planning with off-the-shelf planners, for runtime plan execution, and for managing the PDDL knowledge base, which is updated at runtime with information from the robot's perception system.

\plansys transforms task plans into plan graphs that capture dependencies between actions and allow non-interfering actions to execute in parallel. These graphs are then converted into behavior trees (BTs)~\cite{pddl_plans_bt} for runtime execution. During execution, the BT continuously checks the action preconditions specified in the PDDL formulation and triggers replanning when a precondition is violated. Consequently, the PDDL constructs used in a planning task must be supported by both the selected planner and \plansys's mechanisms.

\plansys partially supports PDDL 2.1 up to level 3~\cite{fox2003pddl2} and lacks several PDDL features that this work requires. \Cref{sec:impl_plansys} lists them and details the extensions to \plansys that add this support. %

\subsection{Description Logic Ontologies}\label{sec:bg_dl}

This work uses \emph{description logic ontologies}~\cite{DBLP:books/daglib/0041477}. In this setting, an \emph{ontology} \Ont is a set of axioms defined over a common vocabulary \N. The vocabulary is partitioned as $\N = \NC \cup \NR \cup \NI$, where \NC, \NR, and \NI denote the sets of concept (class) names, role (property) names, and individual names, respectively.

There are two types of \emph{axioms} that form an ontology: (i) terminological box~(\emph{TBox}) axioms, which describe relations between concepts and (ii) assertional box~(\emph{ABox}) axioms that describe assertions. TBox axioms are of the form $C_1 \sqsubseteq C_2$ where $C_1, C_2$ are concepts. Such axioms describe a sub-concept relationship, i.e., that all elements in $C_1$ are also elements in $C_2$. ABox axioms are of the form $C(a)$ and $r(a,b)$, where $C$ is a concept, $a$ and $b$ are individual names and $r$ is a role name.

\emph{Concepts} are complex expressions built from the vocabulary \N. Depending on the logical fragment used, the concepts allowed vary.
In this work, the fragment \ALCI~(Attributive Language with Complement and Inverse roles) is sufficient. It builds concepts from concept names using negation, conjunction, disjunction, and existential and universal restrictions over roles and their inverses.

Additionally, an ontology can be extended with rules defined with the \emph{semantic web rule language~(SWRL)}~\cite{HorrocksPBTG2004}, which are an additional way to infer facts by providing a consequent that is inferred when the antecedent holds. A rule $(\antecedent, \consequent)$ is a pair of sets of atoms. Atoms are either of the form $C(x)$ or $r(x,y)$, where $C$ is a concept name, $r$ is a role name or a (numeric) comparison operator, and $x$ and $y$ are either variables, individual names or data values.

\subsection{TOMASys Ontology}\label{sec:bg_tomasys}

The Teleological and Ontological Metamodel for Autonomous Systems (\tomasys)~\cite{aguado2021functional,corbato2013model} is an ontology designed to capture the knowledge necessary for enabling architectural adaptation in robotic systems. It models the system's functionalities, the architectural alternatives that realize each functionality, the runtime quality requirements associated with these functionalities, and the expected quality output for each architectural variant.

\paragraph{Classes}
\tomasys' class names \NC are defined as follows:
\[
\NC = \left\lbrace \begin{array}{l}
	\FunctionTomasys, \FunctionDesign, \QAtype, \QAvalue,
	\Component, \Objective, \FunctionGrounding
\end{array}\right\rbrace
\]

An individual of the $\FunctionTomasys$ class represents a system functionality. A $\FunctionDesign$ individual represents an architectural variant that realizes a particular $\FunctionTomasys$, and a single functionality could have several designs that differ in the quality they provide. A $\Component$ individual represents a hardware or software element that a FunctionDesign requires. In SUAVE, there is a \emph{generate search path} functionality that is realized by three function designs that perform spiral search at different altitudes. The six AUV thrusters are represented as $\Component$ individuals, which are required by the function designs of the \emph{maintain motion} functionality.

A $\QAtype$ individual represents a type of quality attribute (QA)~\cite{wiki:sebok}, and a $\QAvalue$ individual represents a specific value of a $\QAtype$, which can be used to express quality requirements, the estimated quality output of function designs, and runtime measurements. In SUAVE, water visibility and search performance are $\QAtypes$. Each generate search path function design has a minimum required water visibility value and an estimated search performance. At runtime, the perception system reports the measured water visibility as a $\QAvalue$ individual. In SUAVE extended, battery level is a $\QAtype$ used to express the requirement that the measured battery level shall not go lower than $25\%$.

\tomasys also defines runtime elements. An $\Objective$ individual represents a runtime goal to have a $\FunctionTomasys$ realized, and a $\FunctionGrounding$ individual records which $\FunctionDesign$ currently meets that $\Objective$. Further relations link them to the measured QA values.

\paragraph{Relations}
In \tomasys, $\solvesF$ represents which $\FunctionDesign$ solves which $\FunctionTomasys$, $\requiredBy$ represents that a $\Component$ is required by a $\FunctionDesign$, $\isQAtype$ represents that a $\QAvalue$ is of a given $\QAtype$, and $\hasQAestimation$ represents the expected $\QAvalue$ of selecting a $\FunctionDesign$. In SUAVE, $\isQAtype$ relates a value such as \emph{low} to the \emph{water visibility} $\QAtype$, and $\hasQAestimation$ gives each spiral design its required \emph{water visibility} $\QAvalue$, e.g., \emph{low} for the lowest-altitude design.

In addition to TBox and ABox axioms, \tomasys contains SWRL rules that infer the system status at runtime, e.g., whether a function design is realizable given the status of its required components.

%% file: sections/model.tex
\section{\pddltomasys}\label{sec:tomasys_pddl}

This work proposes the \pddltomasys ontology to enable TACA solely through ontology-mediated planning. \pddltomasys builds upon the \tomasys ontology~\cite{aguado2021functional,corbato2013model}, but \tomasys cannot be reused for this purpose without modification. It offers no representation of robot actions, and no representation of the relation between an action and the functionalities it requires, i.e., the action's functional requirements. It was also designed under assumptions about the organization of planning and reasoning that do not hold for \acronym. %

The first limitation is that \tomasys gives the planner no way to determine which configuration must be active for a given action to be executable. In SUAVE, for example, the \texttt{search\_pipeline} action requires the \emph{generate search path} and \emph{maintain motion} functionalities. A suitable $\FunctionDesign$ must therefore remain grounded for each functionality throughout the action's execution. To address this limitation, \pddltomasys introduces an explicit representation of action functional requirements, defined below. \acronym combines these explicitly modeled functional requirements with \acronym's runtime precondition monitoring (see \Cref{sec:bg_plansys}). If a constraint violation renders a grounded $\FunctionDesign$ unrealizable, the resulting execution failure prompts the \acronym controller to request replanning. %

The second limitation concerns two assumptions under which \tomasys and the Metacontrol managing system were designed. First, it assumes a separation of responsibilities between task and configuration planning: an external component, without direct access to the knowledge base, is responsible for selecting the robot's actions (task planning), while an internal component within the managing subsystem, with direct access to the knowledge base, handles the architectural configuration (configuration planning). The second assumption is that a monotonic reasoner, driven by an additional procedural component, writes all inferred facts permanently into the knowledge base. In contrast, \acronym tightly integrates task and configuration planning within a single component that has direct access to the knowledge base, and it applies virtualized monotonic reasoning, in which inferred facts exist only transiently during the reasoning cycle and are not written into the knowledge base.

These differences call for a modeling strategy that diverges from \tomasys in three respects: (i) modeling which functionalities of the system must be managed at runtime, (ii) modeling the runtime state of the system, and (iii) defining the inference rules accordingly. Concretely, \pddltomasys removes the $\Objective$ and $\FunctionGrounding$ classes together with their runtime relations, redefines $\functionGrounding$ as an object property, and introduces the $\Action$ class and the $\requiresF$ relation to connect actions to the functionalities they need. The remainder of this section formalizes \pddltomasys using description logic.

\paragraph{Classes and relations} \pddltomasys's classes $\NC$ and relations $\NR = \NRobject \cup \NRdata$ are defined as follows:

\[
\NC = \left\lbrace \begin{array}{l}
	\Action, \FunctionTomasys, \FunctionDesign,
	\Component, \QAtype, \QAvalue
\end{array}\right\rbrace
\]

\[
\NRobject = \left\lbrace\begin{array}{l}
	\requiresF, \solvesF, \requiresC,
	\functionGrounding, \isQAtype,
        \hasQAestimation, \fdBetterUtility
\end{array}\right\rbrace
\]

\[
\NRdata = \left\lbrace\begin{array}{l}
	\fActive, \fdUtility, \fdRealisability,
        \cStatus, \qaHasValue
\end{array}\right\rbrace
\]

To represent the functional requirements of actions and support TACA, this work introduces the class $\Action$ and the object property $\requiresF$ into the ontology. The $\requiresF$ relation specifies the $\FunctionsTomasys$ necessary to perform each action, thereby informing the planner which functionalities must be managed and activated to enable the execution of a given action, e.g., $\requiresF(\aSearchPipelineOwl, \fGenerateSearchPath)$.

To capture which configuration (i.e., $\FunctionDesign$) solves a $\FunctionTomasys$ at runtime and whether it is active, the ontology includes the $\functionGrounding$ relationship\footnote{Formerly a class in \tomasys} that relates a $\FunctionTomasys$ to the $\FunctionDesign$ that solves it, e.g., $\functionGrounding(\fGenerateSearchPath, \fdSpiralHigh)$. Additionally, this work introduces a data property $\fActive$ to indicate whether a $\FunctionTomasys$ is currently activated, i.e., whether a $\functionGrounding$ exists between it and a $\FunctionDesign$ other than $\fdUnground$.

To support the inference of the most suitable $\FunctionDesign$ based on utility, this work introduces the $\fdBetterUtility$ relation, which relates a higher-utility $\FunctionDesign$ to one with lower utility, e.g., $\fdBetterUtility(\fdSpiralHigh, \fdSpiralMed)$. This relationship is inferred at runtime by the planner, enabling \acronym to reason about TACA solely through ontology-mediated planning.

To represent that a $\FunctionTomasys$ should be deactivated, the base ABox includes one additional individual, $\fdUnground$, asserted to be a $\FunctionDesign$ through $\FunctionDesign(\fdUnground)$. Grounding a $\FunctionTomasys$ with $\fdUnground$ denotes that no configuration is currently active for it. The relations $\solvesF(\fdUnground, \varF)$ that let $\fdUnground$ ground any $\FunctionTomasys$ $\varF$ are inferred by the SWRL rule $\FunctionTomasys(\varF) \wedge \FunctionDesign(\fdUnground) \implies \solvesF(\fdUnground, \varF)$ rather than asserted in the ABox. To deactivate a $\FunctionTomasys$ $\varF$, \acronym asserts $\functionGrounding(\varF, \fdUnground)$, e.g., $\functionGrounding(\fGenerateSearchPath, \fdUnground)$.

\paragraph{SWRL rules}
To infer the system status, the following SWRL rules are added to the ontology.\footnote{OWL data properties are binary, relating a class individual to a typed literal value. For example, \texttt{fd\_realisability(?fd, false)} asserts that the \texttt{fd\_realisability} of \texttt{?fd} is \texttt{false}.} %
They infer, in order, that a $\FunctionDesign$ $\varFD$ is not realizable when a required component fails, that a $\FunctionTomasys$ $\varF$ is active when it is grounded with a $\FunctionDesign$ other than $\fdUnground$, which of two $\FunctionDesigns$ solving the same $\FunctionTomasys$ has the better utility, and the utility of a $\FunctionDesign$ from its performance QA estimation. The last rule is generic and can be replaced by application-specific rules.
\begin{align*}
    &\FunctionDesign(\varFD) \wedge \Component(\varC) \wedge \requiresC(\varFD, \varC) \wedge {} \\
    &\quad \cStatus(\varC, \text{ERROR}) \implies \fdRealisability(\varFD, \text{false}) \tag*{(realizability)}\\
    &\FunctionTomasys(\varF) \wedge \FunctionDesign(\varFD) \wedge \solvesF(\varFD, \varF) \wedge \functionGrounding(\varF, \varFD) \wedge (\varFD \neq \fdUnground) \\
    &\quad \implies \fActive(\varF, \text{true}) \tag*{(activation)}\\
    &\FunctionDesign(\varFD) \wedge \FunctionDesign(\varFDtwo) \wedge \FunctionTomasys(\varF) \wedge \solvesF(\varFD, \varF) \wedge \solvesF(\varFDtwo, \varF) \wedge {} \\
    &\quad \fdUtility(\varFD, \varU) \wedge \fdUtility(\varFDtwo, \varUtwo) \wedge (\varU < \varUtwo) \implies \fdBetterUtility(\varFDtwo, \varFD) \tag*{(better utility)}\\
    &\FunctionTomasys(\varF) \wedge \FunctionDesign(\varFD) \wedge \solvesF(\varFD, \varF) \wedge \hasQAestimation(\varFD, \varQA) \wedge {} \\
    &\quad \isQAtype(\varQA, \text{performance}) \wedge \qaHasValue(\varQA, \varQAvalue) \implies \fdUtility(\varFD, \varQAvalue) \tag*{(utility)}
\end{align*}

%% file: sections/framework.tex
\section{Planning Method}\label{sec:method}

To enable ontology-mediated planning with \pddltomasys for solving TACA, this work introduces \acronym (see \Cref{fig:architecture}). In \acronym, the task planning problem is specified in PDDL, while the adaptation logic is modeled in an OWL ontology conforming to \pddltomasys. These two models are automatically combined into a single PDDL model, as described in \Cref{sec:owl_to_pddl}. The resulting PDDL model is transformed once again to include TACA concerns by converting the task actions into \textit{managed} actions. In the running example, $search\_pipeline$ (A1) becomes a managed action. At runtime, a planning system uses this final PDDL model for task and configuration planning, as well as for managing plan execution.
\Cref{sec:owl_to_pddl} explains the OWL to PDDL transformation, and \Cref{sec:plan_transformation_taca} details the final transformation that incorporates all TACA-related concerns.

\subsection{OWL to PDDL translation}\label{sec:owl_to_pddl}

The translation algorithm used in this work for encoding the axioms of the OWL ontology into the PDDL formulation follows the approach proposed by Borgwardt et al.~\cite{DBLP:conf/aaai/Borgwardt0KKNS22}, where the OWL axioms of the ontology are rewritten into derived predicates in PDDL (see also~\Cref{sec:related_ontology_planning}).

To add the OWL axioms to the PDDL domain, one has to translate and incorporate an ontology $\Ont$, with class names $\NC$, relation names $\NR$, and individual names $\NI$, into an existing planning formulation of the form $\pddlTask = \langle \pddlDomain, \pddlInstance \rangle$, with $\pddlDomain = \langle \pddlActions, \pddlDerRules \rangle$ and $\pddlInstance= \langle \pddlObjects, \pddlInitial, \pddlGoal \rangle$ over the set of predicates $\pddlPredicates$. As summarized in \Cref{tab:owl_to_pddl_mapping}, the translation extends the predicates to $\pddlPredicates' = \pddlPredicates \cup \NC \cup \NR$, the objects to $\pddlObjects' = \pddlObjects \cup \NI$, and the initial state to $\pddlInitial' = \pddlInitial \cup \pddlInitial_{\text{ABox}}$, where $\pddlInitial_{\text{ABox}}$ contains the grounded atoms obtained from the class and relation assertions in the ABox.

\begin{table}[tbp]
\centering
\caption{Mapping between OWL ontology and PDDL concepts following the approach of~\cite{DBLP:conf/aaai/Borgwardt0KKNS22}.}%
\label{tab:owl_to_pddl_mapping}
\begin{tabular}{@{}llll@{}}
\toprule
{\textbf{Concept}} & {\textbf{Concept}} & {\textbf{Example}} & {\textbf{Example}}  \\
\textbf{OWL Ontology} & \textbf{PDDL} & \textbf{OWL Ontology} & \textbf{PDDL} \\
\midrule
Individual & Object & \texttt{f1} & \texttt{(f1 - object)} \\
Class  & Unary predicate & $\FunctionTomasys$ & \texttt{(Function ?f)} \\ %
Class Assertion & Grounded unary predicate & $\FunctionTomasys(\texttt{f1})$ & \texttt{(Function f1)} \\
Relation  & Binary predicate & $\functionGrounding$ & \texttt{(functionGrounding ?f ?fd)} \\ %
Relation Assertion & Grounded binary predicate & $\functionGrounding(\texttt{f1}, \texttt{fd1})$ & \texttt{(functionGrounding f1 fd1)} \\
\bottomrule
\end{tabular}
\end{table}

Schema information encoded in the TBox and in SWRL-rules is translated into derivation rules. Sub-concept relations $C_1 \sqsubseteq C_2$ in the ontology are translated to derivation rules of the form $C_2(x) \leftarrow C_1(x)$. SWRL-rules $(\antecedent, \consequent)$ are translated to derivation rules
\[\consequent(\pddlActParam) \leftarrow \exists \pddlActParam_a. \antecedent(\pddlActParam),\]
where \pddlActParam contains variables occurring in the $\consequent$ and $\pddlActParam_a$ contains variables occurring only in $\antecedent$, see \Cref{fig:swrl_to_pddl} for an example. The new set of derivation rules is $\pddlDerRules'= \pddlDerRules \cup \pddlDerRules_{\text{TBox}}$ where $\pddlDerRules_{\text{TBox}}$ are the derivation rules generated from sub-class axioms and SWRL rules.

\begin{figure}[tbp]
\centering
\begin{minipage}{0.41\textwidth}
\[
\begin{aligned}
&\FunctionTomasys(\varF) \wedge \FunctionDesign(\fdUnground) \\
&\implies \solvesF(\fdUnground, \varF)
\end{aligned}
\]
\end{minipage}
\hfill
\begin{minipage}{0.57\textwidth}

\begin{lstlisting}[language=PDDL,numbers=none]
(:derived (solvesF ?fd_unground ?f)
  (and (= ?fd_unground fd_unground)
    (and (Function ?f)
      (FunctionDesign fd_unground))))
\end{lstlisting}

\end{minipage}

\vspace{-0.5cm}

\begin{minipage}[t]{0.41\textwidth}
    \captionof{lstlisting}{SWRL rule}%
    \label{lst:swrl_math}
\end{minipage}
\begin{minipage}[t]{0.57\textwidth}
    \captionof{lstlisting}{PDDL derived predicate}%
    \label{lst:pddl_rule}
\end{minipage}
\caption{Transformation of a SWRL rule into its equivalent PDDL derived predicate.}%
\label{fig:swrl_to_pddl}
\Description{Two side-by-side representations of the same rule. On the left, a SWRL implication
states that if f is a Function and fd\_unground is a FunctionDesign, then fd\_unground
solves f. On the right, a PDDL derived predicate for solvesF expresses the same
conditions and uses an equality test to bind its first parameter to the constant
fd\_unground.}
\end{figure}

The new planning task is then defined as $\Pi' = \langle \pddlDomain', \pddlInstance' \rangle$ over the set of predicates $\pddlPredicates'$,
with $\pddlDomain' = \langle \pddlActions, \pddlDerRules' \rangle$ and $\pddlInstance'= \langle \pddlObjects', \pddlInitial', \pddlGoal \rangle$.

\subsection{Planning task transformation for co-adaptation}\label{sec:plan_transformation_taca}

Finally, to support task and configuration planning, this work proposes transforming the planning task $\Pi' = \langle \pddlDomain', \pddlInstance' \rangle$, which was produced by the OWL to PDDL translation, into a new planning task $\Pi''$ that incorporates adaptation concerns. This includes the addition of reconfiguration actions and the modification of action preconditions to account for functional and quality requirements. Specifically, each action $a \in \pddlActions_{\text{adapt}} \subseteq \pddlActions$, where $\pddlActions_{\text{adapt}}$ is the set of actions subject to adaptation, is replaced with a managed action $a_m \in \pddlActions_m$, the original action $a$ is removed from $\pddlActions$, and a single pair of reconfiguration  actions $a_{r_1}, a_{r_2} \in \pddlActions_r$ is added to the domain.

The effect of this transformation is illustrated using the \texttt{search\_pipeline} action from the running example (Section~\ref{sec:running_example}), which encodes \emph{A1} and requires the \emph{generate search path} and \emph{maintain motion} functionalities.
Listing~\ref{lst:search_pipeline} shows its original form, in which the only precondition is that the robot has started and the sole effect is that the pipeline has been found.
Listing~\ref{lst:search_pipeline_managed} shows the managed version, in which the preconditions are extended with functional and quality requirements for both functionalities, ensuring that each is grounded to an active, non-failed $\FunctionDesign$ with the highest available utility.  For example, while water visibility is high the highest-utility function design for \emph{generate search path} is the high-altitude spiral design ($\fdSpiralHigh$). When visibility drops during \emph{A1} (uncertainty \Utwo), that design is no longer valid, so the managed precondition no longer holds and replanning inserts a reconfiguration action in the plan that re-grounds \emph{generate search path} to the medium-altitude spiral design ($\fdSpiralMed$), which then satisfies the $\bestFD$ requirement.

Additionally, the managed precondition requires that all functions not required by the action are inactive, preventing conflicts between components that serve different task actions. For example, the function used to follow the pipeline in \emph{A2}, $\fFollowPipeline$, must be inactive while \texttt{search\_pipeline} executes, as enforced by the trailing \texttt{(not (F\_active f\_follow\_pipeline \ldots{}))} condition in Listing~\ref{lst:search_pipeline_managed}. This constraint ensures that only the architectural configuration appropriate for the current task action is active at any time. %

\begin{figure}[tbp]
\begin{lstlisting}[language={PDDL},basicstyle=\sffamily\footnotesize, numbers=none,caption={Original \texttt{search\_pipeline} action},label={lst:search_pipeline}]
(:action search_pipeline
  :parameters (?p - pipeline ?r - robot)
  :precondition (and (robot_started ?r))
  :effect (and (pipeline_found ?p)))
\end{lstlisting}
\Description{PDDL listing of the original search\_pipeline action. Its parameters are a pipeline and
a robot. The precondition requires the robot to have started, and the effect marks the
pipeline as found.}
\end{figure}

\begin{figure}[tbp]
\begin{lstlisting}[language={PDDL},basicstyle=\sffamily\footnotesize,numbers=none,caption={Managed \texttt{search\_pipeline} action},label={lst:search_pipeline_managed}]
(:action search_pipeline
  :parameters (?p - pipeline ?r - robot)
  :precondition (and
    (robot_started ?r)
    ;; ?a = a_search_pipeline; ?f1 = f_generate_search_path, ?f2 = f_maintain_motion
    (exists (?a ?f1 ?f2 ?fd1 ?fd2)
      (and
        (Action ?a)
        (= ?a a_search_pipeline)
        (not (= ?f1 ?f2))
        (requiresF ?a ?f1)
        (F_active ?f1 true_boolean)
        (FunctionGrounding ?f1 ?fd1)
        (not (Fd_realisability ?fd1 false_boolean))
        (not
          (exists (?fd1_b)
            (and
              (not (= ?fd1 ?fd1_b))
              (SolvesF ?fd1_b ?f1)
              (not (Fd_realisability ?fd1_b false_boolean))
              (FdBetterUtility  ?fd1_b ?fd1))))
        ;; same conditions for ?f2 (f_maintain_motion)
      ))
    ;; every functionality not required by A1 must be inactive. Here, f_follow_pipeline required by A2
    (not (F_active f_follow_pipeline true_boolean)))
  :effect (and (pipeline_found ?p)))
\end{lstlisting}
\Description{PDDL listing of the managed search\_pipeline action. The original parameters and effect
are retained. Additional existential preconditions require two distinct functions for
the action, each active and grounded to a realizable FunctionDesign with no better
realizable alternative. The follow-pipeline function, which is not required by this
action, must be inactive.}
\end{figure}

Formally, the transformation of an action $a \in \pddlActions$ into a managed action
$a_m \in \pddlActions_m$ is defined as follows. Let $\pddlActName_a$ denote the action name,
$\pddlActParam_a$ the parameters of action $a$, $\pddlActPre_a$ its preconditions, and
$\pddlActAdd_a$ and $\pddlActDel_a$ the sets of atoms added to and removed from the state,
respectively, when $a$ is executed. The managed action is a tuple
$a_m = (\pddlActName_a, \pddlActParam_a, \pddlActPre_{a_m}, \pddlActAdd_a, \pddlActDel_a)$ with

\begin{align*}
    \pddlActPre_{a_m} =& \pddlActPre_a  ~ \cup \pddlActionReq ~ \cup
    \left\lbrace\begin{array}{l}\neg \fActive(\instanceF, \instanceTrue) \mid \\
    \forall \instanceF. \FunctionTomasys(\instanceF) \land \neg(\requiresF(a, \instanceF)) \end{array}\right\rbrace
\end{align*}

Where the action's requirements are expressed in terms of functional and quality requirements:

\begin{align*}
    \pddlActionReq =& \exists \instanceA, \instanceF_1, \ldots{},\instanceF_n,\instanceFD_1,\ldots{},\instanceFD_n .\left(\begin{array}{l} \Action(\instanceA) \land \instanceA = \pddlActName_a \land  \\
    \functionalReq(\instanceA, \instanceF_1, \instanceFD_1) \land \bestFD(\instanceF_1, \instanceFD_1) \land
    \\ \ldots{} \land \\
    \functionalReq(\instanceA, \instanceF_n, \instanceFD_n) \land \bestFD(\instanceF_n, \instanceFD_n) \land \\
    \bigwedge_{1 \leq i < j \leq n} \instanceF_i \neq \instanceF_j \\
 \end{array}\right)
\end{align*}

The final condition requires the $n$ functions to be pairwise distinct, so that a single function cannot satisfy several functional requirements. For \texttt{search\_pipeline}, $n = 2$, and $\instanceF_1$ and $\instanceF_2$ are the \emph{generate search path} and \emph{maintain motion} functions in \Cref{lst:search_pipeline_managed}.

A functional requirement is defined with the $\functionalReq(\instanceA, \instanceF, \instanceFD)$ predicate, which holds when a $\FunctionTomasys$ $\instanceF$ required by an $\Action$ $\instanceA$ has a grounded $\FunctionDesign$ solving it:
\begin{align*}
    \functionalReq(\instanceA, \instanceF, \instanceFD) =& \left\lbrace\begin{array}{l}
    \requiresF(\instanceA, \instanceF) \land \fActive(\instanceF, \instanceTrue) \land \functionGrounding(\instanceF, \instanceFD) \\
    \land \neg \fdRealisability(\instanceFD, \instanceFalse)
    \end{array}\right\rbrace
\end{align*}

Quality requirements can be defined with the predicate $\bestFD(\instanceF, \instanceFD)$, which holds if there does not exist another realizable $\FunctionDesign$ that solves function $\instanceF$ with better utility than $\instanceFD$. When specified as a quality requirement, the $\bestFD$ predicate constrains the planner to select a $\FunctionDesign$ with the highest utility among realizable alternatives for the corresponding function. This local requirement does not imply mission-level optimality, as \acronym computes a satisficing task and reconfiguration plan without optimizing overall mission utility. Different expressions can be used to define other types of quality requirements.

\begin{align*}
    \bestFD(\instanceF, \instanceFD) =& \nexists \instanceFD'.\left(\begin{array}{l}
         \solvesF(\instanceFD', \instanceF)
        \land \neg \fdRealisability(\instanceFD', \instanceFalse)
        \land \fdBetterUtility(\instanceFD', \instanceFD)
    \end{array} \right)\\
\end{align*}

The two additional reconfiguration actions, $a_{r_1}$ and $a_{r_2}$, are responsible for selecting architectural configurations by grounding $\FunctionDesigns$ to their corresponding $\FunctionsTomasys$. This is achieved by inserting predicates of the form $\functionGrounding(\instanceF, \instanceFD)$ into the state. Specifically, $a_{r_1}$ is applied when a $\functionGrounding(\instanceF, \instanceFD)$ already exists for function $\instanceF$, whereas $a_{r_2}$ is applied when the function is not grounded.

Both reconfiguration actions have similar parameters: a target $\FunctionTomasys$, a new $\FunctionDesign$ to be selected, and, in the case of $a_{r_1}$ only, the currently active $\FunctionDesign$. In both actions, the preconditions ensure that the new $\FunctionDesign$ is valid. The only difference between them is whether an existing $\FunctionDesign$ must be deactivated before activating a new one. Formally, the two reconfiguration actions are defined as $a_{r_1} = (\pddlActName_1, \pddlActParam_1, \pddlActPre_1, \pddlActAdd_1, \pddlActDel_1)$ and $a_{r_2} = (\pddlActName_2, \pddlActParam_2, \pddlActPre_2, \pddlActAdd_2, \pddlActDel_2)$, with:

\vspace{-0.5cm}
\begin{center}
\begin{minipage}[t]{0.38\linewidth}
\begin{align*}
    \pddlActName_1 =& \text{reconfigure1} \\
    \pddlActParam_1 =& (\instanceF, \instanceFD_\text{new}, \instanceFD_\text{old})\\
    \pddlActPre_1 =& \functionGrounding(\instanceF, \instanceFD_\text{old})\\
    & \land \FunctionDesign(\instanceFD_\text{new}) \\
    & \land \solvesF(\instanceFD_\text{new}, \instanceF) \\
    & \land (\instanceFD_\text{old} \neq \instanceFD_\text{new})\\
    & \land \lnot  \fdRealisability(\instanceFD_\text{new}, \instanceFalse) \\
    \pddlActAdd_1 =& \functionGrounding(\instanceF, \instanceFD_\text{new}) \\
    \pddlActDel_1 =& \functionGrounding(\instanceF, \instanceFD_\text{old})
\end{align*}
\end{minipage}%
\begin{minipage}[t]{0.38\linewidth}
\begin{align*}
    \pddlActName_2 =& \text{reconfigure2} \\
    \pddlActParam_2 =& (\instanceF, \instanceFD_\text{new})\\
    \pddlActPre_2 =& \nexists \instanceFD. \functionGrounding(\instanceF, \instanceFD) \\
     &\land \FunctionDesign(\instanceFD_\text{new}) \\
     &\land \solvesF(\instanceFD_\text{new}, \instanceF) \\
     & \land \lnot \fdRealisability(\instanceFD_\text{new}, \instanceFalse) \\
    \pddlActAdd_2 =& \functionGrounding(\instanceF, \instanceFD_\text{new}) \\
    \pddlActDel_2 =& \emptyset
\end{align*}
\end{minipage}
\end{center}

In the running example, the visibility drop described above is handled by \texttt{reconfigure1} with $\instanceF$ the \emph{generate search path} function, $\instanceFD_\text{old} = \fdSpiralHigh$, and $\instanceFD_\text{new} = \fdSpiralMed$.

The resulting planning task is defined as $\Pi'' = \langle \pddlDomain'', \pddlInstance' \rangle$ over the set of predicates $\pddlPredicates'$, where $\pddlDomain'' = \langle \pddlActions'', \pddlDerRules' \rangle$, and $\pddlInstance'= \langle \pddlObjects', \pddlInitial', \pddlGoal \rangle$ with $\pddlActions'' = (\pddlActions \setminus \pddlActions_{\text{adapt}}) \cup \pddlActions_m \cup \pddlActions_r$. That is, only the set of actions is modified.

An off-the-shelf PDDL planner solves $\Pi''$ directly and returns a combined task and configuration plan that interleaves reconfiguration and managed actions, for example $\pi = (\underline{a_{r_2}}, \underline{a_{m_1}}, \underline{a_{r_1}}, \ldots, \underline{a_{r_1}}, \underline{a_{m_n}}, \underline{a_{r_1}})$. \acronym executes this plan with \plansys, which re-evaluates action preconditions as runtime information such as measured $\QAvalue$ individuals updates the state, and replans when a precondition no longer holds (\Cref{sec:bg_plansys,sec:impl_plansys}). %

%% file: sections/implementation.tex
\section{Implementation}\label{sec:implementation}

\acronym is implemented as a ROS~2-based system to facilitate its use in robotic applications. It integrates \plansys~\cite{PlanSys2} for planning and plan execution, System Modes~\cite{nordmann2021system} for managing system reconfiguration, and a custom ROS~2 package that implements the OWL to PDDL translation.

\subsection{\owltopddl tool}
To apply the OWL to PDDL translation, this work provides a standalone Kotlin-based executable that performs the translation independently of ROS~2. Additionally, a ROS~2 package is provided to integrate the tool with the ROS~2 ecosystem. While the transformation approach builds on prior work that translates OWL axioms into PDDL rules~\cite{DBLP:conf/aaai/Borgwardt0KKNS22}, existing tools are not well-suited for robotic self-adaptation. In particular, the \pddltomasys ontology includes numerical values that require translation, a feature unsupported by current tools. Additionally, developing a new tool allowed for optimizations, such as specifying the PDDL types of ontology instances. These types are exported to the PDDL domain model to help reduce planning time. Both the OWL to PDDL tool and the associated ROS~2 package are publicly available.\footnote{\url{https://github.com/remaro-network/dlToPlanning}}

\subsection{Planning and plan execution}\label{sec:impl_plansys}

For planning and plan execution, this work leverages \plansys. However, since \plansys does not support all PDDL requirements imposed by the OWL to PDDL transformation step, this work extends \plansys to support existential preconditions (PDDL 1.1), derived predicates (PDDL 2.2), equality between non-numeric objects (e.g., $(\texttt{=}\; ?p\; \texttt{constant})$), negation of composite expressions (e.g., $(\texttt{not}\; (\texttt{=}\; ?a\; ?b))$), and the execution of non-durative actions.

This work uses the Fast Downward~\cite{helmert2006fast} planner for automated planning since it supports derived predicates and existential preconditions. To use Fast Downward with \plansys, this work implements a \plansys plugin for the Fast Downward planner, which is publicly available online\footnote{\url{https://github.com/kas-lab/downward_ros}}.

Adding support for derived predicates and existential preconditions in \plansys resulted in conceptual changes to the \emph{preconditions evaluation} and \emph{apply action effects} algorithms that are used in the plan execution process. Furthermore, it required the addition of a mechanism that is able to evaluate derived predicates at runtime to infer new predicates in a reasonable amount of time. The other extensions to \plansys aforementioned required only additional implementation.

The synthetic domain in \Cref{lst:example_derived} is used throughout this section to keep the examples compact. The derived predicates that \owltopddl generates from the \pddltomasys ontology (\Cref{sec:owl_to_pddl}) share its structure: they are defined over other derived predicates and use existential quantification and negation. Predicates such as \texttt{f\_active} and \texttt{fd\_realisability} are re-derived from base facts such as \texttt{functionGrounding} that the reconfiguration actions modify (\Cref{sec:plan_transformation_taca}).

\paragraph{Preconditions evaluation algorithm}

In \plansys, the \emph{preconditions evaluation algorithm} $\texttt{Evaluate}(\pddlActPre_a, \pddlState)$ determines whether the preconditions $\pddlActPre_a$ of action $a \in \pddlActions$ are satisfied in a given state $\pddlState$. This algorithm is used both during plan graph construction~\cite{pddl_plans_bt} to build a behavior tree for plan execution,
and at runtime by the resulting plan graph BT to continuously verify the feasibility of executing actions. The current evaluation algorithm assumes that there are no free variables in the preconditions, i.e., all parameters are grounded with objects $o \in \pddlObjects$ during the planning process.

This assumption no longer holds when existential preconditions are introduced. For example, consider the plan $\pi = (\underline{a}_1)$ as a valid solution to a \emph{problem instance} $\pddlInstance$ of the \emph{planning domain} shown in~\Cref{lst:example_derived}, where $name_{\underline{a}_1} = \texttt{action1}$ and $\pddlActParam_{\underline{a}_1} = (\texttt{instance1})$. In this case, the existential precondition introduces a free variable $?y$ that remains ungrounded at planning time.

Evaluating such preconditions requires checking whether there exists an object $o \in \pddlObjects$ such that the variable $?y$ can be unified with $o$, and the resulting grounded formula satisfies the logical condition defined in the precondition body. In the example, this corresponds to verifying whether there exists an object $o$ such that the atom $(\texttt{predicateAA}~\texttt{o})$ is contained in the current state $\pddlState$, e.g., $(\texttt{predicateAA}~\texttt{instance2}) \in \pddlState$.

With the addition of derived predicates, the evaluation algorithm is no longer limited to checking action preconditions, it must also evaluate the conditions of derived predicates, i.e., $\texttt{Evaluate}(C, \pddlState)$, where $C ::= \pddlActPre_a \mid \phi(\pddlActParam)$. In this setting, the evaluation of a precondition $c \in C$ cannot be performed by simply verifying whether it is satisfied in the current state $\pddlState$. Instead, the relevant derivation rules $r \in \pddlDerRules$ must be applied to the current state $\pddlState$ to determine whether the predicates $p \in c$ can be derived. For example, in the \emph{planning domain} shown in~\Cref{lst:example_derived}, given a state $s = \{(\texttt{predicateA}~\texttt{instance1}), (\texttt{predicateAA}~\texttt{instance2})\}$, the precondition of $\texttt{action1}$ is satisfied only if the derived predicate $\texttt{(inferredA ?x)}$ (lines 1–2) is successfully inferred. To support the evaluation of existential preconditions and derived predicates in \plansys, this work proposes the evaluation algorithm described in \Cref{S-alg:eval}.

\begin{figure}[tbp]
\centering
\begin{minipage}[t]{0.49\linewidth}
\begin{lstlisting}[language=PDDL, firstnumber=1, numbers=left, stepnumber=1, xleftmargin=2em]
(:derived (inferredA ?x)
  (predicateA ?x))
(:derived (inferredAA ?x)
  (inferredA ?x))
(:derived (inferredB ?x)
  (or (predicateB ?x)
      (inferredAB ?x)))
(:derived (inferredBB ?x)
  (inferredB ?x))
(:derived (inferredAB ?x)
  (or (inferredA ?x)
      (inferredB ?x)))
\end{lstlisting}
\end{minipage}
\hfill
\begin{minipage}[t]{0.49\linewidth}
\begin{lstlisting}[language=PDDL, firstnumber=13, numbers=left, stepnumber=1, xleftmargin=2em]
(:action action1
  :parameters (?x)
  :precondition (and (exists (?y)
    (and (predicateA ?x)
         (predicateAA ?y)
         (inferredA ?x))))
  :effect (and (predicateB ?x)))
(:action action2
  :parameters (?x)
  :precondition (and
    (inferredB ?x)
    (inferredAB ?x))
  :effect (and (predicateC ?x)))
\end{lstlisting}
\end{minipage}
\caption{PDDL example with derived predicates and existential preconditions.}%
\label{lst:example_derived}
\Description{Two-column PDDL example. The left column defines five derived predicates.
\texttt{inferredA} depends on \texttt{predicateA}. \texttt{inferredAA} depends on
\texttt{inferredA}. \texttt{inferredB} depends on \texttt{predicateB} or
\texttt{inferredAB}. \texttt{inferredBB} depends on \texttt{inferredB}. Finally,
\texttt{inferredAB} depends on \texttt{inferredA} or \texttt{inferredB}. The right
column defines two actions. Action1 requires \texttt{predicateA} and \texttt{inferredA}
for its parameter and the existence of an object satisfying \texttt{predicateAA}, then
adds \texttt{predicateB}. Action2 requires \texttt{inferredB} and \texttt{inferredAB},
then adds \texttt{predicateC}.}
\end{figure}

\paragraph{Derived predicates resolution}

The derived predicate resolution implemented in this work adheres to the definition of derived predicates introduced by Thiébaux et al.~\cite{THIEBAUX200538}, which extends the original PDDL~2.2~\cite{edelkamp2004pddl2} specification. This definition allows derived predicates to depend on other derived predicates, including recursive self-dependencies, and allows negated derived predicates in their preconditions, provided that such negation does not occur within a recursive cycle.

\begin{figure}[tbp]
    \centering
    \begin{subfigure}[t]{0.48\linewidth}
        \vspace{0pt} %
        \centering
        \resizebox{\linewidth}{!}{\input{figures/derived_graph_tikz}}
        \caption{Complete graph.}%
        \label{fig:derived_graph}
    \end{subfigure}
    \hfill
    \begin{subfigure}[t]{0.48\linewidth}
        \vspace{0pt} %
        \centering
        \resizebox{\linewidth}{!}{\input{figures/derived_graph_pruned_tikz}}
        \caption{Pruned graph for $\pi=(\texttt{action1}, \texttt{action2})$.}%
        \label{fig:pruned_graph}
    \end{subfigure}
    \caption{Derived predicates resolution graphs for the PDDL formulation in \Cref{lst:example_derived}: (a) the full dependency graph and (b) the pruned graph for the plan $\pi = (\texttt{action1}, \texttt{action2})$. The predicates $\texttt{predicateAA}$ and $\texttt{predicateC}$ are omitted to simplify the visualization, as they are not relevant to the graph structure due to having no child predicates.}%
    \label{fig:derived_graphs_combined}
\Description{Two directed dependency graphs, with arrows pointing from a predicate to predicates or
actions that depend on it. In the complete graph on the left, \texttt{predicateA} leads
to \texttt{inferredA}, which leads to \texttt{inferredAA} and \texttt{inferredAB}.
PredicateB leads to \texttt{inferredB}, which leads to \texttt{inferredAB} and
\texttt{inferredBB}. InferredAB also points back to \texttt{inferredB}, forming a cycle.
In the pruned graph on the right, \texttt{inferredAA} and \texttt{inferredBB} are
removed. \texttt{action1} depends on \texttt{inferredA}, and \texttt{action2} depends on
\texttt{inferredB} and \texttt{inferredAB}. The cycle is retained.}
\end{figure}
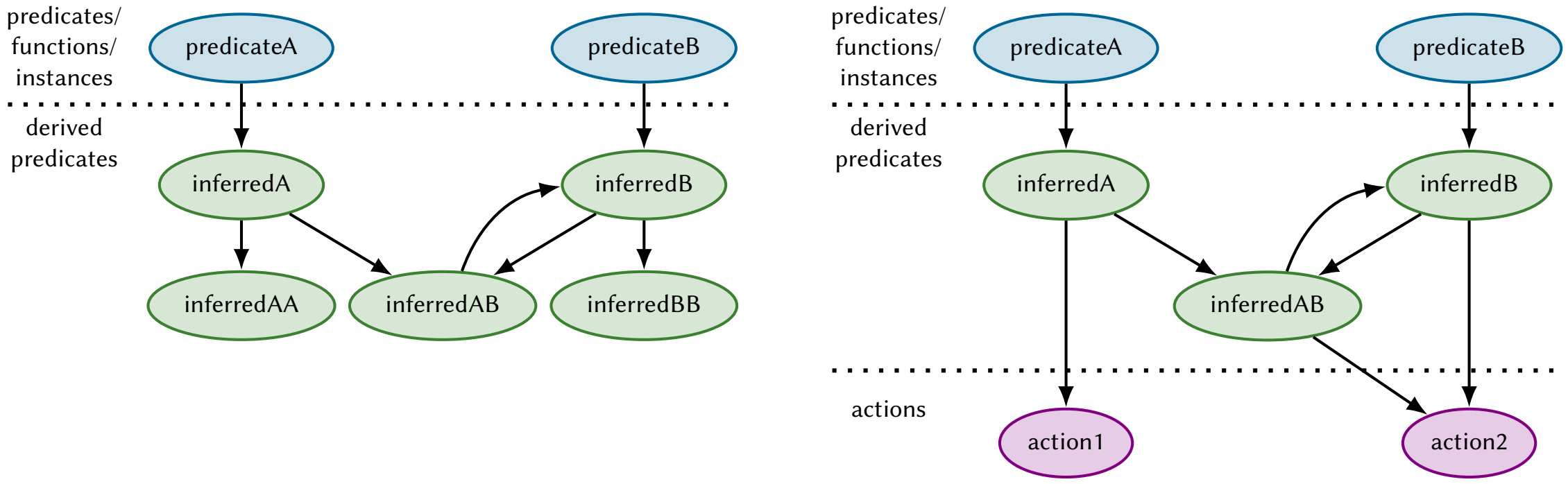

To address this problem, this work proposes representing the derived predicates from a PDDL formulation as a directed graph. For example, the PDDL formulation described in \Cref{lst:example_derived} can be represented as the graph depicted in \Cref{fig:derived_graph}. In this graph, nodes in the upper layer (depth 0) represent predicates, functions, or instances, while nodes in the lower layers (depth one or more) represent derived predicates. The edges indicate dependencies between the nodes, with arrows pointing from parent to child nodes. For example, the edge from $\texttt{predicateA}$ to $\texttt{inferredA}$ indicates that the latter depends on the former.

To solve derived predicates, this work introduces \Cref{S-alg:solve_derived}. It partitions the resolution graph into strongly connected components~(SCCs) using Tarjan's algorithm~\cite{doi:10.1137/0201010}, which separates acyclic from cyclic components. The acyclic components are resolved in topological order, while the cyclic components are resolved using a fixed-point iteration. Root nodes are considered solved by definition, as they correspond to non-derived predicates. For instance, in the graph shown in \Cref{fig:derived_graph}, the derived predicates $\texttt{inferredB}$ and $\texttt{inferredAB}$ form a cycle, whereas the other derived predicates are acyclic.

\paragraph{Apply action effects algorithm}

With the addition of derived predicates, applying an action's effects can invalidate previously derived predicates and make new ones derivable. Instead of recomputing all derived predicates, this work exploits the structure of the derived predicates graph to re-evaluate only the derived predicates that depend on the predicates modified by the effects (\Cref{S-alg:apply_new}). For example, in the PDDL domain described in \Cref{lst:example_derived}, applying the action $(\texttt{action1}~\texttt{instance1})$ adds $(\texttt{predicateB}~\texttt{instance1})$ to the state, which requires re-evaluating only the derived predicates $\texttt{inferredB}$, $\texttt{inferredAB}$, and $\texttt{inferredBB}$.

\paragraph{Derived graph pruning by action}

To optimize the evaluation of derived predicates during plan graph generation and runtime execution in \plansys, this work proposes appending the actions defined in the \emph{planning domain} to the derived graph and pruning it to retain only the predicates relevant to the current action plan. For example, given the \emph{planning domain} in \Cref{lst:example_derived} and the plan \(\pi = (\texttt{action1}, \texttt{action2})\), the derived graph shown in \Cref{fig:derived_graph} can be pruned to the graph in \Cref{fig:pruned_graph}, avoiding the unnecessary resolution of predicates \(\texttt{inferredAA}\) and \(\texttt{inferredBB}\).

%% file: figures/derived_graph_tikz.tex
\begin{tikzpicture}[font=\sffamily]
    
    \coordinate (origin) at (0,0);
    \draw (origin) node[bubbleBlue] (predicateA) {predicateA\strut};
    
    \draw ($(predicateA) + (5,0)$) node[bubbleBlue] (predicateB) {predicateB\strut};
    
    \draw ($(predicateA) + (0,-1.7)$) node[bubbleGreen] (inferredA) {inferredA\strut};
    \draw ($(predicateB) + (0,-1.7)$) node[bubbleGreen] (inferredB) {inferredB\strut};
    
    \draw ($(inferredA) + (0,-1.5)$) node[bubbleGreen] (inferredAA) {inferredAA\strut};
    \draw ($(inferredA) + (2.5,-1.5)$) node[bubbleGreen] (inferredAB) {inferredAB\strut};
    \draw ($(inferredB) + (0,-1.5)$) node[bubbleGreen] (inferredBB) {inferredBB\strut};
    
    \phantom{\draw ($(inferredAA) + (0,-1.7)$) node[bubblePurple] (action1) {action1\strut};} %
    
    \draw[arrow] (predicateA) -- (inferredA);
    \draw[arrow] (predicateB) -- (inferredB);
    
    \draw[arrow] (inferredA) -- (inferredAA);
    \draw[arrow] (inferredA) -- (inferredAB);
    \draw[arrow] (inferredB) -- (inferredAB);
    \draw[arrow] (inferredB) -- (inferredBB);
    
    \draw[arrow] (inferredAB) to [bend left] (inferredB);
    
    \draw ($(predicateA) + (-2.2,0)$) node (topLabel) {\begin{tabular}{c}
            predicates/\\
            functions/\\
            instances
    \end{tabular}};

    \draw ($(topLabel) + (0,-1.2)$) node (botLabel) {\begin{tabular}{c}
            derived\\
            predicates
    \end{tabular}};
    
    \draw[loosely dotted,
        draw=black, 
        line width=1.5pt] ($(topLabel) + (-0.7,-0.7)$) -- ++ (9,0);

\end{tikzpicture}

%% file: figures/derived_graph_pruned_tikz.tex
\begin{tikzpicture}[font=\sffamily]
    
    \coordinate (origin) at (0,0);
    \draw (origin) node[bubbleBlue] (predicateA) {predicateA\strut};
    
    \draw ($(predicateA) + (5,0)$) node[bubbleBlue] (predicateB) {predicateB\strut};
    
    \draw ($(predicateA) + (0,-1.7)$) node[bubbleGreen] (inferredA) {inferredA\strut};
    \draw ($(predicateB) + (0,-1.7)$) node[bubbleGreen] (inferredB) {inferredB\strut};
    
    \draw ($(inferredA) + (0,-1.5)$) node (inferredAA) {};
    \draw ($(inferredA) + (2.5,-1.5)$) node[bubbleGreen] (inferredAB) {inferredAB\strut};
    \draw ($(inferredB) + (0,-1.5)$) node (inferredBB) {};
    
    \draw ($(inferredAA) + (0,-1.7)$) node[bubblePurple] (action1) {action1\strut};
    \draw ($(inferredBB) + (0,-1.7)$) node[bubblePurple] (action2) {action2\strut};

    \draw[arrow] (predicateA) -- (inferredA);
    \draw[arrow] (predicateB) -- (inferredB);
    
    \draw[arrow] (inferredA) -- (action1);
    \draw[arrow] (inferredA) -- (inferredAB);
    \draw[arrow] (inferredB) -- (inferredAB);
    \draw[arrow] (inferredB) -- (action2);
    \draw[arrow] (inferredAB) -- (action2);
    
    \draw[arrow] (inferredAB) to [bend left] (inferredB);
    
    \draw ($(predicateA) + (-2.2,0)$) node (topLabel) {\begin{tabular}{c}
            predicates/\\
            functions/\\
            instances
    \end{tabular}};

    \draw ($(topLabel) + (0,-1.2)$) node (midLabel) {\begin{tabular}{c}
            derived\\
            predicates
    \end{tabular}};

    \draw ($(topLabel) + (0,-4.5)$) node (botLabel) {\begin{tabular}{c}
            actions
    \end{tabular}};
    
    \draw[loosely dotted,
        draw=black, 
        line width=1.5pt] ($(topLabel) + (-0.7,-0.7)$) -- ++ (9,0);
        
   \draw[loosely dotted,
   draw=black, 
   line width=1.5pt] ($(botLabel) + (-0.7,0.5)$) -- ++ (9,0);

\end{tikzpicture}

%% file: sections/evaluation.tex
\section{Evaluation}\label{sec:evaluation}

This work evaluates \acronym on two use cases from the literature: (i) the SUAVE exemplar~\cite{suave}, in which an AUV performs pipeline inspection, and (ii) an AGV scenario from Cámara et al.~\cite{CAMARASoftwareArchitectureTask2020}, in which the AGV navigates through an environment composed of corridors and rooms. The evaluation is organized around the following research questions:
\begin{description}
    \item[RQ1 (Feasibility).] Can \acronym be integrated into a ROS~2-based robotic system to perform task and architecture co-adaptation at runtime, and how does its mission performance compare with that of existing managing subsystems?
    \item[RQ2 (Performance efficiency).] How do \acronym's planning time and peak memory usage compare with those of the approach by Cámara et al.~\cite{CAMARASoftwareArchitectureTask2020}, and does \acronym produce plans of comparable quality?
    \item[RQ3 (Scalability).] How does \acronym's planning time evolve as the mission and adaptation spaces grow, and what are the trade-offs between optimal and satisficing planning?
\end{description}
RQ1 is addressed by the implementation of \acronym and its integration with \plansys~(\Cref{sec:implementation}), and by applying \acronym to the SUAVE exemplar and its TACA extension~(\Cref{sec:suave_exp}), which primarily serve as a feasibility evaluation. RQ2 is addressed with the AGV scenario~(\Cref{sec:nav_exp}). RQ3 is addressed by varying the AGV scenario along four dimensions, independently and in combination~(\Cref{sec:scale_exp}). For all scenarios, the size of the planning instance is reported in terms of the number of task and reconfiguration actions, derived predicates, objects, and initial facts, to characterize the complexity of \acronym's planning formulations~(\Cref{tab:pddl_instances}). The complete source code and experimental artifacts are publicly available online for the SUAVE experiments~\cite{gustavo_rezende_2026_23619542} and the navigation experiments~\cite{gustavo_rezende_2026_23647225}.

\begin{table}[tbp]
\centering
\caption{Complexity of the planning instances.
    \emph{Task actions}: PDDL actions encoding each task in the mission.
    \emph{Reconfig.\ actions}: PDDL actions encoding architecture reconfiguration.
    \emph{Objects}: domain constants and problem-instance objects.
    \emph{Init.\ facts}: ground atoms in the initial state.
    \emph{Derived predicates}: predicates generated by \owltopddl from the ontology.
    The scalability instances are reported at the maximum tested parameter value
    (\Sone: 900 nodes, \Stwo: 15 architectural configurations, \Sthree: 4 corridor types,
    \Sfour: 5 mission-level actions, and \Sfive: 90 nodes).}%
\label{tab:pddl_instances}
\begin{tabular}{lccccc}
\toprule
\textbf{Scenario} & \makecell{\textbf{Task}\\\textbf{actions}} & \makecell{\textbf{Reconfig.}\\\textbf{actions}} & \textbf{Objects} & \makecell{\textbf{Init.}\\\textbf{facts}} & \makecell{\textbf{Derived}\\\textbf{predicates}} \\ \midrule
SUAVE          & 3 & 2 & 44  & 115  & 54 \\
SUAVE extended & 4 & 2 & 49  & 125  & 55 \\ \midrule
Navigation     & 1 & 2 & 537 & 3184 & 53 \\
\Sone          & 1 & 2 & 947 & 8878 & 53 \\
\Stwo          & 1 & 2 & 150 & 1012 & 53 \\
\Sthree        & 1 & 2 & 144 & 1561 & 53 \\
\Sfour         & 5 & 2 & 153 & 988  & 53 \\
\Sfive         & 5 & 2 & 173 & 1635 & 53 \\
\bottomrule
\end{tabular}
\end{table}

\subsection{SUAVE exemplar}\label{sec:suave_exp}
\paragraph{Experimental design}
To assess \acronym's feasibility, mission performance, and reaction time, it is applied to the SUAVE exemplar~(see \Cref{sec:running_example}) and its TACA extension.

To support the benchmarking of different adaptation approaches, the SUAVE system is structured into a managing and a managed subsystem. The managing subsystem, responsible for the adaptation logic, can be replaced by any self-adaptive system that complies with the provided interfaces. SUAVE is packaged with three distinct managing subsystems: one that arbitrarily selects configurations, a behavior tree-based one, and  Metacontrol~\cite{bozhinoski2022mros}. Additionally, Silva et al.~\cite{rosa} made a ROSA implementation of SUAVE and its extended version available.

To apply \acronym to SUAVE, this work proposes a \pddltomasys model of the SUAVE exemplar based on the SUAVE \tomasys ontology. The main differences are the addition of new actions, utility rules, and a rule to handle uncertainty \Utwo~(water visibility). First, the actions \emph{search pipeline} and \emph{inspect pipeline}, as well as their functional requirements, are added to the SUAVE ontology. For the extended use case the \emph{recharge battery} action is also included. For example, $\requiresF$ relates \emph{search pipeline} to the functions \emph{maintain motion} and \emph{generate search path}. Then, rules are added to infer the $\FunctionsTomasys$ utility, for example, a rule that instantiates the generic (utility) rule of \Cref{sec:tomasys_pddl} for the \emph{generate search path} function. Finally, the following rule is defined to infer whether the \emph{generate search path} function designs are realizable, given the measured water visibility:
\begin{align*}
    &\FunctionDesign(?fd1) \wedge
    \hasQAestimation(?fd1, ?eqa) \wedge
    \isQAtype(?eqa, water\_visibility) \wedge \\
    &\qaHasValue(?eqa, ?eqav) \wedge
    \QAvalue(?mqa) \wedge
    sameAs(?mqa, obs\_water\_visibility) \wedge \\
    & \qaHasValue(?mqa, ?mqav) \wedge
    \isQAtype(?mqa, water\_visibility) \wedge
    lessThan(?mqav, ?eqav) \\
    &\implies \fdRealisability(?fd1, false)
\end{align*}

To exemplify the output of the model transformation process, \Cref{lst:search_pipeline_managed} shows the final PDDL formulation for the search pipeline action from the SUAVE use case. The SUAVE task actions in \Cref{tab:pddl_instances} include the non-managed action of starting the robot. The extended instance adds one extra PDDL action schema, recharge battery, and one derived predicate that encodes the battery level condition for uncertainty~\Uthree.

\paragraph{Experimental Setup}
To compare \acronym with existing adaptation approaches, it was evaluated on both the original and extended SUAVE use cases. To reduce the risk of bias from a single parameter set, as in~\cite{suave, rosa}, the evaluation used randomized values for the robot's initial position, the water visibility shift (i.e., its initial value), and the timing of thruster failures. Specifically, three experiments were conducted for each use case, each with a distinct parameter configuration as detailed in \Cref{tab:suave_params}.

\begin{table}[tbp]
\centering
\caption{Parameter configuration for the SUAVE experiments, where $U(a,b)$ is a random uniform distribution from $a$ to $b$. The initial position is relative to the Gazebo reference frame, where $x$ is fixed at $-17.0$ and $z$ at $-18.5$. The time limit for each execution is set to $250$ seconds.}%
\label{tab:suave_params}
\resizebox{.95\textwidth}{!}{%
\begin{tabular}{llllllll}
\toprule
\multirow{2}{*}{\textbf{Exp.}} &
  \textbf{Initial position} &
  \multicolumn{4}{l}{\textbf{Water visibility}} &
  \multicolumn{2}{l}{\textbf{Thruster failures}} \\
 &
  \textbf{Y} &
  \textbf{Min (m)} &
  \textbf{Max (m)} &
  \textbf{Period (s)} &
  \textbf{Shift (s)} &
  \textbf{Thruster} &
  \textbf{Time (s)} \\ \midrule
1 & $2.5 + U(-0.5, 0.5)$ & 2.5  & 2.5  & N/A & N/A      & 1 and 3 & $100 + U(-100, 100)$ \\
2 & $2.5 + U(-0.5, 0.5)$ & 1.25 & 3.75 & 120 & $U(0,120)$ & N/A     & N/A                \\
3 & $2.5 + U(-0.5, 0.5)$ & 1.25 & 3.75 & 120 & $U(0,120)$ & 1 and 3 & $100 + U(-100, 100)$ \\ \bottomrule
\end{tabular}%
}
\end{table}

The original use case was executed 30 times with no managing subsystem, as well as with each of the following managing subsystems: a random subsystem, a behavior-tree-based subsystem (BT), Metacontrol, ROSA, and \acronym. The extended scenario, which includes TACA, was also executed 30 times for the BT-based subsystem, ROSA, and \acronym. In both the original and the extended scenarios, Fast Downward was configured to use A* search with the blind heuristic for \acronym. To ensure that outliers were averaged out, the parameters were kept the same for each group of five runs (i.e., runs 1–5, 6–10, etc.) within the 30 runs. For each run, standard SUAVE metrics~\cite{suave} were collected: \emph{pipeline search time}, the time from mission start until the AUV locates the pipeline, and \emph{pipeline distance inspected}, the total length of pipeline inspected during the mission. A third metric, \emph{reaction time}~\cite{rosa}, was also collected: the time elapsed from a constraint violation until the managing system completes adaptation of the managed subsystem.

\paragraph{Results}
The experimental results for the SUAVE use case are presented in \Cref{tab:suave_results} and for the extended use case in \Cref{tab:suave_extended_results}. To verify whether there is a statistically significant difference between the results, the Wilcoxon signed-rank test was applied to all pairs of managing subsystems for both the \emph{pipeline search time} and \emph{pipeline distance inspected} metrics. The p-values for experiment 3 and extended experiment 3, which combine thruster failures and water visibility changes, are reported in \Cref{tab:wilcoxon}. The p-values for the remaining experiments are reported in \supplementname.

Results were paired by run index, and pairs in which either system did not find the pipeline were excluded from both comparisons. For each metric, the alternative hypothesis is that the managing system in the row achieves a lower \emph{pipeline search time} or, respectively, a greater \emph{pipeline distance inspected} than the one in the column.
p-values less than 0.05 indicate statistical significance in favor of the alternative hypothesis.

\input{tables/suave_result_table.tex}

\input{tables/suave_extended_result_table.tex}

\begin{table}[tbp]
\centering
\caption{Paired Wilcoxon signed-rank comparisons for experiment 3 and extended experiment 3. Each cell reports the unadjusted p-values for the comparison between the managing subsystem in the row and the one in the column for both the \emph{pipeline search time} and \emph{pipeline distance inspected} metrics. Cells with a green background indicate that the alternative hypothesis was favored, and cells with a blue background indicate that the null hypothesis cannot be rejected in favor of the alternative hypothesis. Values rounding to 0.000 are shown as $<0.001$. ``MC'' stands for Metacontrol.}%
\label{tab:wilcoxon}
\begin{subtable}{\textwidth}
\centering
\begingroup
  \renewenvironment{table}[1][]{}{}%
  \renewcommand{\caption}[2][]{}%
  \renewcommand{\label}[1]{}%
  \input{tables/wilcoxon_analysis/exp3_wilcoxon.tex}\endgroup
\caption{Experiment 3}\label{tab:wilcoxon-exp3}
\end{subtable}

\medskip
\begin{subtable}{0.78\textwidth}
\centering
\begingroup
  \renewenvironment{table}[1][]{}{}%
  \renewcommand{\caption}[2][]{}%
  \renewcommand{\label}[1]{}%
  \input{tables/wilcoxon_analysis/extended_exp3_wilcoxon.tex}\endgroup
\caption{Extended experiment 3}\label{tab:wilcoxon-extended-exp3}
\end{subtable}
\end{table}

\emph{SUAVE experiments:} Across the three SUAVE experiments (\Cref{tab:wilcoxon-exp3} and \Cref{S-tab:wilcoxon-exp1,S-tab:wilcoxon-exp2}), the BT managing subsystem achieved significantly lower \emph{pipeline search times} than all other configurations. Metacontrol achieved significantly lower search times than ROSA and \acronym in every experiment. \acronym achieved significantly lower search times than ROSA in experiment 2, while no statistically significant difference between them was detected in experiments 1 and 3. For \emph{pipeline distance inspected}, BT, Metacontrol, and ROSA each achieved significantly greater distances than \acronym in all three experiments. \acronym achieved significantly greater inspected distances than Random in every experiment. Compared with the configuration without a managing subsystem, \acronym inspected significantly less pipeline distance in experiment 1 and significantly more in experiments 2 and 3. %

\emph{SUAVE extended experiments:} Across the three extended SUAVE experiments (\Cref{tab:wilcoxon-extended-exp3} and \Cref{S-tab:wilcoxon-extended-exp1,S-tab:wilcoxon-extended-exp2}), the BT managing subsystem achieved significantly lower \emph{pipeline search times} than ROSA and \acronym, while no statistically significant difference in search time was detected between ROSA and \acronym. For \emph{pipeline distance inspected}, BT achieved significantly greater distances than both ROSA and \acronym, and ROSA achieved significantly greater distances than \acronym in every experiment.

\emph{Answer to RQ1:} These results support the feasibility of using \acronym for runtime task and architecture co-adaptation in SUAVE, but also reveal limitations in mission performance. BT, Metacontrol, and ROSA each achieved significantly better results than \acronym on at least one metric in every experiment in which they were evaluated. \acronym performed significantly better than Random on both metrics in all original experiments, and better than the configuration without a managing subsystem in experiments 2 and 3, but worse in experiment 1. BT's shorter mean reaction times (\Cref{tab:suave_results,tab:suave_extended_results}) may partly explain its advantage, although the experiments do not establish a causal relationship between reaction time and mission performance. %

\subsection{Navigation use case}\label{sec:nav_exp}
\paragraph{Experimental design}
To evaluate \acronym's performance efficiency, in terms of planning time and peak resident set size (RSS) memory usage, and its plan quality, it was applied to the navigation scenario from~\cite{CAMARASoftwareArchitectureTask2020}. Results were compared to those obtained from a replicated implementation of the algorithm proposed in~\cite{CAMARASoftwareArchitectureTask2020}, hereafter referred to as \CamaraApproach.

In the scenario, a robot must navigate from an initial position to a goal position within an environment composed of corridors and rooms. Some corridors might be dark or contain obstacles, reducing their safety. During navigation, the robot dynamically adapts both its architecture (i.e., the sensors and localization algorithms employed) and its task execution (i.e., the path followed) to circumvent failures or prioritize safety or energy consumption. For example, if the robot's planned path includes a dark corridor and its lamp fails during execution, it must reconfigure its architecture and adapt its task execution to avoid dark corridors.

\paragraph{Experimental setup}

To compare \acronym's time performance and memory usage with \CamaraApproach, the map from~\cite{CAMARASoftwareArchitectureTask2020} was discretized to ensure uniform edge lengths. This change was necessary because \acronym does not support numeric fluents and therefore cannot handle edges of varying sizes~(\Cref{sec:limitations}).
\acronym was then applied to perform task and configuration planning on the discretized map. For this comparison, Fast Downward was configured to use eager greedy search with the goal-count heuristic.

Additionally, a model generator was developed to produce PRISM models equivalent to those used in the final step of \CamaraApproach\footnote{The complete algorithm from~\cite{CAMARASoftwareArchitectureTask2020} could not be replicated.}, enabling policy generation for task planning and reconfiguration. A minor correction was introduced to the original PRISM model by adding an effect to the move actions, allowing the selection of distinct configurations for each corridor.  This work assumes that the correction aligns with the authors’ original intent, as the published model appears to contain a typographical error. For all experiments, PRISM was executed with the goal of maximizing the amount of battery remaining at the end of the mission. Both approaches were evaluated by applying them ten times to all possible combinations of initial and goal locations.

\begin{figure}[tbp]
    \centering
    \hfill
    \begin{subfigure}[t]{0.45\linewidth}
        \vspace{0pt} %
        \centering
        \input{navigation_experiment/plots/camara_map_time}
        \caption{Planning time comparison between \acronym and \CamaraApproach.}%
        \label{fig:time_comparison}
    \end{subfigure}
    \hfill
    \begin{subfigure}[t]{0.45\linewidth}
        \vspace{0pt} %
        \centering
        \input{navigation_experiment/plots/camara_map_memory}
        \caption{Peak RSS memory comparison between \acronym and \CamaraApproach.}%
        \label{fig:memory_comparison}
    \end{subfigure}
    \hfill
    \caption{Comparison between \acronym and \CamaraApproach. The number is the median value, the boxes show interquartile range (Q1--Q3), the whiskers reach until $\pm 1.5$~IQR, and the circles show a sample of the observations beyond the whiskers.}%
    \label{fig:camara_pddl_sas}
\Description{Two box-plot panels compare the replicated approach \CamaraApproach with \acronym. The
left panel shows planning time in seconds, with medians of 1.21 for the replicated
approach and 0.40 for \acronym. The right panel shows peak memory in MiB, with
medians of 461.09 and 67.39, respectively. The replicated approach has a wider spread
and larger high-value outliers in both panels.}
\end{figure}
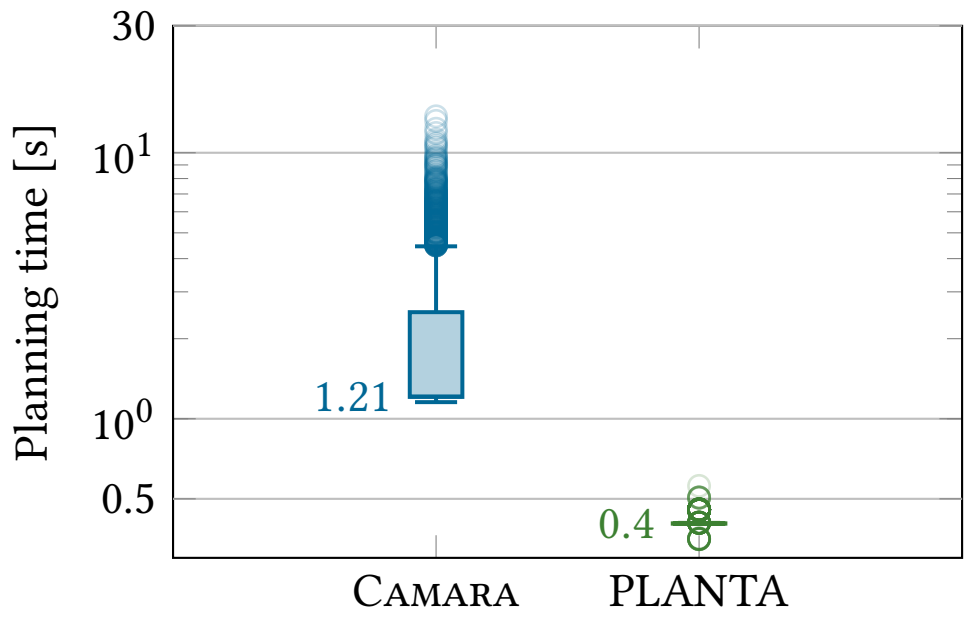
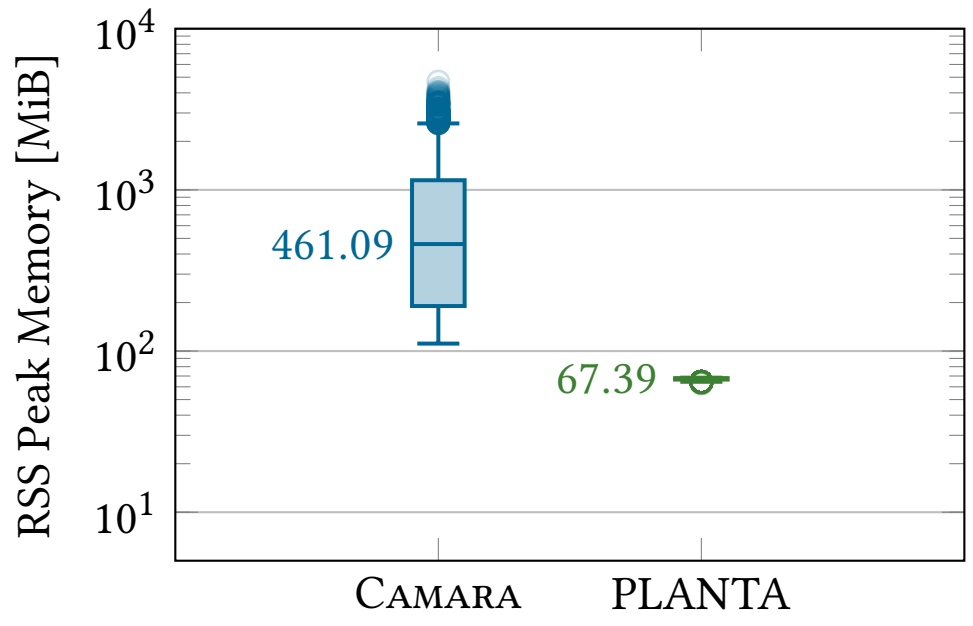

\paragraph{Results}
\Cref{fig:camara_pddl_sas} shows that, at the median, \acronym generates both task and configuration plans $3.00$ times faster than \CamaraApproach generates the corresponding policy from the PRISM models, with $6.84$ times smaller peak memory usage. In the scenarios with the worst planning time, \acronym (from location 16 to 25) takes $0.56$ seconds with $66.88$~MiB peak memory usage, compared to $13.71$ seconds with $4089.03$~MiB peak memory usage for \CamaraApproach (from location 52 to 24). Notably, the spread of execution times and memory consumption is much higher for \CamaraApproach, making the resources required by \acronym more predictable. These results should be interpreted as a comparison of the computational cost of two different ways of addressing TACA rather than as a direct comparison of approaches that pursue the same objective. \CamaraApproach seeks an optimal task and configuration policy under probabilistic uncertainty, whereas \acronym favors satisficing planning over mission-level optimality, planning from the currently observed state and handling changes during execution through replanning. %

Despite the difference in planning time and memory usage, the two approaches produce plans with almost identical path length and remaining battery. The distributions of distance traveled and of the numbers of move and reconfiguration actions nearly coincide. For example, the interquartile range of distance traveled is $36.81$ to $82.70$ for \acronym and $36.62$ to $82.66$ for \CamaraApproach, with the same median of $59.89$. The median remaining battery differs by only $1.02\%$ ($25219.50$ for \acronym against $25480.00$ for \CamaraApproach). Both approaches reconfigure more than once in a minority of the scenarios, $11.85\%$ for \acronym and $15.52\%$ for \CamaraApproach, and up to four times (for example, between locations 11 and 17 for \acronym and between locations 5 and 18 for \CamaraApproach). %

The reported computational performance difference likely underestimates \acronym's advantage. In~\cite{CAMARASoftwareArchitectureTask2020}, the final step, the sole component that could be reproduced for \CamaraApproach, accounts for only 39.34\% of the total execution time. Furthermore, \CamaraApproach generates and verifies PRISM models for the ten best paths between the initial and goal locations, selecting the best policy according to a predefined utility function. In contrast, the experiment described in this paper generates and verifies a PRISM model only for the single best path.

\emph{Answer to RQ2:} In the evaluated AGV scenario, \acronym addresses TACA at a lower computational cost than the replicated final step of \CamaraApproach, in both planning time and peak memory usage, while producing plans with almost identical path length and remaining battery.

\subsection{Scalability}\label{sec:scale_exp}

\paragraph{Experimental design} To assess the scalability of the proposed approach with respect to mission and adaptation spaces, in terms of planning time and memory usage, five additional experiments were conducted using variations of the previous navigation scenario. Specifically, four experiments were designed to independently evaluate the effect of varying the map size, the number of architectural configurations, the number of quality requirements, and the number of task actions. A fifth experiment was designed to stress-test the approach by varying the map size while fixing all other parameters at their maximum values.

Furthermore, to isolate the effect of the adaptation logic on planning time and memory usage, each of the scalability experiments described above was also conducted using only the mission logic (i.e., without reconfiguration). In addition, to compare optimal and satisficing planning, all scalability experiments, both with and without the adaptation logic, were repeated with Fast Downward configured to use A* search with the blind heuristic and eager greedy search with the goal-count heuristic, where an optimal plan corresponds to the shortest plan since all actions have unit cost.

\paragraph{Experimental setup}

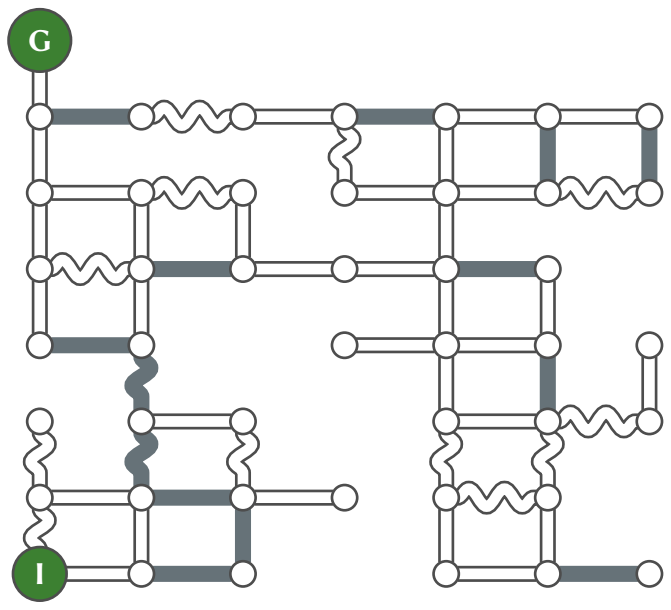
\begin{figure}[tbp]
    \centering
    \resizebox{0.30\textwidth}{!}{
        \input{figures/grid-map}
    }
    \caption{Grid map example with $45$ retained nodes out of $50$ generated nodes.
    White edges represent normal corridors, filled edges represent dark corridors and curved edges represent unsafe corridors. The initial location is marked with {\sffamily\bfseries I} and the goal location is marked with {\sffamily\bfseries G}.
    }%
    \label{fig:grid_map_example}
\Description{Grid-like navigation graph with 45 locations, consisting of a seven-by-seven grid with five locations removed and an additional goal location. Several connections are omitted. The
initial location I is at the lower left, and the goal G is above the upper-left grid
location. Edges connect neighboring locations: white straight edges denote normal
corridors, filled edges denote dark corridors, and curved edges denote unsafe corridors.
Filled curved edges denote corridors that are both dark and unsafe.}
\end{figure}

For each scalability experiment, a map generator produces grid-like maps with the required number of nodes, edges, and corridor types. The generator skips $10\%$ of nodes and $15\%$ of the edges are removed while preserving graph connectivity. In the baseline maps, $25\%$ of the corridors are marked as dark and $25\%$ are marked as unsafe. The generated map is then translated into a PDDL problem file, leveraging the Unified Planning~\cite{unified_planning_2025_micheli} library, with node $0$ as the robot's initial position and the highest-indexed node as the goal waypoint. \Cref{fig:grid_map_example} shows an example map. Each experiment is repeated $10$ times for each search strategy~(i.e., A* and eager greedy) and adaptation mode~(with and without adaptation), while the \emph{planning time}, \emph{peak memory usage}, and the \emph{number of actions} in the plan are collected.

\Cref{tab:scalability_params} lists the parameter ranges of the five scalability experiments. Architectural configurations are combinations of sensors and localization algorithms, and \Stwo adds further combinations\footnote{The additional architectural configurations are included to increase the size of the architectural configuration space, not to model realistic selection criteria or quality-attribute trade-offs.}. \Sthree models additional quality requirements as two new corridor types: outdoor corridors ($20\%$) and dusty corridors ($20\%$). Outdoor corridors require sensors that are robust to sunlight, for example, lidars, whereas dusty corridors require sensors that are robust to floating particles, for example, RGB cameras. \Sfour adds the mission-level PDDL actions inspect, deliver, recharge, and report. The map-size upper bound in \Sfive is lower than in \Sone because \Sfive combines the largest architectural configuration space, the largest set of corridor requirements, and the largest mission-action set, which substantially increases the planning space.

\begin{table}[tbp]
  \centering
  \caption{Parameter configurations for the five scalability experiments~(S1--S5).
    Map nodes are generated nodes. The map generator skips $10\%$ of them, so the maximum of $1000$ generated nodes in S1 and $100$ in S5 corresponds to $900$ and $90$ retained nodes, respectively.
    For S4 and S5, the number of architectural configurations grows with the number of action types: each additional mission-level PDDL action introduces two further configurations.
  }%
  \label{tab:scalability_params}
  \begin{tabular}{llcccc}
    \toprule
    & \textbf{Experiment} &
    \makecell{\textbf{Map nodes}} &
    \makecell{\textbf{Arch.}\\\textbf{configs}} &
    \makecell{\textbf{Corridor}\\\textbf{types}} &
    \makecell{\textbf{Actions}} \\
    \midrule
    \Sone & Map size               & $10$--$1000$, step $10$ & $6$                 & $2$     & $1$ \\
    \Stwo & Arch.\ configurations  & $100$                   & $2,4,6,8,10,12,15$  & $2$     & $1$ \\
    \Sthree & Quality requirements   & $100$                   & $6$                 & $2,3,4$ & $1$ \\
    \Sfour & Mission-level actions           & $100$                   & $6$--$14$           & $2$     & $1,2,3,4,5$ \\
    \Sfive & Combined (stress test) & $10$--$100$, step $10$  & $23$                & $4$     & $5$ \\
    \bottomrule
  \end{tabular}
\end{table}

Experiments \Stwo, \Sthree, and \Sfour use maps with $100$ nodes, so that changes in planning time and memory usage can be interpreted primarily with respect to the dimension varied in each experiment rather than map size, which is evaluated separately in \Sone. \Sfive combines the action-driven configuration growth of \Sfour with the additional architectural configurations introduced in \Stwo, yielding $23$ configurations. The configurations added in \Sfour are intentionally not ranked by utility, which avoids introducing a separate utility-optimization factor into the mission-action scalability experiment. \Cref{tab:pddl_instances} reports the resulting complexity of each experiment's planning instance at its maximum tested parameter value.

\paragraph{Results}

\begin{figure}[tbp]
    \centering
    \pgfplotsset{every axis post/.append style={legend style={inner sep=1pt,row sep=-2pt}}}%
    \begin{subfigure}[t]{0.32\linewidth}
        \centering
        {%
        \pgfplotsset{every axis post/.append style={xtick={0,300,600,900}}}%
        \input{navigation_experiment/plots/s1_time_plot}}
        \caption{\Sone: map size.}%
        \label{plot:s1_planning_time}
    \end{subfigure}\hfill
    \begin{subfigure}[t]{0.32\linewidth}
        \centering
        \input{navigation_experiment/plots/s4_time_plot}
        \caption{\Sfour: mission-level actions.}%
        \label{plot:s4_planning_time}
    \end{subfigure}\hfill
    \begin{subfigure}[t]{0.32\linewidth}
        \centering
        \input{navigation_experiment/plots/s5_time_plot}
        \caption{\Sfive: combined stress test.}%
        \label{plot:s5_planning_time}
    \end{subfigure}
    \caption{Planning time in the scalability experiments \Sone, \Sfour, and \Sfive. The shaded area shows the standard deviation. EG stands for the eager greedy search.}%
    \label{fig:scalability_time}
\Description{Three line-plot panels show mean planning time in seconds. In S1, as map size increases
from 9 to 900 retained nodes, A* and eager greedy search have similar planning times, reaching
0.92 and 0.95 seconds, respectively, and the no-adaptation baseline reaches about 0.15 seconds. In
S4, as the number of mission-level action schemas increases from 1 to 5, A* planning time rises
sharply to 15.45 seconds, while eager greedy search reaches 1.05 seconds and the no-adaptation
baseline remains near 0.10 seconds. In S5, the combined stress test, as map size increases from 9
to 90 retained nodes, A* planning time rises to 188.96 seconds, whereas eager greedy search reaches
13.73 seconds. Shaded bands indicate standard deviations.}
\end{figure}
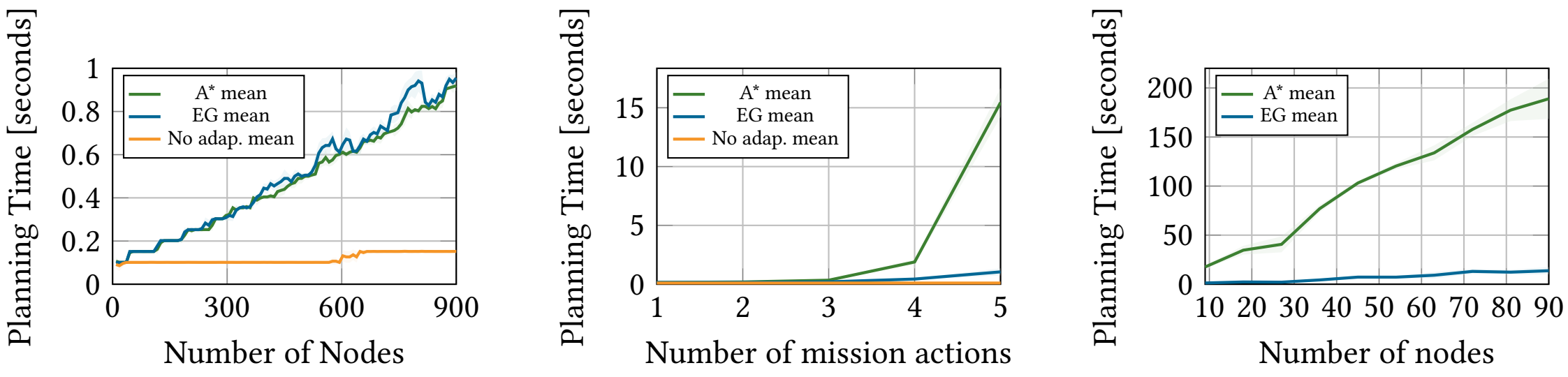

Across the five scalability experiments, the number of mission-level PDDL actions is the dominant driver of planning time, with map size as a secondary driver. Architectural configurations (\Stwo) and quality requirements (\Sthree) have a small effect on planning time. Map size (\Sone) produces a moderate increase for both search strategies, whereas increasing the number of mission-level PDDL actions (\Sfour) leads to a much larger increase, with A* search requiring more than an order of magnitude more time than eager greedy search while both strategies produce plans of similar length. This disparity becomes even more pronounced when all dimensions are combined (\Sfive), where A* search becomes impractical for online TACA, while eager greedy search remains manageable. Peak memory usage grows mainly with map size and, for A* search, with the number of mission-level PDDL actions. It ranges from $37.69$~MiB to $104.66$~MiB in \Sone and reaches $308.43$~MiB for A* search and $90.32$~MiB for eager greedy search in the largest \Sfive instances. \Cref{tab:scalability_summary} summarizes the results, and \Cref{fig:scalability_time} shows the planning-time trends of \Sone, \Sfour, and \Sfive. The plots for all five experiments, including plan length, are given in \Crefrange{S-fig:s1_results}{S-fig:s5_results} of \supplementname.

\input{navigation_experiment/tables/scalability_summary}

\emph{\Sone: Map size.}
\Cref{plot:s1_planning_time} shows that the planning time increases with map size. For the largest maps, with $900$ retained nodes, computing a plan takes less than one second with both search strategies (\Cref{tab:scalability_summary}). This is adequate for non-safety-critical robotic applications in which replanning is not required at high frequency, such as the experiment's navigation application. The A* and eager greedy curves are similar, indicating little difference between optimal and satisficing planning in this experiment. A* and eager greedy also result in plans with the same length (\Cref{tab:scalability_summary}).

\emph{\Stwo and \Sthree: Architectural configurations and quality requirements.}
Planning time remains below $0.31$ seconds over the evaluated range of architectural configurations and corridor types (\Cref{tab:scalability_summary}). A* and eager greedy return plans with the same length.

\emph{\Sfour: Mission-level PDDL actions.}
\Cref{plot:s4_planning_time} shows that planning time increases with the number of mission-level PDDL actions, especially for A* search. With $5$ mission-level actions, A* requires $15.45$ seconds on average, whereas eager greedy search requires $1.05$ seconds. \Cref{tab:scalability_summary} also shows that this reduction in planning time does not come with a substantial increase in plan length. This latency matters for online TACA\@: in the navigation scenario, for example, a lamp failure before traversing a dark corridor requires replanning so that the resulting task plan and architectural configuration satisfy the corridor requirements. Therefore, eager greedy search is the more practical choice for online TACA in this scenario.

\emph{\Sfive: Combined stress test.}
When the dimensions varied individually across \Sone--\Sfour are combined, planning time increases substantially compared to varying map size alone. In \Sfive, planning time reaches $188.96$ seconds on average for A* search and $13.73$ seconds on average for eager greedy search at just $90$ retained nodes, as shown in \Cref{plot:s5_planning_time}. This represents an increase of $205.62$ times in A* planning time relative to \Sone, for a map that is one-tenth the size~(90 vs 900 retained nodes), confirming that combining all dimensions simultaneously amplifies planning time well beyond what any single-dimension experiment produces. Despite the difference in planning time between both search strategies, plan length has a modest difference (\Cref{tab:scalability_summary}). For online TACA under the full-complexity configuration, eager greedy search is the more tractable option. Nevertheless, its planning time is still high for applications requiring frequent replanning.

\emph{No adaptation.}
\Cref{plot:s1_planning_time} and \Cref{plot:s4_planning_time}\footnote{The no-adaptation curve in both plots reports A* search. The corresponding eager greedy curve is omitted because it follows the same trend and does not add relevant information.} demonstrate that planning with only the mission logic~(i.e., task planning) is substantially faster than planning with the adaptation logic included~(i.e., task and architecture planning). For the single-action mission used in \Sone--\Sthree, the no-adaptation baseline reaches approximately $0.15$ seconds at $900$ retained nodes. For the mission used in \Sfour--\Sfive, the no-adaptation baseline reaches approximately $0.10$ seconds with five mission-level actions. These results indicate that the adaptation logic is the primary source of the planning-time increase observed with adaptation.

\emph{Answer to RQ3:} The scalability experiments \Sone--\Sthree show that planning time remains manageable as map size, the number of architectural configurations, and the number of quality requirements increase individually, for both satisficing and optimal planning. \Sfour shows that increasing the number of mission-level PDDL actions has a substantially greater impact on planning time under optimal planning than under satisficing planning. Since both approaches produce plans of similar length, satisficing planning is the more suitable choice for this scenario. \Sfive shows that, when all dimensions are combined, planning time reaches impractical levels under optimal planning and high, though potentially acceptable, levels under satisficing planning for non-safety-critical applications with low replanning frequency. Across all experiments, \Cref{tab:pddl_instances} shows that raw instance size, in objects and initial facts, is driven primarily by map size rather than by architectural or mission complexity, indicating that the planning-time increase observed in \Sfour and \Sfive stems from the added branching factor of additional PDDL action schemas rather than from a larger planning instance.

%% file: tables/suave_result_table.tex
\begin{table}[tb]
\centering
\caption{SUAVE results. The Random managing subsystem found the pipeline in 21/30 runs (70.00\%) for Experiment 1, 22/30 runs (73.33\%) for Experiment 2, and 22/30 runs (73.33\%) for Experiment 3. All other managing subsystems found the pipeline in every run. Search time and inspected distance report the mean and sample standard deviation only for runs that found the pipeline. \Uone denotes thruster failure and \Utwo water visibility. Reaction times are arithmetic means of recorded events per uncertainty across all retained runs. N/A denotes an inapplicable metric or insufficient observations. The initial visibility adjustment in Experiment 1 is excluded.}\label{tab:suave_results}
\resizebox{.95\textwidth}{!}{%
\begin{tabular}{llcccccc}
\toprule
\multirow{2}{*}{\textbf{Exp.}} & \multirow{2}{*}{\textbf{Managing subsystem}} & \multicolumn{2}{c}{\textbf{Search time (s)}} & \multicolumn{2}{c}{\textbf{Distance inspected (m)}} & \multicolumn{2}{c}{\textbf{Mean reaction time (s)}} \\
 & & \textbf{Mean} & \textbf{Std.\ dev.} & \textbf{Mean} & \textbf{Std.\ dev.} & \textbf{\Uone} & \textbf{\Utwo} \\
\midrule
\multirow{6}{*}{1} & None & 58.77 & 26.07 & 49.50 & 12.56 & N/A & N/A \\
 & Random & 126.71 & 52.90 & 11.65 & 10.08 & N/A & N/A \\
 & BT & 59.63 & 27.46 & 58.17 & 8.43 & 0.07 & N/A \\
 & Metacontrol & 59.20 & 25.50 & 58.92 & 8.63 & 1.11 & N/A \\
 & ROSA & 65.33 & 27.48 & 57.76 & 9.06 & 0.75 & N/A \\
 & \textbf{\acronym} & \textbf{65.07} & \textbf{33.95} & \textbf{43.56} & \textbf{8.77} & \textbf{0.88} & \textbf{N/A} \\
\midrule
\multirow{6}{*}{2} & None & 152.13 & 30.32 & 27.31 & 11.23 & N/A & N/A \\
 & Random & 113.32 & 48.30 & 10.91 & 8.15 & N/A & N/A \\
 & BT & 83.83 & 30.96 & 50.42 & 10.25 & N/A & 0.13 \\
 & Metacontrol & 85.53 & 31.51 & 50.22 & 10.37 & N/A & 1.37 \\
 & ROSA & 89.03 & 31.19 & 49.69 & 10.46 & N/A & 1.52 \\
 & \textbf{\acronym} & \textbf{87.83} & \textbf{31.41} & \textbf{44.45} & \textbf{10.11} & \textbf{N/A} & \textbf{1.78} \\
\midrule
\multirow{6}{*}{3} & None & 148.97 & 33.14 & 25.90 & 9.47 & N/A & N/A \\
 & Random & 119.50 & 52.35 & 8.76 & 7.31 & N/A & N/A \\
 & BT & 84.43 & 30.94 & 49.82 & 10.30 & 0.07 & 0.12 \\
 & Metacontrol & 86.77 & 31.48 & 49.30 & 10.08 & 1.22 & 1.52 \\
 & ROSA & 89.57 & 31.71 & 49.31 & 10.53 & 0.78 & 1.56 \\
 & \textbf{\acronym} & \textbf{88.73} & \textbf{32.07} & \textbf{43.58} & \textbf{9.45} & \textbf{1.45} & \textbf{1.86} \\
\bottomrule
\end{tabular}%
}
\end{table}

%% file: tables/suave_extended_result_table.tex
\begin{table}[tb]
\centering
\caption{SUAVE extended results. All managing subsystems found the pipeline in every run. Search time and inspected distance report the mean and sample standard deviation only for runs that found the pipeline. \Uone denotes thruster failure and \Utwo water visibility, with \Uthree denoting battery level. Reaction times are arithmetic means of recorded events per uncertainty across all retained runs. N/A denotes an inapplicable metric or insufficient observations. The initial visibility adjustment in Experiment 1 is excluded.}\label{tab:suave_extended_results}
\resizebox{.95\textwidth}{!}{%
\begin{tabular}{llccccccc}
\toprule
\multirow{2}{*}{\textbf{Exp.}} & \multirow{2}{*}{\textbf{Managing subsystem}} & \multicolumn{2}{c}{\textbf{Search time (s)}} & \multicolumn{2}{c}{\textbf{Distance inspected (m)}} & \multicolumn{3}{c}{\textbf{Mean reaction time (s)}} \\
 & & \textbf{Mean} & \textbf{Std.\ dev.} & \textbf{Mean} & \textbf{Std.\ dev.} & \textbf{\Uone} & \textbf{\Utwo} & \textbf{\Uthree} \\
\midrule
\multirow{3}{*}{1} & BT & 64.30 & 34.37 & 23.40 & 8.17 & 0.09 & N/A & 1.09 \\
 & ROSA & 61.10 & 24.85 & 22.41 & 6.95 & 0.81 & N/A & 1.95 \\
 & \textbf{\acronym} & \textbf{71.47} & \textbf{37.23} & \textbf{17.47} & \textbf{8.79} & \textbf{0.89} & \textbf{N/A} & \textbf{1.44} \\
\midrule
\multirow{3}{*}{2} & BT & 83.93 & 30.88 & 16.74 & 9.09 & N/A & 0.12 & 1.08 \\
 & ROSA & 88.70 & 31.50 & 15.40 & 8.96 & N/A & 1.57 & 2.03 \\
 & \textbf{\acronym} & \textbf{88.20} & \textbf{31.86} & \textbf{14.72} & \textbf{8.44} & \textbf{N/A} & \textbf{2.37} & \textbf{2.25} \\
\midrule
\multirow{3}{*}{3} & BT & 89.07 & 40.14 & 16.25 & 9.07 & 0.07 & 0.12 & 1.07 \\
 & ROSA & 88.10 & 31.71 & 15.48 & 8.96 & 0.76 & 1.67 & 1.96 \\
 & \textbf{\acronym} & \textbf{89.30} & \textbf{32.06} & \textbf{14.29} & \textbf{8.47} & \textbf{1.57} & \textbf{2.40} & \textbf{2.31} \\
\bottomrule
\end{tabular}%
}
\end{table}

%% file: tables/wilcoxon_analysis/exp3_wilcoxon.tex
\begin{table}[tb]
\centering
\caption{Experiment 3: paired Wilcoxon signed-rank comparisons. Each cell reports the unadjusted p-values for the comparison between the managing subsystem in the row and the one in the column for both the \emph{pipeline search time} and \emph{pipeline distance inspected} metrics. Cells with a green background indicate that the alternative hypothesis was favored, and cells with a blue background indicate that the null hypothesis cannot be rejected in favor of the alternative hypothesis. Values rounding to 0.000 are shown as $<0.001$. ``MC'' stands for Metacontrol.}%
\label{tab:wilcoxon-exp3}
\renewcommand{\arraystretch}{1.2} %
\resizebox{\textwidth}{!}{%
\begin{tabular}{p{1.65cm}*{6}{p{1.3cm}}*{6}{p{1.3cm}}}
\toprule
\multirow{2}{1.65cm}{\raggedright \textbf{Managing subsystem}} & \multicolumn{6}{c}{\textbf{Pipeline search time}} & \multicolumn{6}{c}{\textbf{Pipeline distance inspected}} \\ \cline{2-13}
 & \textbf{None} & \textbf{Random} & \textbf{BT} & \textbf{MC} & \textbf{ROSA} & \textbf{PLANTA} & \textbf{None} & \textbf{Random} & \textbf{BT} & \textbf{MC} & \textbf{ROSA} & \textbf{PLANTA} \\
\midrule
\textbf{None} & - & \cellcolor{pvalue_blue}0.975 & \cellcolor{pvalue_blue}1.000 & \cellcolor{pvalue_blue}1.000 & \cellcolor{pvalue_blue}1.000 & \cellcolor{pvalue_blue}1.000 & - & \cellcolor{pvalue_green}$<0.001$ & \cellcolor{pvalue_blue}1.000 & \cellcolor{pvalue_blue}1.000 & \cellcolor{pvalue_blue}1.000 & \cellcolor{pvalue_blue}1.000 \\
\textbf{Random} & \cellcolor{pvalue_green}0.025 & - & \cellcolor{pvalue_blue}0.984 & \cellcolor{pvalue_blue}0.954 & \cellcolor{pvalue_blue}0.942 & \cellcolor{pvalue_blue}0.963 & \cellcolor{pvalue_blue}1.000 & - & \cellcolor{pvalue_blue}1.000 & \cellcolor{pvalue_blue}1.000 & \cellcolor{pvalue_blue}1.000 & \cellcolor{pvalue_blue}1.000 \\
\textbf{BT} & \cellcolor{pvalue_green}$<0.001$ & \cellcolor{pvalue_green}0.016 & - & \cellcolor{pvalue_green}$<0.001$ & \cellcolor{pvalue_green}$<0.001$ & \cellcolor{pvalue_green}$<0.001$ & \cellcolor{pvalue_green}$<0.001$ & \cellcolor{pvalue_green}$<0.001$ & - & \cellcolor{pvalue_blue}0.093 & \cellcolor{pvalue_blue}0.084 & \cellcolor{pvalue_green}$<0.001$ \\
\textbf{MC} & \cellcolor{pvalue_green}$<0.001$ & \cellcolor{pvalue_green}0.046 & \cellcolor{pvalue_blue}1.000 & - & \cellcolor{pvalue_green}$<0.001$ & \cellcolor{pvalue_green}0.001 & \cellcolor{pvalue_green}$<0.001$ & \cellcolor{pvalue_green}$<0.001$ & \cellcolor{pvalue_blue}0.907 & - & \cellcolor{pvalue_blue}0.457 & \cellcolor{pvalue_green}$<0.001$ \\
\textbf{ROSA} & \cellcolor{pvalue_green}$<0.001$ & \cellcolor{pvalue_blue}0.058 & \cellcolor{pvalue_blue}1.000 & \cellcolor{pvalue_blue}1.000 & - & \cellcolor{pvalue_blue}0.922 & \cellcolor{pvalue_green}$<0.001$ & \cellcolor{pvalue_green}$<0.001$ & \cellcolor{pvalue_blue}0.916 & \cellcolor{pvalue_blue}0.543 & - & \cellcolor{pvalue_green}$<0.001$ \\
\textbf{PLANTA} & \cellcolor{pvalue_green}$<0.001$ & \cellcolor{pvalue_green}0.037 & \cellcolor{pvalue_blue}1.000 & \cellcolor{pvalue_blue}0.999 & \cellcolor{pvalue_blue}0.078 & - & \cellcolor{pvalue_green}$<0.001$ & \cellcolor{pvalue_green}$<0.001$ & \cellcolor{pvalue_blue}1.000 & \cellcolor{pvalue_blue}1.000 & \cellcolor{pvalue_blue}1.000 & - \\
\bottomrule
\end{tabular}
}
\end{table}

%% file: tables/wilcoxon_analysis/extended_exp3_wilcoxon.tex
\begin{table}[tb]
\centering
\caption{Extended experiment 3: paired Wilcoxon signed-rank comparisons. Each cell reports the unadjusted p-values for the comparison between the managing subsystem in the row and the one in the column for both the \emph{pipeline search time} and \emph{pipeline distance inspected} metrics. Cells with a green background indicate that the alternative hypothesis was favored, and cells with a blue background indicate that the null hypothesis cannot be rejected in favor of the alternative hypothesis. Values rounding to 0.000 are shown as $<0.001$.}%
\label{tab:wilcoxon-extended-exp3}
\renewcommand{\arraystretch}{1.2} %
\resizebox{0.7\textwidth}{!}{%
\begin{tabular}{p{1.65cm}*{3}{p{1.3cm}}*{3}{p{1.3cm}}}
\toprule
\multirow{2}{1.65cm}{\raggedright \textbf{Managing subsystem}} & \multicolumn{3}{c}{\textbf{Pipeline search time}} & \multicolumn{3}{c}{\textbf{Pipeline distance inspected}} \\ \cline{2-7}
 & \textbf{BT} & \textbf{ROSA} & \textbf{PLANTA} & \textbf{BT} & \textbf{ROSA} & \textbf{PLANTA} \\
\midrule
\textbf{BT} & - & \cellcolor{pvalue_green}$<0.001$ & \cellcolor{pvalue_green}$<0.001$ & - & \cellcolor{pvalue_green}$<0.001$ & \cellcolor{pvalue_green}$<0.001$ \\
\textbf{ROSA} & \cellcolor{pvalue_blue}1.000 & - & \cellcolor{pvalue_blue}0.621 & \cellcolor{pvalue_blue}1.000 & - & \cellcolor{pvalue_green}$<0.001$ \\
\textbf{PLANTA} & \cellcolor{pvalue_blue}1.000 & \cellcolor{pvalue_blue}0.379 & - & \cellcolor{pvalue_blue}1.000 & \cellcolor{pvalue_blue}1.000 & - \\
\bottomrule
\end{tabular}
}
\end{table}

%% file: navigation_experiment/plots/camara_map_time.tex
\input{navigation_experiment/data/camara_map_prism_experiment_time_boxstats.tex}%
\input{navigation_experiment/data/camara_map_pddl_experiment_time_boxstats.tex}%
\begin{tikzpicture}[baseline=(paperAxis.north)]
\begin{axis}[name=paperAxis,
width=\textwidth,
height=0.75\textwidth,
boxplot/draw direction=y,
xmin=0,
xmax=3,
xtick={1,2},
xticklabels={\CamaraApproach,\paperEagerLabel},
ylabel={Planning time [s]},
ymajorgrids=true,
ymode=log,
ymin=0.3,
ymax=30,
ytick={1,10},
extra y ticks={0.5,30},
extra y tick labels={0.5,30}
]
\addplot+[boxplot prepared={lower whisker=\camaraPrismTimeLW,lower quartile=\camaraPrismTimeLQ,median=\camaraPrismTimeMED,upper quartile=\camaraPrismTimeUQ,upper whisker=\camaraPrismTimeUW,draw position=1,box extend=0.2},draw=\blueColor,fill=\blueColor!30,line width=0.75pt,mark=o,mark options={fill=\blueColor,draw=\blueColor,opacity=0.2,line width=0.5pt}] table[y=y,col sep=comma] {navigation_experiment/data/camara_map_prism_experiment_time_outliers.csv}
node[left,text=\blueColor] at (boxplot box cs: \camaraPrismTimeMED,0) {\pgfmathprintnumber{\camaraPrismTimeMED}};
\addplot+[boxplot prepared={lower whisker=\camaraPddlTimeLW,lower quartile=\camaraPddlTimeLQ,median=\camaraPddlTimeMED,upper quartile=\camaraPddlTimeUQ,upper whisker=\camaraPddlTimeUW,draw position=2,box extend=0.2},draw=\greenColor,fill=\greenColor!30,line width=0.75pt,mark=o,mark options={fill=\greenColor,draw=\greenColor,opacity=0.2,line width=0.5pt}] table[y=y,col sep=comma] {navigation_experiment/data/camara_map_pddl_experiment_time_outliers.csv}
node[left,text=\greenColor] at (boxplot box cs: \camaraPddlTimeMED,0) {\pgfmathprintnumber{\camaraPddlTimeMED}};
\end{axis}
\end{tikzpicture}

%% file: navigation_experiment/plots/camara_map_memory.tex
\input{navigation_experiment/data/camara_map_prism_experiment_peak_memory_boxstats.tex}%
\input{navigation_experiment/data/camara_map_pddl_experiment_peak_memory_boxstats.tex}%
\begin{tikzpicture}[baseline=(paperAxis.north)]
\begin{axis}[name=paperAxis,
width=\textwidth,
height=0.75\textwidth,
boxplot/draw direction=y,
xmin=0,
xmax=3,
xtick={1,2},
xticklabels={\CamaraApproach,\paperEagerLabel},
ylabel={RSS Peak Memory [MiB]},
ymajorgrids=true,
ymode=log,
ymin=5,
ymax=10000,
ytick={10,100,1000,10000}
]
\addplot+[boxplot prepared={lower whisker=\camaraPrismMemLW,lower quartile=\camaraPrismMemLQ,median=\camaraPrismMemMED,upper quartile=\camaraPrismMemUQ,upper whisker=\camaraPrismMemUW,draw position=1,box extend=0.2},draw=\blueColor,fill=\blueColor!30,line width=0.75pt,mark=o,mark options={fill=\blueColor,draw=\blueColor,opacity=0.2,line width=0.5pt}] table[y=y,col sep=comma] {navigation_experiment/data/camara_map_prism_experiment_peak_memory_outliers.csv}
node[left,text=\blueColor] at (boxplot box cs: \camaraPrismMemMED,0) {\pgfmathprintnumber{\camaraPrismMemMED}};
\addplot+[boxplot prepared={lower whisker=\camaraPddlMemLW,lower quartile=\camaraPddlMemLQ,median=\camaraPddlMemMED,upper quartile=\camaraPddlMemUQ,upper whisker=\camaraPddlMemUW,draw position=2,box extend=0.2},draw=\greenColor,fill=\greenColor!30,line width=0.75pt,mark=o,mark options={fill=\greenColor,draw=\greenColor,opacity=0.2,line width=0.5pt}] table[y=y,col sep=comma] {navigation_experiment/data/camara_map_pddl_experiment_peak_memory_outliers.csv}
node[left,text=\greenColor] at (boxplot box cs: \camaraPddlMemMED,0) {\pgfmathprintnumber{\camaraPddlMemMED}};
\end{axis}
\end{tikzpicture}

%% file: figures/grid-map.tex
\begin{tikzpicture}[font=\sffamily\bfseries]
    
    \foreach \i in {0,1,2,3,4,5,6} {
        \foreach \j in {0,1,2,3,4,5,6} {
            \node at (\i, -0.75*\j) (node-\i-\j) {};
            \node[gridNode] at (\i, -0.75*\j)  {};
        }
    }
    
    \node[goal] at (0, 0.75) (goal) {G};
    
    \node[start] at (node-0-6) (start) {I};

    \node[hideNode] at (node-2-3) {};
    \node[hideNode] at (node-3-4) {};
    \node[hideNode] at (node-3-6) {};
    \node[hideNode] at (node-6-2) {};
    \node[hideNode] at (node-6-5) {};

    \begin{pgfonlayer}{background}
        \draw[corridor] (goal.center) -- (node-0-0);
        
        \draw[corridor] (node-0-0) -- (node-0-1);
        \draw[corridor] (node-0-1) -- (node-0-2);
        \draw[corridor] (node-0-2) -- (node-0-3);
        \draw[corridorUnsafe] (node-0-4) -- (node-0-5);
        \draw[corridorUnsafe] (node-0-5) -- (node-0-6);
        
        \draw[corridor] (node-1-1) -- (node-1-2);
        \draw[corridor] (node-1-2) -- (node-1-3);
        \draw[corridorDarkUnsafe] (node-1-3) -- (node-1-4);
        \draw[corridorDarkUnsafe] (node-1-4) -- (node-1-5);
        \draw[corridor] (node-1-5) -- (node-1-6);
        
        \draw[corridor] (node-2-1) -- (node-2-2);
        \draw[corridorUnsafe] (node-2-4) -- (node-2-5);
        \draw[corridorDark] (node-2-5) -- (node-2-6);
        
        \draw[corridorUnsafe] (node-3-0) -- (node-3-1);
        
        \draw[corridor] (node-4-0) -- (node-4-1);
        \draw[corridor] (node-4-1) -- (node-4-2);
        \draw[corridor] (node-4-2) -- (node-4-3);
        \draw[corridor] (node-4-3) -- (node-4-4);
        \draw[corridorUnsafe] (node-4-4) -- (node-4-5);
        \draw[corridor] (node-4-5) -- (node-4-6);
        
        \draw[corridorDark] (node-5-0) -- (node-5-1);
        \draw[corridor] (node-5-2) -- (node-5-3);
        \draw[corridorDark] (node-5-3) -- (node-5-4);
        \draw[corridorUnsafe] (node-5-4) -- (node-5-5);
        \draw[corridor] (node-5-5) -- (node-5-6);
        
        \draw[corridorDark] (node-6-0) -- (node-6-1);
        \draw[corridor] (node-6-3) -- (node-6-4);
        
        \draw[corridorDark] (node-0-0) -- (node-1-0);
        \draw[corridorUnsafe] (node-1-0) -- (node-2-0);
        \draw[corridor] (node-2-0) -- (node-3-0);
        \draw[corridorDark] (node-3-0) -- (node-4-0);
        \draw[corridor] (node-4-0) -- (node-5-0);
        \draw[corridor] (node-5-0) -- (node-6-0);
        
        \draw[corridor] (node-0-1) -- (node-1-1);
        \draw[corridorUnsafe] (node-1-1) -- (node-2-1);
        \draw[corridor] (node-3-1) -- (node-4-1);
        \draw[corridor] (node-4-1) -- (node-5-1);
        \draw[corridorUnsafe] (node-5-1) -- (node-6-1);
        
        \draw[corridorUnsafe] (node-0-2) -- (node-1-2);
        \draw[corridorDark] (node-1-2) -- (node-2-2);
        \draw[corridor] (node-2-2) -- (node-3-2);
        \draw[corridor] (node-3-2) -- (node-4-2);
        \draw[corridorDark] (node-4-2) -- (node-5-2);
        
        \draw[corridorDark] (node-0-3) -- (node-1-3);
        \draw[corridor] (node-3-3) -- (node-4-3);
        \draw[corridor] (node-4-3) -- (node-5-3);
        
        \draw[corridor] (node-1-4) -- (node-2-4);
        \draw[corridor] (node-4-4) -- (node-5-4);
        \draw[corridorUnsafe] (node-5-4) -- (node-6-4);
        
        \draw[corridor] (node-0-5) -- (node-1-5);
        \draw[corridorDark] (node-1-5) -- (node-2-5);
        \draw[corridor] (node-2-5) -- (node-3-5);
        \draw[corridorUnsafe] (node-4-5) -- (node-5-5);
        
        \draw[corridor] (node-0-6) -- (node-1-6);
        \draw[corridorDark] (node-1-6) -- (node-2-6);
        \draw[corridor] (node-4-6) -- (node-5-6);
        \draw[corridorDark] (node-5-6) -- (node-6-6);
   \end{pgfonlayer}
    
\end{tikzpicture}

%% file: navigation_experiment/tables/scalability_summary.tex
\begin{table}[tb]
\centering
\caption{Scalability experiment results at the maximum tested parameter value for each experiment (\Sone: 900 nodes, \Stwo: 15 architectural configurations, \Sthree: 4 corridor types, \Sfour: 5 mission-level actions, and \Sfive: 90 nodes).}\label{tab:scalability_summary}
\resizebox{.75\textwidth}{!}{%
\begin{tabular}{lcccccccc}
\toprule
\multirow{3}{*}{\textbf{Exp.}} & \multicolumn{4}{c}{\textbf{A*}} & \multicolumn{4}{c}{\textbf{Eager greedy}} \\
\cmidrule(lr){2-5}\cmidrule(lr){6-9}
 & \multicolumn{2}{c}{\textbf{Planning time (s)}} & \multicolumn{2}{c}{\textbf{Plan length}} & \multicolumn{2}{c}{\textbf{Planning time (s)}} & \multicolumn{2}{c}{\textbf{Plan length}} \\
\cmidrule(lr){2-3}\cmidrule(lr){4-5}\cmidrule(lr){6-7}\cmidrule(lr){8-9}
 & \textbf{Mean} & \textbf{Std.} & \textbf{Mean} & \textbf{Std.} & \textbf{Mean} & \textbf{Std.} & \textbf{Mean} & \textbf{Std.} \\
\midrule
\Sone & 0.92 & 0.02 & 47.20 & 7.74 & 0.95 & 0.03 & 47.20 & 7.74 \\
\Stwo & 0.30 & 0.00 & 20.70 & 1.42 & 0.30 & 0.02 & 20.70 & 1.42 \\
\Sthree & 0.20 & 0.00 & 20.30 & 2.26 & 0.20 & 0.00 & 20.30 & 2.26 \\
\Sfour & 15.45 & 1.24 & 46.00 & 5.52 & 1.05 & 0.30 & 46.80 & 6.21 \\
\Sfive & 188.96 & 20.43 & 48.10 & 3.73 & 13.73 & 3.66 & 51.60 & 5.64 \\
\bottomrule
\end{tabular}%
}
\end{table}

%% file: navigation_experiment/plots/s1_time_plot.tex
\begin{tikzpicture}[baseline=(paperAxis.north)]
\begin{axis}[name=paperAxis,
width=\textwidth,
height=0.75\textwidth,
xlabel={Number of Nodes},
ylabel={Planning Time [seconds]},
xmin=0,
xmax=900,
ymin=0,
ymax=1,
xtick={0,200,400,600,800,900},
grid=major,
legend pos=north west,
legend style={font=\footnotesize,nodes={scale=0.8,transform shape}},
legend image post style={scale=0.5}
]
\addplot[name path=upper0,draw=none,forget plot] table[x=nodes,y=mean_time_plus_std,col sep=comma] {navigation_experiment/data/s1_a_star_experiment_processed.csv};
\addplot[name path=lower0,draw=none,forget plot] table[x=nodes,y=mean_time_minus_std,col sep=comma] {navigation_experiment/data/s1_a_star_experiment_processed.csv};
\addplot[fill=\greenColor!20,fill opacity=0.3,draw=none,forget plot] fill between[of=upper0 and lower0];
\addplot[thick,color=\greenColor] table[x=nodes,y=mean_time,col sep=comma] {navigation_experiment/data/s1_a_star_experiment_processed.csv};
\addlegendentry{A* mean}
\addplot[name path=upper1,draw=none,forget plot] table[x=nodes,y=mean_time_plus_std,col sep=comma] {navigation_experiment/data/s1_eager_experiment_processed.csv};
\addplot[name path=lower1,draw=none,forget plot] table[x=nodes,y=mean_time_minus_std,col sep=comma] {navigation_experiment/data/s1_eager_experiment_processed.csv};
\addplot[fill=\blueColor!20,fill opacity=0.3,draw=none,forget plot] fill between[of=upper1 and lower1];
\addplot[thick,color=\blueColor] table[x=nodes,y=mean_time,col sep=comma] {navigation_experiment/data/s1_eager_experiment_processed.csv};
\addlegendentry{EG mean}
\addplot[name path=upper2,draw=none,forget plot] table[x=nodes,y=mean_time_plus_std,col sep=comma] {navigation_experiment/data/s1_a_star_no_sas_experiment_processed.csv};
\addplot[name path=lower2,draw=none,forget plot] table[x=nodes,y=mean_time_minus_std,col sep=comma] {navigation_experiment/data/s1_a_star_no_sas_experiment_processed.csv};
\addplot[fill=\orangeColor!20,fill opacity=0.3,draw=none,forget plot] fill between[of=upper2 and lower2];
\addplot[thick,color=\orangeColor] table[x=nodes,y=mean_time,col sep=comma] {navigation_experiment/data/s1_a_star_no_sas_experiment_processed.csv};
\addlegendentry{No adap. mean}
\end{axis}
\end{tikzpicture}

%% file: navigation_experiment/plots/s4_time_plot.tex
\begin{tikzpicture}[baseline=(paperAxis.north)]
\begin{axis}[name=paperAxis,
width=\textwidth,
height=0.75\textwidth,
xlabel={Number of mission actions},
ylabel={Planning Time [seconds]},
xmin=1,
xmax=5,
ymin=0,
ymax=18.3531757133,
xtick={1,2,3,4,5},
grid=major,
legend pos=north west,
legend style={font=\footnotesize,nodes={scale=0.8,transform shape}},
legend image post style={scale=0.5}
]
\addplot[name path=upper0,draw=none,forget plot] table[x=m,y=mean_time_plus_std,col sep=comma] {navigation_experiment/data/s4_astar_experiment_processed.csv};
\addplot[name path=lower0,draw=none,forget plot] table[x=m,y=mean_time_minus_std,col sep=comma] {navigation_experiment/data/s4_astar_experiment_processed.csv};
\addplot[fill=\greenColor!20,fill opacity=0.3,draw=none,forget plot] fill between[of=upper0 and lower0];
\addplot[thick,color=\greenColor] table[x=m,y=mean_time,col sep=comma] {navigation_experiment/data/s4_astar_experiment_processed.csv};
\addlegendentry{A* mean}
\addplot[name path=upper1,draw=none,forget plot] table[x=m,y=mean_time_plus_std,col sep=comma] {navigation_experiment/data/s4_eager_experiment_processed.csv};
\addplot[name path=lower1,draw=none,forget plot] table[x=m,y=mean_time_minus_std,col sep=comma] {navigation_experiment/data/s4_eager_experiment_processed.csv};
\addplot[fill=\blueColor!20,fill opacity=0.3,draw=none,forget plot] fill between[of=upper1 and lower1];
\addplot[thick,color=\blueColor] table[x=m,y=mean_time,col sep=comma] {navigation_experiment/data/s4_eager_experiment_processed.csv};
\addlegendentry{EG mean}
\addplot[name path=upper2,draw=none,forget plot] table[x=m,y=mean_time_plus_std,col sep=comma] {navigation_experiment/data/s4_astar_no_sas_experiment_processed.csv};
\addplot[name path=lower2,draw=none,forget plot] table[x=m,y=mean_time_minus_std,col sep=comma] {navigation_experiment/data/s4_astar_no_sas_experiment_processed.csv};
\addplot[fill=\orangeColor!20,fill opacity=0.3,draw=none,forget plot] fill between[of=upper2 and lower2];
\addplot[thick,color=\orangeColor] table[x=m,y=mean_time,col sep=comma] {navigation_experiment/data/s4_astar_no_sas_experiment_processed.csv};
\addlegendentry{No adap. mean}
\end{axis}
\end{tikzpicture}

%% file: navigation_experiment/plots/s5_time_plot.tex
\begin{tikzpicture}[baseline=(paperAxis.north)]
\begin{axis}[name=paperAxis,
width=\textwidth,
height=0.75\textwidth,
xlabel={Number of nodes},
ylabel={Planning Time [seconds]},
xmin=9,
xmax=90,
ymin=0,
ymax=220,
xtick={10,20,30,40,50,60,70,80,90},
grid=major,
legend pos=north west,
legend style={font=\footnotesize,nodes={scale=0.8,transform shape}},
legend image post style={scale=0.5}
]
\addplot[name path=upper0,draw=none,forget plot] table[x=nodes,y=mean_time_plus_std,col sep=comma] {navigation_experiment/data/s5_astar_experiment_processed.csv};
\addplot[name path=lower0,draw=none,forget plot] table[x=nodes,y=mean_time_minus_std,col sep=comma] {navigation_experiment/data/s5_astar_experiment_processed.csv};
\addplot[fill=\greenColor!20,fill opacity=0.3,draw=none,forget plot] fill between[of=upper0 and lower0];
\addplot[thick,color=\greenColor] table[x=nodes,y=mean_time,col sep=comma] {navigation_experiment/data/s5_astar_experiment_processed.csv};
\addlegendentry{A* mean}
\addplot[name path=upper1,draw=none,forget plot] table[x=nodes,y=mean_time_plus_std,col sep=comma] {navigation_experiment/data/s5_eager_experiment_processed.csv};
\addplot[name path=lower1,draw=none,forget plot] table[x=nodes,y=mean_time_minus_std,col sep=comma] {navigation_experiment/data/s5_eager_experiment_processed.csv};
\addplot[fill=\blueColor!20,fill opacity=0.3,draw=none,forget plot] fill between[of=upper1 and lower1];
\addplot[thick,color=\blueColor] table[x=nodes,y=mean_time,col sep=comma] {navigation_experiment/data/s5_eager_experiment_processed.csv};
\addlegendentry{EG mean}
\end{axis}
\end{tikzpicture}

%% file: sections/discussion.tex
\section{Discussion}\label{sec:discussion}

This section discusses the SUAVE and navigation results, explains the choice of OWL rather than PDDL for modeling the system architecture, and examines the limitations of \acronym.

\subsection{SUAVE exemplar}

The SUAVE experiments show that applying \acronym to a ROS~2-based robotic exemplar is feasible, but they also reveal mission-performance limitations relative to the other managing subsystems evaluated. BT achieved lower \emph{pipeline search times} and greater \emph{pipeline distances inspected} than \acronym across the original and extended experiments, and ROSA and Metacontrol also outperformed \acronym on several mission-performance comparisons. Therefore, the SUAVE results should be interpreted primarily as evidence that \acronym can be integrated with and executed on a realistic ROS~2-based robotic exemplar, rather than as evidence of mission-performance superiority. The SUAVE exemplar provides a concrete scenario and an experimental setup for evaluating whether managing systems can bridge the gap between theory and practice. The challenges involved in achieving this integration are often underestimated. As shown in~\Cref{sec:impl_plansys}, integrating \acronym with \plansys required extending the PDDL features that \plansys supports, adapting its plan-execution algorithms, and implementing a planner plugin, which introduced substantial engineering overhead.

At the same time, the SUAVE experiment highlights the need for new ROS~2-based exemplars for self-adaptive robotic systems that enable a more fine-grained evaluation of how overall mission quality is affected by performance efficiency, including scalability, reaction time, and memory usage.

\subsection{Navigation use case}\label{sec:disc_nav}
Although both \acronym and the approach proposed by Cámara et al.~\cite{CAMARASoftwareArchitectureTask2020} address TACA for the entire mission, they differ fundamentally in their underlying planning philosophy. \acronym computes a satisficing task and configuration plan that selects the best configuration for each action based on a predefined utility function. In contrast, the approach by Cámara et al.\ seeks to generate an optimal task and configuration policy that maximizes the overall mission utility. Furthermore, \acronym assumes deterministic action effects and configuration utilities, while the method of Cámara et al.\ models probabilistic outcomes, such as collision risks during execution. %

In the experiment in \Cref{sec:nav_exp}, the replicated final step of \CamaraApproach, which computes an optimal policy under a probabilistic model, required more planning time than \acronym. Optimal planning can yield better results when the robot and its environment behave as modeled, but in dynamic environments the assumptions made at design time may not hold during operation. Under these conditions, it can be more advantageous to search for a satisficing solution, monitor during execution whether the current plan remains valid, and replan when it does not, because optimal planning is generally more expensive and therefore incurs a higher penalty at each replanning. A practical strategy could therefore be to use optimal planning for the initial mission plan and satisficing planning for runtime replanning.

\subsection{Modeling the System Architecture in OWL}
There are three main reasons why the robotic system and its functional designs are modeled in OWL rather than directly in PDDL\@.

First, although PDDL supports type hierarchies, OWL offers substantially richer mechanisms for representing class axioms, relations, and inferred knowledge. In PDDL, objects are assigned a type, and type membership follows from explicitly declared type hierarchies. This limits the representation of relationships between classes, which is particularly relevant for robotic components. For example, a specific camera model can be represented as a camera, a visual sensor, and a device requiring electrical power, but these relationships must be explicitly encoded in the type hierarchy. Moreover, it remains unclear whether PDDL types can be treated as predicates~\cite{GOLDMANpddlTypeProblems2012}. In contrast, OWL allows classes to be defined independently and related through axioms, enabling class membership to be inferred rather than explicitly declared. For example, a device requiring electrical power can be defined as any component with a power-consumption property, allowing camera models with this property to be classified accordingly. Together with SWRL rules, OWL also supports inferring relations that are not explicitly asserted. \acronym relies on these inference capabilities to determine which $\FunctionDesigns$ can ground a $\FunctionTomasys$ and which $\FunctionDesign$ has higher utility, as described in \Cref{sec:tomasys_pddl}. These modeling capabilities are particularly valuable for representing robotic systems and are extensively used in \acronym. %

Secondly, there exist many tools for OWL ontologies that assist in modeling classes and their relationships, including the SWRL rules used in this work for inference. These tools allow automatic checking of the model for common errors, such as verifying the consistency of class descriptions and ensuring that instances of the classes can exist without violating other axioms (i.e., satisfiability). In addition, they can visualize the class structure, which further facilitates model development.

Lastly, established standards exist for representing robotic systems and their components using ontologies, such as the IEEE Standard for Autonomous Robotics (AuR) ontology~\cite{IEEEStandardAutonomous2022}. No equivalent standards are known to exist for PDDL\@. Therefore, modeling with OWL ontologies increases the likelihood of producing reusable and interoperable models.

\subsection{Limitations}\label{sec:limitations}
A limitation of \acronym is that it evaluates architectural alternatives locally, i.e., for each $\FunctionTomasys$ required by an action, it selects the realizable $\FunctionDesign$ with the highest utility with respect to that $\FunctionTomasys$. This selection should not be interpreted as mission-level utility optimization. For example, an architectural configuration that provides better performance for an individual function may consume significantly more energy throughout the mission, yet \acronym would still select it.
In domains where architectural choices have long-term effects, such as energy consumption or resource depletion, accounting for such trade-offs would require explicitly modeling their mission-level consequences and incorporating an appropriate optimization objective.

A further limitation concerns the costs associated with architectural reconfiguration. Although reconfiguration is represented through planning actions, the current formulation does not model transition-related costs, such as reconfiguration duration and energy consumption. Consequently, the resulting plans may overlook relevant overhead associated with changing architectural configurations. These aspects could be incorporated by extending the planning model with temporal or cost-aware features, such as durative actions or action costs.

Another limitation of \acronym is its reliance on PDDL derived predicates, a feature not supported by most PDDL-based planners, limiting the number of planners that can be used with \acronym. More specifically, to the best of our knowledge, no existing planner supports both derived predicates and numeric fluents. This poses a significant constraint in the robotics domain, where numeric modeling is essential for representing properties such as battery level, power consumption, and distance.

To mitigate this limitation, the proposed \owltopddl transformation encodes numerical values as symbolic constants and explicitly represents inequality relations among them. This enables symbolic reasoning over ``numeric'' values using inequalities. However, it does not support traditional numerical planning as defined in PDDL 2.1~\cite{fox2003pddl2}, which involves arithmetic operations and numeric fluents.

This work nevertheless keeps derived predicates explicit in the planning domains of \acronym, because compiling them away can, in the worst case, cause an exponential blow-up in the size of the domain formulation or in the length of the shortest plans~\cite{THIEBAUX200538}.

%% file: sections/threats.tex
\section{Threats to Validity}\label{sec:threats}

\paragraph{Construct validity}
The comparison with \CamaraApproach involves different planning guarantees: an optimal policy under probabilistic uncertainty versus a satisficing deterministic plan with replanning (\Cref{sec:disc_nav}). The plan-quality comparison shows that, on the evaluated instances, both approaches produced plans with near-identical path length and estimated remaining battery. However, these metrics are computed for nominal execution and do not capture how each approach performs when probabilistic events, such as collisions, occur. Moreover, only the final step of the approach of Cámara et al.\ was replicated. As discussed in \Cref{sec:nav_exp}, the reported computational performance difference therefore likely underestimates the advantage of \acronym.

\paragraph{Internal validity}
Planner choice and planner configuration influence planning time. \acronym uses Fast Downward in all experiments because it supports the derived predicates and existential preconditions required by the generated planning models (\Cref{sec:impl_plansys}), and other planners or search configurations may yield different planning times. To expose this effect, the scalability experiments compare A* search with the blind heuristic and eager greedy search with the goal-count heuristic, each with and without the adaptation logic (\Cref{sec:scale_exp}). To reduce the effect of random variation, the navigation and scalability experiments were repeated ten times. Each SUAVE experiment was executed 30 times per managing subsystem, with randomized parameters drawn for each group of five consecutive runs (\Cref{tab:suave_params}), and differences between managing subsystems were assessed with the Wilcoxon signed-rank test. All source code and experimental artifacts are publicly available to support replication. %

\paragraph{External validity}
The evaluation covers only two use cases and therefore does not establish that the results generalize to other robotic systems. Both use cases come from the literature and were originally introduced to evaluate other self-adaptation approaches, which reduces the risk of selecting scenarios that favor \acronym. Moreover, \Sone--\Sfive are synthetic variations of the navigation use case. They use generated grid maps and, depending on the experiment, synthetic corridor types, mission-level PDDL actions, or additional architectural configurations that do not model realistic selection criteria. Their results therefore characterize planning-time trends for grid-map navigation missions within the tested parameter ranges (\Cref{tab:scalability_params}). The SUAVE experiments were executed in the Gazebo simulator rather than on a physical vehicle, so they do not capture the behavior of \acronym on a physical AUV\@. All managing subsystems were evaluated in the same simulated setup, which keeps their comparison consistent. Finally, \acronym does not support PDDL numeric fluents (\Cref{sec:limitations}), and the navigation map was discretized to uniform edge lengths, so the results do not extend to problems that require numerical planning.

%% file: sections/conclusions.tex
\section{Conclusion and Future Work}\label{sec:conclusion}

This paper presented \acronym, an ontology-mediated approach to task and architecture co-adaptation that integrates architectural knowledge represented in OWL with task planning problems defined in PDDL\@. By automatically combining these models, \acronym enables joint planning of task and reconfiguration actions using satisficing planning, with replanning when changes during execution invalidate the current plan. An open-source ROS~2 implementation of the approach is provided, including extensions to \plansys to support the PDDL features required for planning and plan execution.

The evaluation in the SUAVE underwater inspection scenario supports the feasibility of runtime co-adaptation, although it also reveals mission-performance limitations relative to other managing subsystems. In the ground navigation scenario, \acronym required less planning time and peak memory than the replicated final step of the approach by Cámara et al.~\cite{CAMARASoftwareArchitectureTask2020}, while producing plans with similar path length and estimated remaining battery. However, these approaches pursue different planning objectives, with \acronym computing satisficing plans and Cámara et al.\ generating optimal policies under probabilistic uncertainty. The scalability experiments further showed that planning time remains manageable as map size, architectural configurations, or quality requirements increase individually, but grows substantially with additional mission-level actions under optimal planning and when multiple problem dimensions are combined. In these more complex scenarios, satisficing planning remained tractable where optimal planning became impractical, although its planning time in the combined scenario remained high for applications requiring frequent replanning. These results indicate that, in the evaluated scenarios, satisficing planning with runtime replanning is a practical alternative to optimal policy generation for task and architecture co-adaptation.

Future work includes addressing the limitations discussed in \Cref{sec:limitations}: selecting architectural alternatives with respect to mission-level objectives rather than per function, modeling reconfiguration costs through action costs or durative actions, and supporting numeric quality attributes without symbolic encoding. Further directions include applying \acronym to diverse real-world robotic systems to assess its reusability and further identify its limitations. Another promising direction is exploring the integration of the proposed planning approach into existing reference architectures for robot self-adaptation, such as ROSA~\cite{rosa}.

%% file: sections/supplement.tex
\section{Algorithms of the \plansys extensions}\label{sec:plansys_algorithms}

This section details the algorithms of the \plansys extensions described in \Cref{M-sec:impl_plansys}.

To support the evaluation of existential preconditions and derived predicates in \plansys, this work proposes the evaluation algorithm described in \Cref{alg:eval}. Lines~2-3 evaluate predicates by unifying their free parameters with all objects in $\pddlState$ that satisfy the predicate,  returning a set of dictionaries of the form $\{\{?x_1:o_1, \ldots{},?x_n:o_n\} \mid (p~o_1,\ldots{},o_n) \in S\}$. Lines 4-6 solve negated conditions by evaluating the condition and returning the complement of the evaluated set $E$ with respect to the current state $\pddlState$, with the form $\{\{?x_1:o_1,\ldots{},?x_n:o_n\} \mid (o_1, \ldots{}, o_n)  \in S \wedge  \{?x_1:o_1, \ldots{},?x_n:o_n\} \notin E\}$. Lines 7-13 evaluate conjunctions of conditions by individually evaluating each condition and joining the solution sets. Lines 14-17 evaluate disjunctions by evaluating each condition and merging the solution sets. Lines 18-20 evaluate existential preconditions by evaluating their body and returning a dictionary without free variables in the form $\{\{?x_1:o_1,\ldots{},?x_n:o_n\} \mid \{?x_1:o_1,\ldots{},?x_n:o_n,?x_{e_{1}}:o_{e_{1}},\ldots{}, ?x_{e_{k}}:o_{e_{k}}\} \in E\}$.

\begin{algorithm}[tb]
    \caption{Evaluation algorithm}%
    \label{alg:eval}
    \begin{algorithmic}[1] %
        \Procedure{Evaluate}{$\algC, \algS$}
            \If{$ \algC = (p~?x_1 \ldots{} ?x_n)$}
                \State \textbf{return} $\Call{UnifyPredicate}{\algC, \algS}$
            \ElsIf{$\algC = (not~(\algC'))$}
                \State $\algE \gets \Call{Evaluate}{\algC', \algS}$
                \State \textbf{return} $\Call{ComplementSet}{\algE, \algS}$
            \ElsIf{$\algC = (and~(\algC'~\ldots{}~\algC''))$}
                \State $\algE \gets \{\emptyset\}$
                \For{$c \in \{\algC', \ldots{}, \algC''\}$}
                    \State $\algE' \gets \Call{Evaluate}{c,~\algS}$
                    \State $\algE \gets \Call{Join}{\algE,~\algE'}$
                \EndFor
                \State \textbf{return} $\algE$
            \ElsIf{$\algC = (or~(\algC'~\algC''))$}
                \State $\algE' \gets \Call{Evaluate}{\algC',~\algS}$
                \State $\algE'' \gets \Call{Evaluate}{\algC'',~\algS}$
                \State \textbf{return} $\Call{Union}{\algE',~\algE''}$
             \ElsIf{$\algC = (exists~(?x_{e_{1}} \ldots{} ?x_{e_{k}})~\algC')$}
                \State $\algE \gets \Call{Evaluate}{\algC',~\algS}$
                \State \textbf{return} $\algE \setminus \{?x_{e_{1}},\ldots{},?x_{e_{k}}\}$
            \EndIf
        \EndProcedure
    \end{algorithmic}
\end{algorithm}

To solve derived predicates, this work introduces Algorithm~\ref{alg:solve_derived}. It partitions the resolution graph into acyclic and cyclic components. The acyclic components are resolved in topological order, while the cyclic components are resolved using a fixed-point iteration. Root nodes are considered solved by definition, as they correspond to non-derived predicates. For instance, in the graph shown in \Cref{M-fig:derived_graph}, the derived predicates $\texttt{inferredB}$ and $\texttt{inferredAB}$ form a cycle, whereas the other derived predicates are acyclic.

Line~2 calls the $\texttt{GetDerivedSCCs}$ function, which computes the strongly connected components~(SCCs) of the subgraph formed by traversing all descendant nodes of the $\texttt{Roots}$ nodes in the resolution graph, using Tarjan's algorithm~\cite{doi:10.1137/0201010}. The function returns a vector of SCCs in topological order. Each SCC contains only derived predicates, explicitly excluding predicates, functions, and instances. For example, if $\texttt{Roots} = \{\texttt{predicateA}, \texttt{predicateB}\}$,
\begin{equation*}
\texttt{GetDerivedSCCs}(\texttt{Graph}, \texttt{Roots}) = \big\{ \{\texttt{inferredA}\},\, \{\texttt{inferredAA}\},\, \{\texttt{inferredB},\, \texttt{inferredAB}\},\, \{\texttt{inferredBB}\} \big\}.
\end{equation*}

Lines~4--15 solve the cyclic SCCs using a fixed-point iteration, while lines~17--22 handle each acyclic SCC by solving its single derived predicate. The $\texttt{Unground}$ function removes all existing assertions of $d$ and its descendants from the state $\pddlState$. The $\texttt{Evaluate}$ function evaluates the preconditions of the derived predicate $d$, storing all valid parameter bindings in the set $E$ (e.g., $\{?x = \texttt{instance1}\}$). The $\texttt{GroundPredicates}$ function grounds the derived predicate for each substitution in $E$ (e.g., $(\texttt{inferredA}~\texttt{instance1})$) and adds the resulting assertions to the state.

\begin{algorithm}[tb]
\caption{Derived Predicates Evaluation}%
\label{alg:solve_derived}
    \begin{algorithmic}[1]
    \Procedure{SolveDerived}{$\algGraph, \algRoots, \algS$}
        \State $\algSCCs \gets \Call{GetDerivedSCCs}{\algGraph, \algRoots}$
        \For{$\algSCC \in \algSCCs$}
            \If{$\Call{IsCyclic}{\algSCC}$}
                \For{$d \in \algSCC$}
                    \State $\algS \gets \Call{Unground}{\algS, d}$
                \EndFor
                \Repeat
                    \State $\algS' \gets \algS$
                    \For{$d \in \algSCC$}
                        \State $\algE \gets \Call{Evaluate}{\mathrm{pre}(d), \algS}$
                        \State $\algP' \gets \Call{GroundPredicates}{d, \algE}$
                        \State $\algS \gets \algS \cup \algP'$
                    \EndFor
                \Until{$\algS = \algS'$}
            \Else
                \For{$d \in \algSCC$}
                    \State $\algS \gets \Call{Unground}{\algS, d}$
                    \State $\algE \gets \Call{Evaluate}{\mathrm{pre}(d), \algS}$
                    \State $\algP' \gets \Call{GroundPredicates}{d, \algE}$
                    \State $\algS \gets \algS \cup \algP'$
                \EndFor
            \EndIf
        \EndFor
    \State \Return $\algS$
    \EndProcedure
    \end{algorithmic}
\end{algorithm}

\Cref{alg:apply_new} defines the $\texttt{ApplyEffects}$ function. Its inputs are the current state $\algS$, the effects $\algEff$ of an action, the resolution graph $\algGraph$, the set $\algNodes$ of modified nodes, and a boolean $\algDerive$ that indicates whether derived predicates should be recomputed. Lines~2--4 add a predicate to the state, and lines~5--7 remove a negated predicate from it. In both cases the predicate is added to $\algNodes$. Lines~8--11 apply conjunctive effects recursively, with $\algDerive$ set to false so that derived predicates are not recomputed after each individual effect. If $\algDerive$ is true (line~13), the function calls $\texttt{SolveDerived}$ (\Cref{alg:solve_derived}) with $\algNodes$ as root nodes, so only the derived predicates that depend on the modified predicates are re-evaluated. The function returns the updated state and the set of modified nodes.

\begin{algorithm}[tb]
    \caption{Apply Action Effects and Selectively Re-evaluate Derived Predicates}%
    \label{alg:apply_new}
    \begin{algorithmic}[1]
    \Procedure{ApplyEffects}{$\algS, \algEff, \algGraph, \algNodes, \algDerive$}
        \If{$\algEff = (p~\underline{o}_1~\dots~\underline{o}_n)$}
            \State $\algS \gets \algS \cup \{\algEff\}$
            \State $\algNodes \gets \algNodes \cup \{\algEff\}$
        \ElsIf{$\algEff = (not~(\algEff'))$}
            \State $\algS \gets \algS \setminus \{\algEff'\}$
            \State $\algNodes \gets \algNodes \cup \{\algEff'\}$
        \ElsIf{$\algEff = (and~(\algEff'_1~\dots~\algEff'_k))$}
            \For{$e \in \{\algEff'_1, \dots, \algEff'_k\}$}
                \State $\algS, \algNodes \gets \Call{ApplyEffects}{\algS, e, \algGraph, \algNodes, \text{false}}$
            \EndFor
        \EndIf
        \If{$\algDerive$}
            \State $\algS \gets \Call{SolveDerived}{\algGraph, \algNodes, \algS}$
        \EndIf
        \State \Return $\algS, \algNodes$
    \EndProcedure
    \end{algorithmic}
\end{algorithm}

\section{Wilcoxon signed-rank comparisons for the SUAVE experiments}\label{sec:wilcoxon}

This section reports the paired Wilcoxon signed-rank comparisons for experiments 1 and 2 and extended experiments 1 and 2 of the SUAVE exemplar. The test setup and hypotheses are described in \Cref{M-sec:suave_exp}, and the comparisons for experiment 3 and extended experiment 3 are reported in \Cref{M-tab:wilcoxon}, whose caption also explains the layout and color legend of the tables below.

\begin{table}[tb]
\centering
\caption{Experiment 1: paired Wilcoxon signed-rank comparisons.}%
\label{tab:wilcoxon-exp1}
\begingroup
  \renewenvironment{table}[1][]{}{}%
  \renewcommand{\caption}[2][]{}%
  \renewcommand{\label}[1]{}%
  \input{tables/wilcoxon_analysis/exp1_wilcoxon.tex}\endgroup
\end{table}

\begin{table}[tb]
\centering
\caption{Experiment 2: paired Wilcoxon signed-rank comparisons.}%
\label{tab:wilcoxon-exp2}
\begingroup
  \renewenvironment{table}[1][]{}{}%
  \renewcommand{\caption}[2][]{}%
  \renewcommand{\label}[1]{}%
  \input{tables/wilcoxon_analysis/exp2_wilcoxon.tex}\endgroup
\end{table}

\begin{table}[tb]
\centering
\caption{Extended experiment 1: paired Wilcoxon signed-rank comparisons.}%
\label{tab:wilcoxon-extended-exp1}
\begingroup
  \renewenvironment{table}[1][]{}{}%
  \renewcommand{\caption}[2][]{}%
  \renewcommand{\label}[1]{}%
  \input{tables/wilcoxon_analysis/extended_exp1_wilcoxon.tex}\endgroup
\end{table}

\begin{table}[tb]
\centering
\caption{Extended experiment 2: paired Wilcoxon signed-rank comparisons.}%
\label{tab:wilcoxon-extended-exp2}
\begingroup
  \renewenvironment{table}[1][]{}{}%
  \renewcommand{\caption}[2][]{}%
  \renewcommand{\label}[1]{}%
  \input{tables/wilcoxon_analysis/extended_exp2_wilcoxon.tex}\endgroup
\end{table}

\section{Per-experiment results of the scalability experiments}\label{sec:scalability_plots}

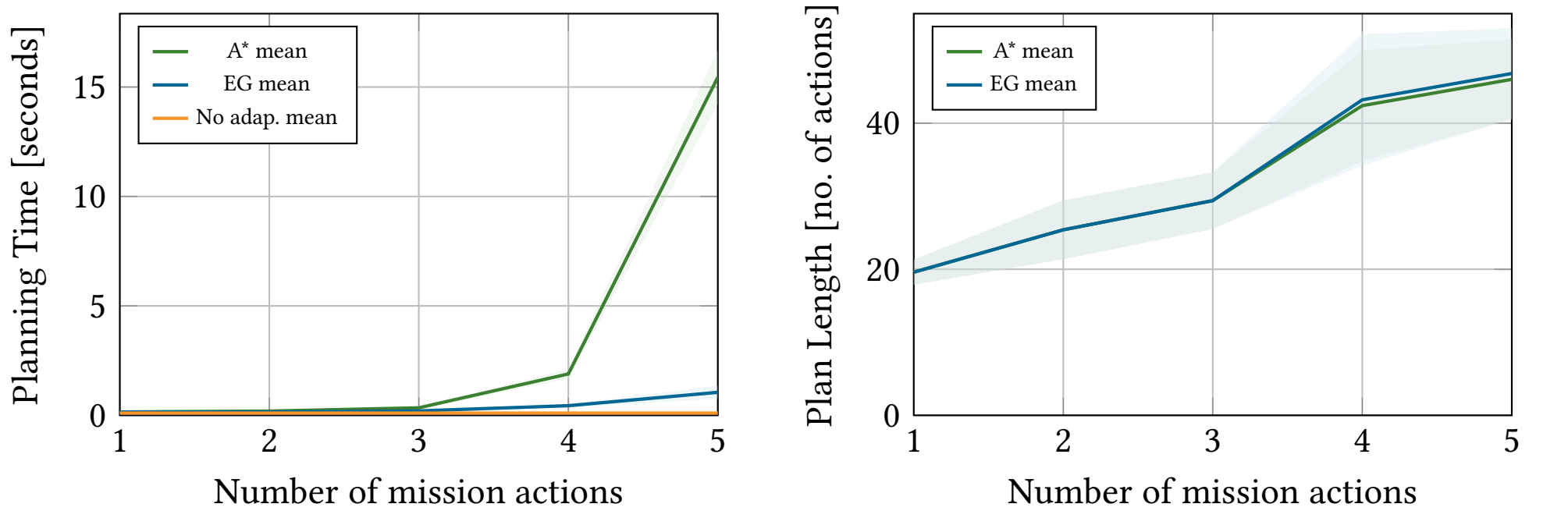
\begin{figure}[tb]
    \centering
    \begin{subfigure}[t]{0.44\linewidth}
        \centering
        \input{navigation_experiment/plots/s4_time_plot}
        \caption{Planning time as mission-level actions increase.}%
        \label{plot:supp_s4_planning_time}
    \end{subfigure}
    \begin{subfigure}[t]{0.44\linewidth}
        \centering
        \input{navigation_experiment/plots/s4_action_plot}
        \caption{Plan length as mission-level actions increase.}%
        \label{plot:s4_action_number}
    \end{subfigure}
    \caption{\Sfour results: Planning time and plan length as mission-level actions increase. The shaded area shows the standard deviation. EG stands for the eager greedy search.}%
    \label{fig:s4_results}
\Description{Two line-plot panels show mean planning time in seconds and mean plan length in action
counts as the number of mission-level action schemas increases from 1 to 5. A* planning
time rises sharply to 15.45 seconds, while eager greedy search reaches 1.05 seconds and
the no-adaptation baseline remains near 0.10 seconds. Plan lengths increase for both
search strategies and remain similar, reaching 46.00 and 46.80 actions, respectively.
Shaded bands indicate standard deviations.}
\end{figure}

This section reports planning time and plan length for each scalability experiment of the navigation use case. The experimental design and setup are described in \Cref{M-sec:scale_exp}, and the results are summarized in \Cref{M-tab:scalability_summary} and \Cref{M-fig:scalability_time}. In \Cref{fig:s1_results,fig:s4_results}, the no-adaptation curve reports A* search.

\begin{figure}[tb]
    \centering
    \begin{subfigure}[t]{0.44\linewidth}
        \centering
        \input{navigation_experiment/plots/s1_time_plot}
        \caption{Planning time as map size increases.}%
        \label{plot:supp_s1_planning_time}
    \end{subfigure}
    \begin{subfigure}[t]{0.44\linewidth}
        \centering
        \input{navigation_experiment/plots/s1_action_plot}
        \caption{Plan length as map size increases. Both search strategies overlap.}%
        \label{plot:s1_action_number}
    \end{subfigure}
    \caption{\Sone results: Planning time and plan length as map size increases. The shaded area shows the standard deviation. EG stands for the eager greedy search.}%
    \label{fig:s1_results}
\Description{Two line-plot panels show mean planning time in seconds and mean plan length in action
counts against map size, from 9 to 900 retained nodes. A* and eager greedy search have
similar planning times, reaching 0.92 and 0.95 seconds, respectively, at 900 nodes, and overlapping
plan-length curves, reaching 47.20 actions. The left panel also shows the faster
no-adaptation baseline. Shaded bands indicate standard deviations.}
\end{figure}

\begin{figure}[tb]
    \centering
    \begin{subfigure}[t]{0.44\linewidth}
        \centering
        \input{navigation_experiment/plots/s2_time_plot}
        \caption{Planning time as architectural configurations increase.}%
        \label{plot:s2_planning_time}
    \end{subfigure}
    \begin{subfigure}[t]{0.44\linewidth}
        \centering
        \input{navigation_experiment/plots/s2_action_plot}
        \caption{Plan length as architectural configurations increase. Both search strategies overlap.}%
        \label{plot:s2_action_number}
    \end{subfigure}
    \caption{\Stwo results: Planning time and plan length as architectural configurations increase. The shaded area shows the standard deviation. EG stands for the eager greedy search.}%
    \label{fig:s2_results}
\Description{Two line-plot panels show mean planning time in seconds and mean plan length in action
counts as the number of architectural configurations increases from 2 to 15. Planning
time changes modestly, reaching 0.30 seconds for both A* and eager
greedy search. The plan-length curves overlap, reaching 20.70 actions at 15
configurations. Shaded bands indicate standard deviations.}
\end{figure}
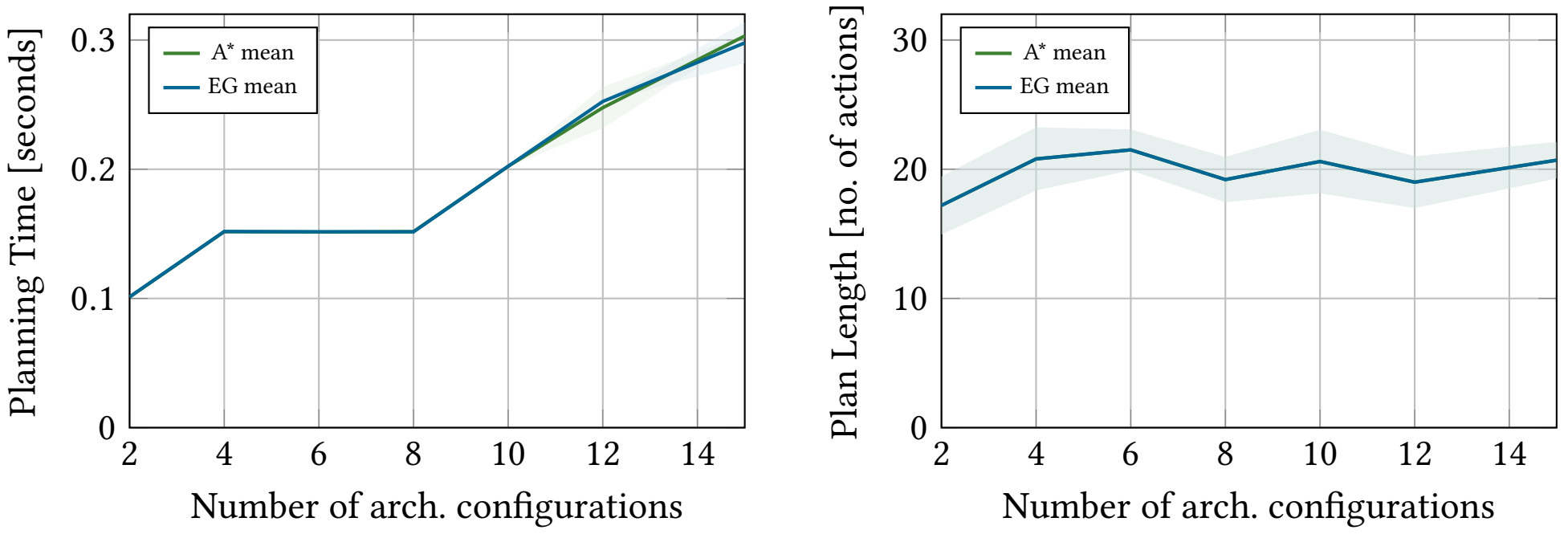

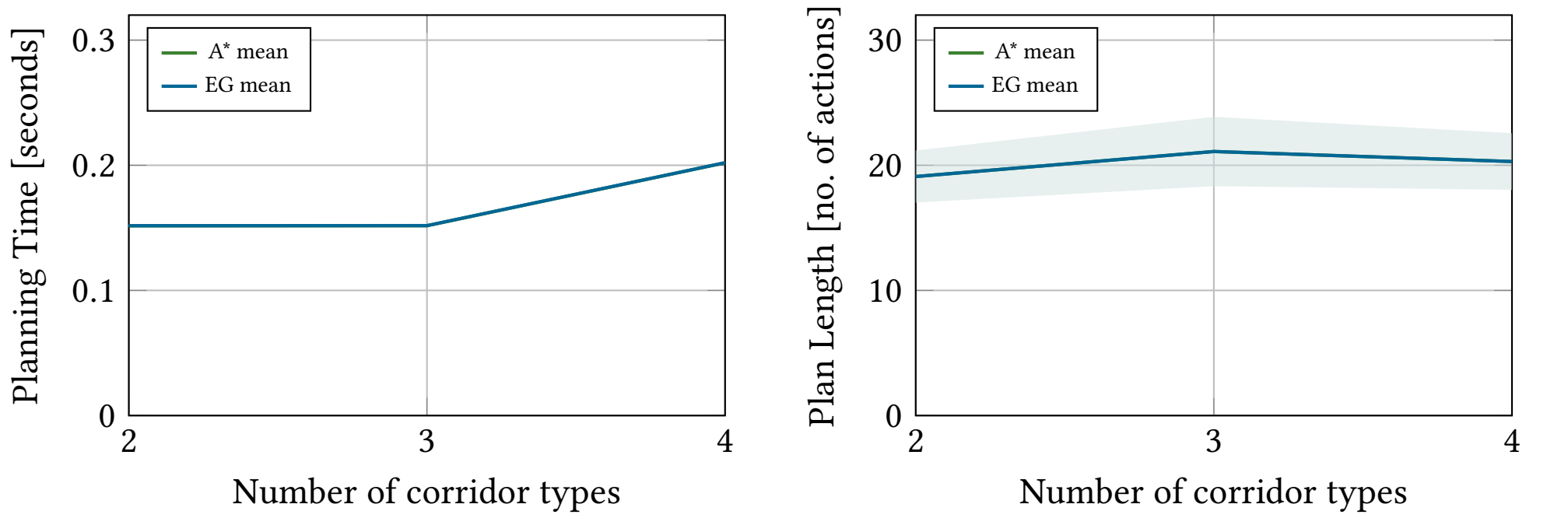
\begin{figure}[tb]
    \centering
    \begin{subfigure}[t]{0.44\linewidth}
        \centering
        \input{navigation_experiment/plots/s3_time_plot}
        \caption{Planning time as corridor types increase.}%
        \label{plot:s3_planning_time}
    \end{subfigure}
    \begin{subfigure}[t]{0.44\linewidth}
        \centering
        \input{navigation_experiment/plots/s3_action_plot}
        \caption{Plan length as corridor types increase. Both search strategies overlap.}%
        \label{plot:s3_action_number}
    \end{subfigure}
    \caption{\Sthree results: Planning time and plan length as corridor types increase. The shaded area shows the standard deviation. EG stands for the eager greedy search.}%
    \label{fig:s3_results}
\Description{Two line-plot panels show mean planning time in seconds and mean plan length in action
counts as the number of corridor types increases from 2 to 4. A* and eager greedy search
have nearly identical planning times, increasing from about 0.15 to 0.20 seconds. Their
plan-length curves overlap, with means of 19.1 actions at two corridor types and 20.3 at
four. Shaded bands indicate standard deviations.}
\end{figure}

\begin{figure}[tb]
    \centering
    \begin{subfigure}[t]{0.44\linewidth}
        \centering
        \input{navigation_experiment/plots/s5_time_plot}
        \caption{Planning time as map size increases.}%
        \label{plot:supp_s5_planning_time}
    \end{subfigure}
    \begin{subfigure}[t]{0.44\linewidth}
        \centering
        \input{navigation_experiment/plots/s5_action_plot}
        \caption{Plan length as map size increases.}%
        \label{plot:s5_action_number}
    \end{subfigure}
    \caption{\Sfive results: Planning time and plan length as map size increases. The shaded area shows the standard deviation. EG stands for the eager greedy search.}%
    \label{fig:s5_results}
\Description{Two line-plot panels show mean planning time and mean plan length against map size, from
9 to 90 retained nodes, in the combined stress test. A* planning time rises to 188.96
seconds, whereas eager greedy search reaches 13.73 seconds. Plan lengths increase with
map size, reaching 48.10 actions for A* and 51.60 for eager greedy search. Shaded bands
indicate standard deviations.}
\end{figure}

%% file: tables/wilcoxon_analysis/exp1_wilcoxon.tex
\begin{table}[tb]
\centering
\caption{Experiment 1: paired Wilcoxon signed-rank comparisons. Each cell reports the unadjusted p-values for the comparison between the managing subsystem in the row and the one in the column for both the \emph{pipeline search time} and \emph{pipeline distance inspected} metrics. Cells with a green background indicate that the alternative hypothesis was favored, and cells with a blue background indicate that the null hypothesis cannot be rejected in favor of the alternative hypothesis. Values rounding to 0.000 are shown as $<0.001$. ``MC'' stands for Metacontrol.}%
\label{tab:wilcoxon-exp1}
\renewcommand{\arraystretch}{1.2} %
\resizebox{\textwidth}{!}{%
\begin{tabular}{p{1.65cm}*{6}{p{1.3cm}}*{6}{p{1.3cm}}}
\toprule
\multirow{2}{1.65cm}{\raggedright \textbf{Managing subsystem}} & \multicolumn{6}{c}{\textbf{Pipeline search time}} & \multicolumn{6}{c}{\textbf{Pipeline distance inspected}} \\ \cline{2-13}
 & \textbf{None} & \textbf{Random} & \textbf{BT} & \textbf{MC} & \textbf{ROSA} & \textbf{PLANTA} & \textbf{None} & \textbf{Random} & \textbf{BT} & \textbf{MC} & \textbf{ROSA} & \textbf{PLANTA} \\
\midrule
\textbf{None} & - & \cellcolor{pvalue_green}$<0.001$ & \cellcolor{pvalue_blue}0.997 & \cellcolor{pvalue_green}0.013 & \cellcolor{pvalue_green}$<0.001$ & \cellcolor{pvalue_green}$<0.001$ & - & \cellcolor{pvalue_green}$<0.001$ & \cellcolor{pvalue_blue}1.000 & \cellcolor{pvalue_blue}1.000 & \cellcolor{pvalue_blue}1.000 & \cellcolor{pvalue_green}0.010 \\
\textbf{Random} & \cellcolor{pvalue_blue}1.000 & - & \cellcolor{pvalue_blue}1.000 & \cellcolor{pvalue_blue}1.000 & \cellcolor{pvalue_blue}0.999 & \cellcolor{pvalue_blue}0.999 & \cellcolor{pvalue_blue}1.000 & - & \cellcolor{pvalue_blue}1.000 & \cellcolor{pvalue_blue}1.000 & \cellcolor{pvalue_blue}1.000 & \cellcolor{pvalue_blue}1.000 \\
\textbf{BT} & \cellcolor{pvalue_green}0.003 & \cellcolor{pvalue_green}$<0.001$ & - & \cellcolor{pvalue_green}0.001 & \cellcolor{pvalue_green}$<0.001$ & \cellcolor{pvalue_green}$<0.001$ & \cellcolor{pvalue_green}$<0.001$ & \cellcolor{pvalue_green}$<0.001$ & - & \cellcolor{pvalue_blue}0.767 & \cellcolor{pvalue_blue}0.241 & \cellcolor{pvalue_green}$<0.001$ \\
\textbf{MC} & \cellcolor{pvalue_blue}0.987 & \cellcolor{pvalue_green}$<0.001$ & \cellcolor{pvalue_blue}0.999 & - & \cellcolor{pvalue_green}$<0.001$ & \cellcolor{pvalue_green}0.016 & \cellcolor{pvalue_green}$<0.001$ & \cellcolor{pvalue_green}$<0.001$ & \cellcolor{pvalue_blue}0.233 & - & \cellcolor{pvalue_green}0.015 & \cellcolor{pvalue_green}$<0.001$ \\
\textbf{ROSA} & \cellcolor{pvalue_blue}1.000 & \cellcolor{pvalue_green}0.001 & \cellcolor{pvalue_blue}1.000 & \cellcolor{pvalue_blue}1.000 & - & \cellcolor{pvalue_blue}0.846 & \cellcolor{pvalue_green}$<0.001$ & \cellcolor{pvalue_green}$<0.001$ & \cellcolor{pvalue_blue}0.759 & \cellcolor{pvalue_blue}0.985 & - & \cellcolor{pvalue_green}$<0.001$ \\
\textbf{PLANTA} & \cellcolor{pvalue_blue}1.000 & \cellcolor{pvalue_green}0.001 & \cellcolor{pvalue_blue}1.000 & \cellcolor{pvalue_blue}0.984 & \cellcolor{pvalue_blue}0.154 & - & \cellcolor{pvalue_blue}0.991 & \cellcolor{pvalue_green}$<0.001$ & \cellcolor{pvalue_blue}1.000 & \cellcolor{pvalue_blue}1.000 & \cellcolor{pvalue_blue}1.000 & - \\
\bottomrule
\end{tabular}
}
\end{table}

%% file: tables/wilcoxon_analysis/exp2_wilcoxon.tex
\begin{table}[tb]
\centering
\caption{Experiment 2: paired Wilcoxon signed-rank comparisons. Each cell reports the unadjusted p-values for the comparison between the managing subsystem in the row and the one in the column for both the \emph{pipeline search time} and \emph{pipeline distance inspected} metrics. Cells with a green background indicate that the alternative hypothesis was favored, and cells with a blue background indicate that the null hypothesis cannot be rejected in favor of the alternative hypothesis. Values rounding to 0.000 are shown as $<0.001$. ``MC'' stands for Metacontrol.}%
\label{tab:wilcoxon-exp2}
\renewcommand{\arraystretch}{1.2} %
\resizebox{\textwidth}{!}{%
\begin{tabular}{p{1.65cm}*{6}{p{1.3cm}}*{6}{p{1.3cm}}}
\toprule
\multirow{2}{1.65cm}{\raggedright \textbf{Managing subsystem}} & \multicolumn{6}{c}{\textbf{Pipeline search time}} & \multicolumn{6}{c}{\textbf{Pipeline distance inspected}} \\ \cline{2-13}
 & \textbf{None} & \textbf{Random} & \textbf{BT} & \textbf{MC} & \textbf{ROSA} & \textbf{PLANTA} & \textbf{None} & \textbf{Random} & \textbf{BT} & \textbf{MC} & \textbf{ROSA} & \textbf{PLANTA} \\
\midrule
\textbf{None} & - & \cellcolor{pvalue_blue}0.997 & \cellcolor{pvalue_blue}1.000 & \cellcolor{pvalue_blue}1.000 & \cellcolor{pvalue_blue}1.000 & \cellcolor{pvalue_blue}1.000 & - & \cellcolor{pvalue_green}$<0.001$ & \cellcolor{pvalue_blue}1.000 & \cellcolor{pvalue_blue}1.000 & \cellcolor{pvalue_blue}1.000 & \cellcolor{pvalue_blue}1.000 \\
\textbf{Random} & \cellcolor{pvalue_green}0.003 & - & \cellcolor{pvalue_blue}0.999 & \cellcolor{pvalue_blue}0.994 & \cellcolor{pvalue_blue}0.960 & \cellcolor{pvalue_blue}0.974 & \cellcolor{pvalue_blue}1.000 & - & \cellcolor{pvalue_blue}1.000 & \cellcolor{pvalue_blue}1.000 & \cellcolor{pvalue_blue}1.000 & \cellcolor{pvalue_blue}1.000 \\
\textbf{BT} & \cellcolor{pvalue_green}$<0.001$ & \cellcolor{pvalue_green}0.001 & - & \cellcolor{pvalue_green}$<0.001$ & \cellcolor{pvalue_green}$<0.001$ & \cellcolor{pvalue_green}$<0.001$ & \cellcolor{pvalue_green}$<0.001$ & \cellcolor{pvalue_green}$<0.001$ & - & \cellcolor{pvalue_blue}0.333 & \cellcolor{pvalue_blue}0.052 & \cellcolor{pvalue_green}$<0.001$ \\
\textbf{MC} & \cellcolor{pvalue_green}$<0.001$ & \cellcolor{pvalue_green}0.006 & \cellcolor{pvalue_blue}1.000 & - & \cellcolor{pvalue_green}$<0.001$ & \cellcolor{pvalue_green}$<0.001$ & \cellcolor{pvalue_green}$<0.001$ & \cellcolor{pvalue_green}$<0.001$ & \cellcolor{pvalue_blue}0.667 & - & \cellcolor{pvalue_blue}0.104 & \cellcolor{pvalue_green}$<0.001$ \\
\textbf{ROSA} & \cellcolor{pvalue_green}$<0.001$ & \cellcolor{pvalue_green}0.040 & \cellcolor{pvalue_blue}1.000 & \cellcolor{pvalue_blue}1.000 & - & \cellcolor{pvalue_blue}0.988 & \cellcolor{pvalue_green}$<0.001$ & \cellcolor{pvalue_green}$<0.001$ & \cellcolor{pvalue_blue}0.948 & \cellcolor{pvalue_blue}0.896 & - & \cellcolor{pvalue_green}$<0.001$ \\
\textbf{PLANTA} & \cellcolor{pvalue_green}$<0.001$ & \cellcolor{pvalue_green}0.026 & \cellcolor{pvalue_blue}1.000 & \cellcolor{pvalue_blue}1.000 & \cellcolor{pvalue_green}0.012 & - & \cellcolor{pvalue_green}$<0.001$ & \cellcolor{pvalue_green}$<0.001$ & \cellcolor{pvalue_blue}1.000 & \cellcolor{pvalue_blue}1.000 & \cellcolor{pvalue_blue}1.000 & - \\
\bottomrule
\end{tabular}
}
\end{table}

%% file: tables/wilcoxon_analysis/extended_exp1_wilcoxon.tex
\begin{table}[tb]
\centering
\caption{Extended experiment 1: paired Wilcoxon signed-rank comparisons. Each cell reports the unadjusted p-values for the comparison between the managing subsystem in the row and the one in the column for both the \emph{pipeline search time} and \emph{pipeline distance inspected} metrics. Cells with a green background indicate that the alternative hypothesis was favored, and cells with a blue background indicate that the null hypothesis cannot be rejected in favor of the alternative hypothesis. Values rounding to 0.000 are shown as $<0.001$.}
\label{tab:wilcoxon-extended-exp1}
\renewcommand{\arraystretch}{1.2} %
\resizebox{0.7\textwidth}{!}{%
\begin{tabular}{p{1.65cm}|*{3}{p{1.3cm}}|*{3}{p{1.3cm}}}
\hline
\multirow{2}{1.65cm}{\raggedright \textbf{Managing subsystem}} & \multicolumn{3}{c|}{\textbf{Pipeline search time}} & \multicolumn{3}{c}{\textbf{Pipeline distance inspected}} \\ \cline{2-7}
 & \textbf{BT} & \textbf{ROSA} & \textbf{PLANTA} & \textbf{BT} & \textbf{ROSA} & \textbf{PLANTA} \\
\hline
\textbf{BT} & - & \cellcolor{pvalue_green}0.001 & \cellcolor{pvalue_green}$<0.001$ & - & \cellcolor{pvalue_green}0.002 & \cellcolor{pvalue_green}$<0.001$ \\
\textbf{ROSA} & \cellcolor{pvalue_blue}0.999 & - & \cellcolor{pvalue_blue}0.352 & \cellcolor{pvalue_blue}0.998 & - & \cellcolor{pvalue_green}$<0.001$ \\
\textbf{PLANTA} & \cellcolor{pvalue_blue}1.000 & \cellcolor{pvalue_blue}0.648 & - & \cellcolor{pvalue_blue}1.000 & \cellcolor{pvalue_blue}1.000 & - \\
\hline
\end{tabular}
}
\end{table}

%% file: tables/wilcoxon_analysis/extended_exp2_wilcoxon.tex
\begin{table}[tb]
\centering
\caption{Extended experiment 2: paired Wilcoxon signed-rank comparisons. Each cell reports the unadjusted p-values for the comparison between the managing subsystem in the row and the one in the column for both the \emph{pipeline search time} and \emph{pipeline distance inspected} metrics. Cells with a green background indicate that the alternative hypothesis was favored, and cells with a blue background indicate that the null hypothesis cannot be rejected in favor of the alternative hypothesis. Values rounding to 0.000 are shown as $<0.001$.}
\label{tab:wilcoxon-extended-exp2}
\renewcommand{\arraystretch}{1.2} %
\resizebox{0.7\textwidth}{!}{%
\begin{tabular}{p{1.65cm}|*{3}{p{1.3cm}}|*{3}{p{1.3cm}}}
\hline
\multirow{2}{1.65cm}{\raggedright \textbf{Managing subsystem}} & \multicolumn{3}{c|}{\textbf{Pipeline search time}} & \multicolumn{3}{c}{\textbf{Pipeline distance inspected}} \\ \cline{2-7}
 & \textbf{BT} & \textbf{ROSA} & \textbf{PLANTA} & \textbf{BT} & \textbf{ROSA} & \textbf{PLANTA} \\
\hline
\textbf{BT} & - & \cellcolor{pvalue_green}$<0.001$ & \cellcolor{pvalue_green}$<0.001$ & - & \cellcolor{pvalue_green}$<0.001$ & \cellcolor{pvalue_green}$<0.001$ \\
\textbf{ROSA} & \cellcolor{pvalue_blue}1.000 & - & \cellcolor{pvalue_blue}0.862 & \cellcolor{pvalue_blue}1.000 & - & \cellcolor{pvalue_green}$<0.001$ \\
\textbf{PLANTA} & \cellcolor{pvalue_blue}1.000 & \cellcolor{pvalue_blue}0.138 & - & \cellcolor{pvalue_blue}1.000 & \cellcolor{pvalue_blue}1.000 & - \\
\hline
\end{tabular}
}
\end{table}

%% file: navigation_experiment/plots/s1_action_plot.tex
\begin{tikzpicture}[baseline=(paperAxis.north)]
\begin{axis}[name=paperAxis,
width=\textwidth,
height=0.75\textwidth,
xlabel={Number of Nodes},
ylabel={Plan Length [no. of actions]},
xmin=0,
xmax=900,
ymin=0,
ymax=70,
xtick={0,200,400,600,800,900},
grid=major,
legend pos=north west,
legend style={font=\footnotesize,nodes={scale=0.8,transform shape}},
legend image post style={scale=0.5}
]
\addplot[name path=upper0,draw=none,forget plot] table[x=nodes,y=mean_action_count_plus_std,col sep=comma] {navigation_experiment/data/s1_a_star_experiment_processed.csv};
\addplot[name path=lower0,draw=none,forget plot] table[x=nodes,y=mean_action_count_minus_std,col sep=comma] {navigation_experiment/data/s1_a_star_experiment_processed.csv};
\addplot[fill=\greenColor!20,fill opacity=0.3,draw=none,forget plot] fill between[of=upper0 and lower0];
\addplot[thick,color=\greenColor] table[x=nodes,y=mean_action_count,col sep=comma] {navigation_experiment/data/s1_a_star_experiment_processed.csv};
\addlegendentry{A* mean}
\addplot[name path=upper1,draw=none,forget plot] table[x=nodes,y=mean_action_count_plus_std,col sep=comma] {navigation_experiment/data/s1_eager_experiment_processed.csv};
\addplot[name path=lower1,draw=none,forget plot] table[x=nodes,y=mean_action_count_minus_std,col sep=comma] {navigation_experiment/data/s1_eager_experiment_processed.csv};
\addplot[fill=\blueColor!20,fill opacity=0.3,draw=none,forget plot] fill between[of=upper1 and lower1];
\addplot[thick,color=\blueColor] table[x=nodes,y=mean_action_count,col sep=comma] {navigation_experiment/data/s1_eager_experiment_processed.csv};
\addlegendentry{EG mean}
\end{axis}
\end{tikzpicture}

%% file: navigation_experiment/plots/s2_time_plot.tex
\begin{tikzpicture}[baseline=(paperAxis.north)]
\begin{axis}[name=paperAxis,
width=\textwidth,
height=0.75\textwidth,
xlabel={Number of arch. configurations},
ylabel={Planning Time [seconds]},
xmin=2,
xmax=15,
ymin=0,
ymax=0.32,
xtick={2,4,6,8,10,12,14},
grid=major,
legend pos=north west,
legend style={font=\footnotesize,nodes={scale=0.8,transform shape}},
legend image post style={scale=0.5}
]
\addplot[name path=upper0,draw=none,forget plot] table[x=n_fd,y=mean_time_plus_std,col sep=comma] {navigation_experiment/data/s2_astar_experiment_processed.csv};
\addplot[name path=lower0,draw=none,forget plot] table[x=n_fd,y=mean_time_minus_std,col sep=comma] {navigation_experiment/data/s2_astar_experiment_processed.csv};
\addplot[fill=\greenColor!20,fill opacity=0.3,draw=none,forget plot] fill between[of=upper0 and lower0];
\addplot[thick,color=\greenColor] table[x=n_fd,y=mean_time,col sep=comma] {navigation_experiment/data/s2_astar_experiment_processed.csv};
\addlegendentry{A* mean}
\addplot[name path=upper1,draw=none,forget plot] table[x=n_fd,y=mean_time_plus_std,col sep=comma] {navigation_experiment/data/s2_eager_experiment_processed.csv};
\addplot[name path=lower1,draw=none,forget plot] table[x=n_fd,y=mean_time_minus_std,col sep=comma] {navigation_experiment/data/s2_eager_experiment_processed.csv};
\addplot[fill=\blueColor!20,fill opacity=0.3,draw=none,forget plot] fill between[of=upper1 and lower1];
\addplot[thick,color=\blueColor] table[x=n_fd,y=mean_time,col sep=comma] {navigation_experiment/data/s2_eager_experiment_processed.csv};
\addlegendentry{EG mean}
\end{axis}
\end{tikzpicture}

%% file: navigation_experiment/plots/s2_action_plot.tex
\begin{tikzpicture}[baseline=(paperAxis.north)]
\begin{axis}[name=paperAxis,
width=\textwidth,
height=0.75\textwidth,
xlabel={Number of arch. configurations},
ylabel={Plan Length [no. of actions]},
xmin=2,
xmax=15,
ymin=0,
ymax=32,
xtick={2,4,6,8,10,12,14},
grid=major,
legend pos=north west,
legend style={font=\footnotesize,nodes={scale=0.8,transform shape}},
legend image post style={scale=0.5}
]
\addplot[name path=upper0,draw=none,forget plot] table[x=n_fd,y=mean_action_count_plus_std,col sep=comma] {navigation_experiment/data/s2_astar_experiment_processed.csv};
\addplot[name path=lower0,draw=none,forget plot] table[x=n_fd,y=mean_action_count_minus_std,col sep=comma] {navigation_experiment/data/s2_astar_experiment_processed.csv};
\addplot[fill=\greenColor!20,fill opacity=0.3,draw=none,forget plot] fill between[of=upper0 and lower0];
\addplot[thick,color=\greenColor] table[x=n_fd,y=mean_action_count,col sep=comma] {navigation_experiment/data/s2_astar_experiment_processed.csv};
\addlegendentry{A* mean}
\addplot[name path=upper1,draw=none,forget plot] table[x=n_fd,y=mean_action_count_plus_std,col sep=comma] {navigation_experiment/data/s2_eager_experiment_processed.csv};
\addplot[name path=lower1,draw=none,forget plot] table[x=n_fd,y=mean_action_count_minus_std,col sep=comma] {navigation_experiment/data/s2_eager_experiment_processed.csv};
\addplot[fill=\blueColor!20,fill opacity=0.3,draw=none,forget plot] fill between[of=upper1 and lower1];
\addplot[thick,color=\blueColor] table[x=n_fd,y=mean_action_count,col sep=comma] {navigation_experiment/data/s2_eager_experiment_processed.csv};
\addlegendentry{EG mean}
\end{axis}
\end{tikzpicture}

%% file: navigation_experiment/plots/s3_time_plot.tex
\begin{tikzpicture}[baseline=(paperAxis.north)]
\begin{axis}[name=paperAxis,
width=\textwidth,
height=0.75\textwidth,
xlabel={Number of corridor types},
ylabel={Planning Time [seconds]},
xmin=2,
xmax=4,
ymin=0,
ymax=0.32,
xtick={2,3,4},
grid=major,
legend pos=north west,
legend style={font=\footnotesize,nodes={scale=0.8,transform shape}},
legend image post style={scale=0.5}
]
\addplot[name path=upper0,draw=none,forget plot] table[x=k,y=mean_time_plus_std,col sep=comma] {navigation_experiment/data/s3_astar_experiment_processed.csv};
\addplot[name path=lower0,draw=none,forget plot] table[x=k,y=mean_time_minus_std,col sep=comma] {navigation_experiment/data/s3_astar_experiment_processed.csv};
\addplot[fill=\greenColor!20,fill opacity=0.3,draw=none,forget plot] fill between[of=upper0 and lower0];
\addplot[thick,color=\greenColor] table[x=k,y=mean_time,col sep=comma] {navigation_experiment/data/s3_astar_experiment_processed.csv};
\addlegendentry{A* mean}
\addplot[name path=upper1,draw=none,forget plot] table[x=k,y=mean_time_plus_std,col sep=comma] {navigation_experiment/data/s3_eager_experiment_processed.csv};
\addplot[name path=lower1,draw=none,forget plot] table[x=k,y=mean_time_minus_std,col sep=comma] {navigation_experiment/data/s3_eager_experiment_processed.csv};
\addplot[fill=\blueColor!20,fill opacity=0.3,draw=none,forget plot] fill between[of=upper1 and lower1];
\addplot[thick,color=\blueColor] table[x=k,y=mean_time,col sep=comma] {navigation_experiment/data/s3_eager_experiment_processed.csv};
\addlegendentry{EG mean}
\end{axis}
\end{tikzpicture}

%% file: navigation_experiment/plots/s3_action_plot.tex
\begin{tikzpicture}[baseline=(paperAxis.north)]
\begin{axis}[name=paperAxis,
width=\textwidth,
height=0.75\textwidth,
xlabel={Number of corridor types},
ylabel={Plan Length [no. of actions]},
xmin=2,
xmax=4,
ymin=0,
ymax=32,
xtick={2,3,4},
grid=major,
legend pos=north west,
legend style={font=\footnotesize,nodes={scale=0.8,transform shape}},
legend image post style={scale=0.5}
]
\addplot[name path=upper0,draw=none,forget plot] table[x=k,y=mean_action_count_plus_std,col sep=comma] {navigation_experiment/data/s3_astar_experiment_processed.csv};
\addplot[name path=lower0,draw=none,forget plot] table[x=k,y=mean_action_count_minus_std,col sep=comma] {navigation_experiment/data/s3_astar_experiment_processed.csv};
\addplot[fill=\greenColor!20,fill opacity=0.3,draw=none,forget plot] fill between[of=upper0 and lower0];
\addplot[thick,color=\greenColor] table[x=k,y=mean_action_count,col sep=comma] {navigation_experiment/data/s3_astar_experiment_processed.csv};
\addlegendentry{A* mean}
\addplot[name path=upper1,draw=none,forget plot] table[x=k,y=mean_action_count_plus_std,col sep=comma] {navigation_experiment/data/s3_eager_experiment_processed.csv};
\addplot[name path=lower1,draw=none,forget plot] table[x=k,y=mean_action_count_minus_std,col sep=comma] {navigation_experiment/data/s3_eager_experiment_processed.csv};
\addplot[fill=\blueColor!20,fill opacity=0.3,draw=none,forget plot] fill between[of=upper1 and lower1];
\addplot[thick,color=\blueColor] table[x=k,y=mean_action_count,col sep=comma] {navigation_experiment/data/s3_eager_experiment_processed.csv};
\addlegendentry{EG mean}
\end{axis}
\end{tikzpicture}

%% file: navigation_experiment/plots/s4_action_plot.tex
\begin{tikzpicture}[baseline=(paperAxis.north)]
\begin{axis}[name=paperAxis,
width=\textwidth,
height=0.75\textwidth,
xlabel={Number of mission actions},
ylabel={Plan Length [no. of actions]},
xmin=1,
xmax=5,
ymin=0,
ymax=55,
xtick={1,2,3,4,5},
grid=major,
legend pos=north west,
legend style={font=\footnotesize,nodes={scale=0.8,transform shape}},
legend image post style={scale=0.5}
]
\addplot[name path=upper0,draw=none,forget plot] table[x=m,y=mean_action_count_plus_std,col sep=comma] {navigation_experiment/data/s4_astar_experiment_processed.csv};
\addplot[name path=lower0,draw=none,forget plot] table[x=m,y=mean_action_count_minus_std,col sep=comma] {navigation_experiment/data/s4_astar_experiment_processed.csv};
\addplot[fill=\greenColor!20,fill opacity=0.3,draw=none,forget plot] fill between[of=upper0 and lower0];
\addplot[thick,color=\greenColor] table[x=m,y=mean_action_count,col sep=comma] {navigation_experiment/data/s4_astar_experiment_processed.csv};
\addlegendentry{A* mean}
\addplot[name path=upper1,draw=none,forget plot] table[x=m,y=mean_action_count_plus_std,col sep=comma] {navigation_experiment/data/s4_eager_experiment_processed.csv};
\addplot[name path=lower1,draw=none,forget plot] table[x=m,y=mean_action_count_minus_std,col sep=comma] {navigation_experiment/data/s4_eager_experiment_processed.csv};
\addplot[fill=\blueColor!20,fill opacity=0.3,draw=none,forget plot] fill between[of=upper1 and lower1];
\addplot[thick,color=\blueColor] table[x=m,y=mean_action_count,col sep=comma] {navigation_experiment/data/s4_eager_experiment_processed.csv};
\addlegendentry{EG mean}
\end{axis}
\end{tikzpicture}

%% file: navigation_experiment/plots/s5_action_plot.tex
\begin{tikzpicture}[baseline=(paperAxis.north)]
\begin{axis}[name=paperAxis,
width=\textwidth,
height=0.75\textwidth,
xlabel={Number of nodes},
ylabel={Plan Length [no. of actions]},
xmin=9,
xmax=90,
ymin=0,
ymax=65,
xtick={10,20,30,40,50,60,70,80,90},
grid=major,
legend pos=north west,
legend style={font=\footnotesize,nodes={scale=0.8,transform shape}},
legend image post style={scale=0.5}
]
\addplot[name path=upper0,draw=none,forget plot] table[x=nodes,y=mean_action_count_plus_std,col sep=comma] {navigation_experiment/data/s5_astar_experiment_processed.csv};
\addplot[name path=lower0,draw=none,forget plot] table[x=nodes,y=mean_action_count_minus_std,col sep=comma] {navigation_experiment/data/s5_astar_experiment_processed.csv};
\addplot[fill=\greenColor!20,fill opacity=0.3,draw=none,forget plot] fill between[of=upper0 and lower0];
\addplot[thick,color=\greenColor] table[x=nodes,y=mean_action_count,col sep=comma] {navigation_experiment/data/s5_astar_experiment_processed.csv};
\addlegendentry{A* mean}
\addplot[name path=upper1,draw=none,forget plot] table[x=nodes,y=mean_action_count_plus_std,col sep=comma] {navigation_experiment/data/s5_eager_experiment_processed.csv};
\addplot[name path=lower1,draw=none,forget plot] table[x=nodes,y=mean_action_count_minus_std,col sep=comma] {navigation_experiment/data/s5_eager_experiment_processed.csv};
\addplot[fill=\blueColor!20,fill opacity=0.3,draw=none,forget plot] fill between[of=upper1 and lower1];
\addplot[thick,color=\blueColor] table[x=nodes,y=mean_action_count,col sep=comma] {navigation_experiment/data/s5_eager_experiment_processed.csv};
\addlegendentry{EG mean}
\end{axis}
\end{tikzpicture}